%% file: sn-article.tex
\documentclass[pdflatex,sn-nature]{sn-jnl}

\usepackage{graphicx}%
\usepackage{multirow}%
\usepackage{amsmath,amssymb,amsfonts}%
\usepackage{amsthm}%
\usepackage{mathrsfs}%
\usepackage[title]{appendix}%
\usepackage[dvipsnames, table]{xcolor}
\usepackage{textcomp}%
\usepackage{manyfoot}%
\usepackage{booktabs, tabularx}%
\usepackage{algorithm}%
\usepackage{algorithmicx}%
\usepackage{algpseudocode}%
\usepackage{listings}%
\usepackage{tikz}%
\usepackage{placeins}%
\usepackage{wrapfig}%

\definecolor{myblue}{RGB}{0,112,168}
\definecolor{myyellow}{RGB}{255,204,0}
\definecolor{mygreen}{RGB}{0,166,157}
\definecolor{myred}{RGB}{190,24,24}
\definecolor{mypink}{RGB}{204,121,167}
\definecolor{mypurple}{RGB}{122,31,167}
\newcommand{\colordot}[1]{\raisebox{-.5ex}{\tikz\draw[draw=none,fill=#1] (0,0) circle (4pt);}}
\newcommand{\mycirc}[1][black]{\textcolor{#1}{\ensuremath\bullet}}
\newcommand{\mysquare}[1][black]{\textcolor{#1}{\ensuremath\blacksquare}}

\theoremstyle{thmstyleone}%
\theoremstyle{thmstyletwo}%

\theoremstyle{thmstylethree}%

\begin{document}

\title{Recombination without composition leads to systematic limits of meta-learned neural networks for compositionality}

\author*[1,2]{\fnm{Hannah} \sur{Willkomm}}\email{hannah.willkomm@stud.tu-darmstadt.de}
\equalcont{These authors contributed equally to this work.}

\author[3]{\fnm{Tim} \sur{Woydt}}\email{tim.woydt@tu-darmstadt.de}
\equalcont{These authors contributed equally to this work.}

\author[3]{\fnm{Antonia} \sur{Wüst}}\email{antonia.wuest@tu-darmstadt.de}

\author[3]{\fnm{Moritz} \sur{Willig}}\email{moritz.willig@tu-darmstadt.de}

\author[1,3,4,5]{\fnm{Kristian} \sur{Kersting}}\email{kristian.kersting@tu-darmstadt.de}

\author[1,2,3,4]{\fnm{Constantin A.} \sur{Rothkopf}}\email{constantin.rothkopf@cogsci.tu-darmstadt.de}

\affil*[1]{\orgdiv{Centre for Cognitive Science}, \orgname{Technical University of Darmstadt}, \orgaddress{\street{Alexanderstrasse 10}, \city{Darmstadt}, \postcode{64283}, \country{Germany}}}

\affil[2]{\orgdiv{Institute of Psychology}, \orgname{Technical University of Darmstadt}, \orgaddress{\street{Alexanderstrasse 10}, \city{Darmstadt}, \postcode{64283}, \country{Germany}}}

\affil[3]{\orgdiv{Department of Computer Science}, \orgname{Technical University of Darmstadt}, \orgaddress{\street{Hochschulstrasse 10}, \city{Darmstadt}, \postcode{64289}, \country{Germany}}}

\affil[4]{\orgdiv{Hessian Center for Artificial Intelligence (hessian.ai)}, \orgaddress{\street{Landwehrstrasse 50a}, \city{Darmstadt}, \postcode{64293}, \country{Germany}}}

\affil[5]{\orgdiv{German Research Center for AI (DFKI)}, \orgaddress{\street{Hochschulstrasse 10}, \city{Darmstadt}, \postcode{64289}, \country{Germany}}}

\abstract{
Fodor and Pylyshyn argued that connectionist networks, lacking symbolic machinery, cannot support the structure-sensitive operations that systematic thought requires. Meta-learning for compositionality (MLC) has been offered as evidence that statistical learning alone suffices, a premise that has since been extended to human cognition and visual reasoning. We introduce a hierarchy of evidence for compositional competence, from benchmark accuracy through behavioural tests to grammar-level analysis, and evaluate MLC-trained Transformer and LSTM models. High benchmark accuracy mostly requires assigning new symbols to structures attested in meta-training. Systematic tests reveal within-range failures: accuracy declines with sequence and argument length, and the Transformer answers both parts of a two-rule chain correctly yet fails at their composition, while novel pairings of attested rule components succeed where single property changes fail at matched demonstration frequency. Success thus reflects recombination, not rule abstraction. MLC as trained provides no guarantee of systematicity, leaving Fodor and Pylyshyn’s challenge open. 
}

\maketitle

\section{Introduction}\label{sec1}
Since the beginning of artificial intelligence and cognitive science, an intense debate has centred on the nature of mental representations: is thought language-like, symbolic, thereby embodying compositionality, systematicity, and productivity \citep{Fodor2001}, or is it sub-symbolic, of a statistical nature alone \citep{smolensky1988proper}? 
A prominent recent entry in this debate by Lake and Baroni \citep{lake2023human} suggests that this debate is now resolved by providing a standard neural network trained through meta-learning for compositionality (MLC) as evidence for systematic generalization. Trained via MLC across many hidden grammars to infer the rules of a new grammar from a few examples, the network achieves high benchmark accuracy, although the claim of systematic generalization has since been questioned \citep{woydt2025fodor, goodale2026fodor}. 
The same claim, i.e. that practice on compositional episodes induces compositional competence, underlies earlier meta-learning approaches \citep{lake2019compositional,conklin2021meta}, has been proposed as a general principle \citep{irie2025overcoming} and as a theory of human compositionality \citep{russin2024human,binz2024meta}, and has been extended beyond language \citep{mondorf2026compositional}.
In each case, the evidence is the same: accuracy on held-out episodes drawn from the meta-training distribution. Whether such accuracy shows compositional competence is therefore a question about this entire line of work, and we address it here for the setting in which the claim was made most strongly \citep{lake2023human}.
Originally, Fodor and Pylyshyn \cite{FodorPylyshyn1988connectionism} argued that neural networks (NNs), as traditionally conceived, are inadequate models of mental representation, because they rely on associative statistical learning, which, as they argue, does not guarantee an ability to understand and generate novel expressions by systematically combining familiar elements according to learned rules, i.e., compositional generalization. Their claim was not that no NN could behave systematically, but that purely associative mechanisms, lacking symbolic structure, cannot guarantee such capabilities by design. 
In contrast, symbolic systems, including formal grammars or rule-based inference engines \citep{kemp2009structured, tenenbaum2011grow, lake2015human}, encode compositional representations and support structure-sensitive operations, enabling the kind of rule-based, general-purpose reasoning observed in human cognition \citep{FodorPylyshyn1988connectionism,Fodor2001,marcus2003algebraic}.  

Evaluating these claims requires disentangling performance from competence \citep{firestone2020performance,mitchell2026six}, an important but fundamentally hard task that should receive at least as much attention as developing new algorithms. 
Therefore, we use a hierarchy of progressively stronger evidence for compositional competence: benchmark evidence, behavioural evidence, and grammar-level evidence. 
The weakest evidence comes from standard benchmarking (Level~1), in which the grammar is operationalized as a dataset and evaluation reduces to accuracy on predefined train-test splits. For compositional generalization, however, one must demonstrate that performance under a particular evaluation distribution extends reliably to novel configurations that do not follow the statistical properties of the training data \citep{lake2018generalization, hupkes2020compositionality}. Otherwise, benchmark performance may rely on statistical regularities or shortcut strategies shared across training and test distributions \citep{geirhos2020shortcut,hupkes2020compositionality}.
To obtain stronger behavioural evidence (Level~2), evaluation must
assess whether a model continues to apply the relevant rules systematically when relevant grammar dimensions are varied beyond the statistical regularities encountered during meta-training \citep{hupkes2020compositionality,keysers2020measuring,kim2020cogs}. This targets structure-sensitivity at the level of the model's predictions, but does not establish that the mechanisms producing them are compositional. The strongest form of evidence is grammar-level evidence (Level~3), which requires recovering an explicit grammar from internal states and computations.

Guided by this hierarchy, we systematically evaluate compositional generalization in MLC-trained long short-term memory (LSTM) \citep{lake2019compositional} and Transformer \citep{lake2023human} sequence-to-sequence (seq2seq) architectures using the first two levels.
Benchmark and behavioural evidence can reveal failures of systematic rule application, but cannot by themselves establish its presence. This would additionally require recovering an explicit grammar from a model's internal computations (Level~3), which goes beyond identifying circuits for individual tasks \citep{wang2022interpretability,tang2026circuit}.
Using benchmark and behavioural evaluations, we find systematic failures of compositional generalization already within the meta-training range. Under controlled variations of query structure, the models fail to apply demonstrated rules of forms absent from meta-training, and to compose chains of demonstrated rules, even when each part is answered correctly in isolation.
 
We discuss these results in light of the enduring debate in AI and cognitive science and the implications for NN systems based on the considered architectures. 
First, the results show that MLC as currently trained provides no guarantee of systematic compositional generalization, even within the meta-training range. Thus, Fodor and Pylyshyn's challenge \cite{FodorPylyshyn1988connectionism} remains open.
Second, the results speak to the pervasive difficulty of disentangling performance and competence in AI systems, as much as it is in cognitive systems in general \citep{firestone2020performance,mitchell2026six}. 
Third, these results underscore the ongoing challenge of understanding the relationship among natural language, the language of thought, formal languages, and the implementation of these competencies in animals and humans \citep{kean2026evidence, fedorenko2024language,mahowald2024dissociating}. While the notion of a ``compositionality continuum'' has recently been proposed \citep{riveland_compositionality_2026} with Lake and Baroni's results \citep{lake2023human} cited as evidence of systematicity, our findings suggest that benchmark success is not sufficient evidence that the underlying compositional rules have been learned. 
Finally, the current study suggests that recent successes of neurosymbolic approaches in AI \citep{ellis2020dreamcodergrowinggeneralizableinterpretable, dinu2024symbolicaiframeworklogicbasedapproaches, wust2026synthesizing, shindo2025iclr} could guide investigations in cognitive science and neuroscience of how animals and humans achieve compositionality, systematicity, and productivity.

\section{Results}\label{sec2}

\subsection{Evaluating compositionality with episodic grammar tasks} \label{sec:evaluating_comp}

\begin{figure}[!t]
    \centering
    \includegraphics[width=1.0\textwidth]{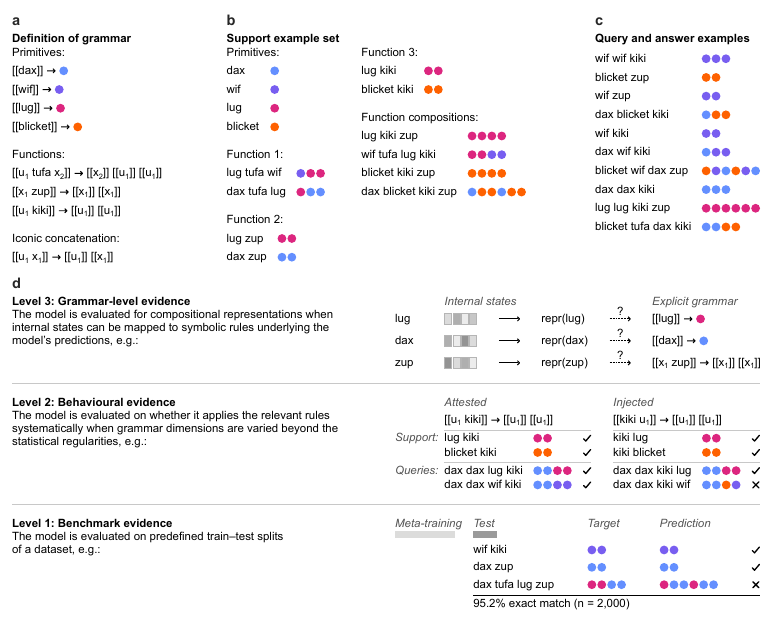}
     \caption{\textbf{Assessing systematic compositionality in a few-shot grammar-learning task.}
    \textbf{a,} Each episode has a hidden interpretation grammar comprising primitive mappings, function rules that combine or repeat their arguments, and a fixed concatenation rule. Here, $u$ denotes a single token, $x$ a non-empty string, and $F$ a function symbol. Double brackets mark expressions interpreted under the grammar.
    \textbf{b,} The support set ($n_s=14$) provides examples of the episode-specific mappings, rules and their compositions. It remains available to the model at test time without weight updates.
    \textbf{c,} At test time, the model uses the support set examples to predict the outputs of held-out queries from the same hidden grammar without weight updates.
    \textbf{d,} Benchmark evidence (Level~1) is obtained by measuring exact-match accuracy on queries drawn from the same meta-grammar and sampling regime as meta-training. The displayed $95.2\%$ is the replicated Transformer's accuracy on $2{,}000$ novel queries from the $200$ released validation episodes. Behavioural evidence (Level~2) is obtained by testing whether a demonstrated rule transfers under a controlled change in structure. The example compares the attested form $u_1\,F$ with the operator-initial form $F\,u_1$, which does not occur in meta-training. Grammar-level evidence (Level~3) would additionally require recovering an explicit grammar underlying the model's predictions from the model's internal computations.}
    \label{fig:overview}
\end{figure}

To evaluate compositional generalization in meta-learned networks, we use the seq2seq grammar task of ref.~\citep{lake2023human} (Fig.~\ref{fig:overview}a--c; see \nameref*{sec:methods} Sec.~\ref{sec:metaLearnSeq2Seq}, ``Meta-learning seq2seq grammar tasks'', and Sec.~\ref{sec:metaLearnCompositionalityMLC}, ``Meta-learning for compositionality (MLC) framework''). In each episode, a meta-grammar samples a new interpretation grammar and generates support and query pairs from its production rules under global and sequence-level constraints. Models must infer episode-specific meanings and rules from the support set to answer held-out queries without parameter updates. To test whether the results depend on the architecture, we evaluate two meta-learned architectures: an LSTM with key--value episodic memory \citep{lake2019compositional} and an encoder--decoder Transformer \citep{lake2023human} (Supplementary Section~\ref{secA}). 
We distinguish three levels of evidence for compositional competence, ordered from weakest to strongest (Fig.~\ref{fig:overview}d). Benchmark evidence (Level~1) is based on performance under the standard MLC protocol, where high accuracy does not establish whether demonstrated rules are applied beyond structures familiar from meta-training. 
Lake and Baroni \citep{lake2023human} themselves note that MLC is expected to succeed only when a test episode is in-distribution with respect to the meta-training episodes, and earlier work found that meta-learned sequence models tend to fit the meta-training task distribution rather than acquire a compositional inductive bias \citep{mitchell2021challenges}. Level 2 evaluation is designed to separate these two readings: whether the demonstrated rules are applied as rules, or whether accuracy tracks the structures practised during meta-training.
Accordingly, evaluation for behavioural evidence (Level~2) tests whether demonstrated rules transfer under systematic shifts in episode, sequence and rule structure beyond meta-training. Continued application under these shifts would support rule inference, whereas deterioration would show that accuracy depends on structures familiar from meta-training. Grammar-level evidence (Level~3) requires prediction-consistent recovery of an explicit grammar from model internals. Here, we only\linebreak 
address the first two levels.

\subsection{Benchmark accuracy is high within a narrow structural range}\label{sec:benchmark_interpolation}
\begin{figure}[!htbp]
    \centering
    \includegraphics[width=1\textwidth]{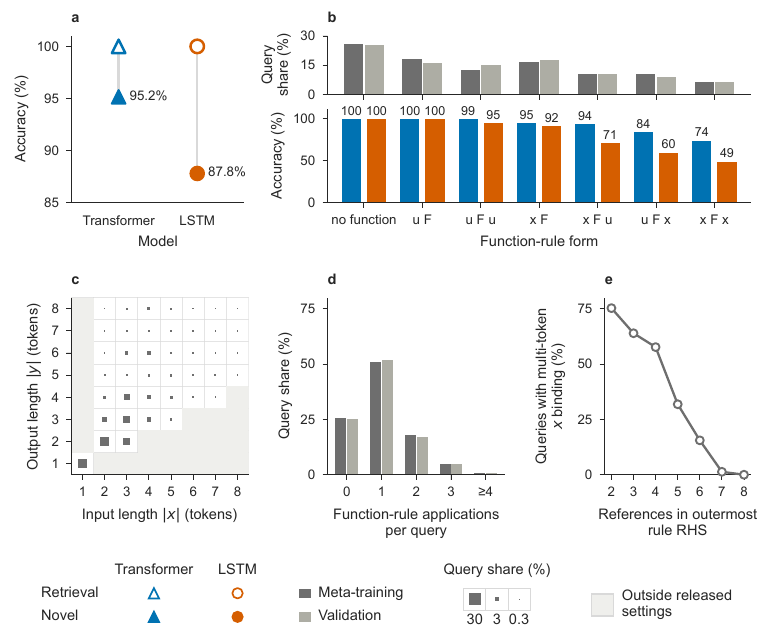}
    \caption{\textbf{High benchmark accuracy within a restricted space of grammar and sequence structures.}
    \textbf{a,} Exact-match accuracy for the replicated Transformer (blue) and LSTM (orange), both with context, on $2{,}000$ novel queries (filled) and $2{,}800$ support items (open) from $200$ released validation episodes. \textbf{b,} Meta-training and validation query shares (top) and validation accuracy (bottom) by outermost function-rule form. $F$ denotes a function symbol, $u$ a single token and $x$ a non-empty string. \textbf{c,} Joint distribution of input and output lengths across $1{,}000{,}000$ meta-training queries. Square area scales with query share (minimum width $0.25$\,mm). Grey cells fall outside the released benchmark parameters. \textbf{d,} Function-rule applications per query in both splits (four or more pooled). \textbf{e,} Among meta-training queries whose outermost rule takes a string argument, the share with at least one such argument spanning multiple tokens, by that rule's right-hand-side reference count.}
    \label{fig:benchmark_scope}
\end{figure}
At the benchmark-evidence level (Level~1), accuracy is high. The Transformer and LSTM reach $95.2\%$ and $87.8\%$ validation accuracy, respectively, and reproduce all support items (Fig.~\ref{fig:benchmark_scope}a). However, these aggregate scores mask substantial variation across episodes and query structures (Fig.~\ref{fig:benchmark_scope}b and Supplementary Section~\ref{sec:supp_episode_qualitative}). Episode~\#133, for example, reaches only $37$--$42\%$ accuracy under repeated stochastic decoding for our two models and the three released checkpoints of ref.~\citep{lake2023human}, ranking lowest or second lowest among the $200$ validation episodes for each (Supplementary Table~\ref{tab:supp_worst_episodes}).
Both models exceed $99\%$ on queries without a function application or with $u\,F$ as the outermost function rule, but score only $74.0\%$ (Transformer) and $48.9\%$ (LSTM) with $x\,F\,x$. The four categories on which both models exceed $90\%$ accuracy account for $74.2\%$ of validation queries, so the aggregate score is dominated by the easiest structures.

To assess what this performance implies for compositional generalization, we examined the grammar and sequence-level novelty required by validation.
The meta-grammar permits six left-hand-side forms and non-erasing right-hand sides containing two to eight references, but otherwise restricts functions to unary or binary, fixes operator position, prohibits argument erasure and maps each primitive to one token (Fig.~\ref{fig:benchmark_scope}b). Sampling is concentrated on shorter rules, with two-reference rules comprising $60.0\%$ of sampled function rules and those with five to eight references only $6.4\%$ (Supplementary Fig.~\ref{fig:supp_meta_grammar_stats}). Validation introduces little grammar-level novelty. Of its $200$ grammars, $178$ use a combination of three function-rule forms seen in meta-training ($147$ in the same order), $21$ recombine attested forms, and only one contains an unattested form (Supplementary Section~\ref{sec:B_grammar_structure}). 
At the sequence level, eight-token input and output bounds restrict how rules in sequences can be instantiated. Inputs of two to four tokens comprise $69.3\%$ of meta-training queries, and $76.5\%$ of queries apply at most one function rule  (Fig.~\ref{fig:benchmark_scope}c,d). The output bound also couples rule length to binding width, so eight-reference rules take only single-token arguments, and thus cannot take the output of another function as an argument (Fig.~\ref{fig:benchmark_scope}e). 
Validation therefore primarily tests new symbol assignments within narrow bounds and structures familiar from meta-training. High benchmark accuracy shows that the models handle new symbol assignments within this familiar range, but does not establish whether they transfer to rare or absent query structures and rule forms. 
This characterization of the benchmark does not by itself contradict \citep{lake2023human}, which reports test accuracy as evidence of systematicity relative to human performance on the same episodes. Instead, it shows what the validation set can and cannot test, and thus why benchmark accuracy on its own is Level 1 evidence.

\subsection{Meta-training design and model size shape benchmark performance}\label{sec:level1c}
\begin{figure}[!htbp]
    \centering
    \includegraphics[width=1\textwidth]{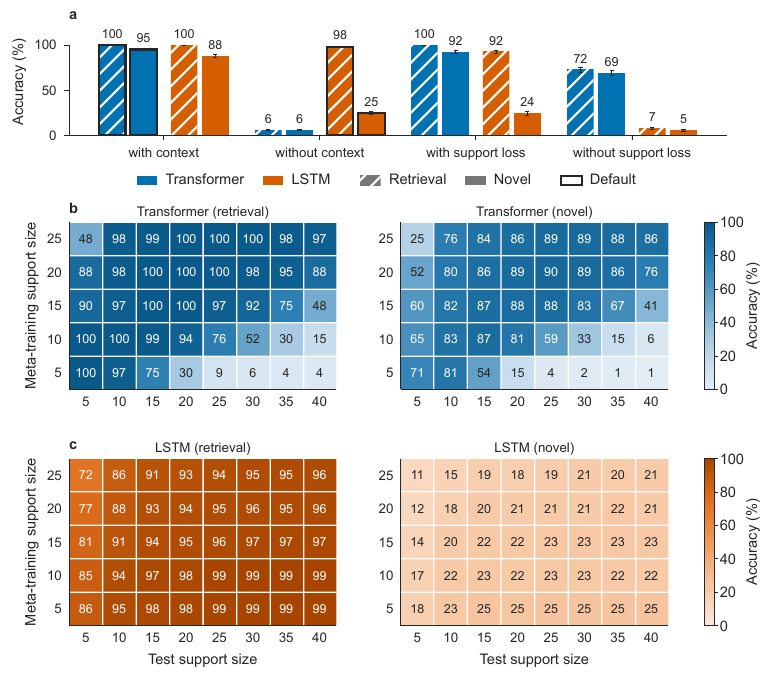}
   \caption{\textbf{Retrieval and generalization depend on support-set size and training configuration.}
    \textbf{a,} Exact-match accuracy on support items (hatched bars) and novel queries (plain bars) on the $200$ released validation episodes for the Transformer (blue) and LSTM (orange) with varied meta-training configurations. Without context, the LSTM still accesses the support pairs through its key--value memory, while the Transformer receives the query alone. For the support-loss ablation, the Transformer is trained with context and the LSTM without it. Black outlines mark the original default input configurations, and error bars show $95\%$ episode-clustered bootstrap confidence intervals. \textbf{b,c,} Support-item and novel-query accuracy across meta-training support sizes (rows, $5$--$25$) and test support sizes (columns, $5$--$40$) for the Transformer with context (\textbf{b}) and the LSTM without context (\textbf{c}). Each cell contains $500$ generated episodes with ten novel queries per episode. Each configuration in \textbf{a--c} was trained once.}
    \label{fig:meta_training_ablations}
\end{figure}

To test whether the remaining benchmark errors reflect insufficient capacity, we varied model width, depth and dropout (Supplementary Section~\ref{secE}). Increasing Transformer width from $32$ to $128$ raised novel-query accuracy from $22.1\%$ to $93.4\%$, but doubling it again to $256$ added merely $2.1$ percentage points, and greater depth did not consistently improve accuracy (Extended Data Fig.~\ref{fig:supp_hidden_size_accuracy}).

The support set provides the only evidence of episode-specific meanings and rules, so we tested its effects on retrieval and novel-query accuracy using the default architectures (Fig.~\ref{fig:meta_training_ablations}a). Without the support pairs serialized with each query (\emph{without context}), the Transformer's retrieval and novel-query accuracy fall from $100\%$ and $95\%$ to about $6\%$. The LSTM's key--value memory retains $98\%$ retrieval accuracy without context, but its novel-query accuracy falls from $88\%$ to $25\%$. Without the auxiliary loss for reproducing support outputs, retrieval and novel-query accuracy fall to $72\%$ and $69\%$ for the Transformer and to $7\%$ and $5\%$ for the LSTM without context (Fig.~\ref{fig:meta_training_ablations}a).

To test adaptation to evidence quantity, we varied support-set size from the standard $14$ items to $5$--$25$ during training and $5$--$40$ at test (Fig.~\ref{fig:meta_training_ablations}b,c). For the Transformer, novel-query accuracy generally increased with support size when training and test support sizes matched, but large mismatches reduced it. With five test items, novel-query accuracy was $71\%$ after training with five support examples, but $25\%$ after training with $25$ examples. Adding support examples at test did not necessarily improve accuracy. After meta-training with $15$ support examples, novel-query accuracy fell from $88\%$ to $41\%$ when test support increased from $25$ to $40$ examples. Because support is serialized with each Transformer query, varying its size changes input length and composition alongside evidence quantity. 
That accuracy tracks the match between meta-training and test support statistics, rather than the amount of evidence available, is consistent with findings that in-context learning in Transformers is based on, and remains bounded by, the distributional properties of the training stream \citep{chan2022data}, and with the sensitivity of meta-learned models to how meta-training tasks are constructed \citep{mitchell2021challenges}.
In the LSTM sweep without context, larger test support sets improved retrieval, but novel-query accuracy remained at $11$--$25\%$ (Fig.~\ref{fig:meta_training_ablations}c). With context, the LSTM meta-trained with $14$ support examples per episode maintained $80$--$83\%$ novel-query accuracy and near-perfect retrieval across $14$--$40$ test examples (Supplementary Fig.~\ref{fig:supp_nsupport_default_models}). Therefore, the remaining benchmark errors do not reflect limited capacity, and the following evaluations use both models in their best configuration, with context.

\subsection{Accuracy declines with sequence length and argument width}\label{sec:behavioral_ood}
\begin{figure}[!t]
\centering
\includegraphics[keepaspectratio,width=1.0\linewidth ]{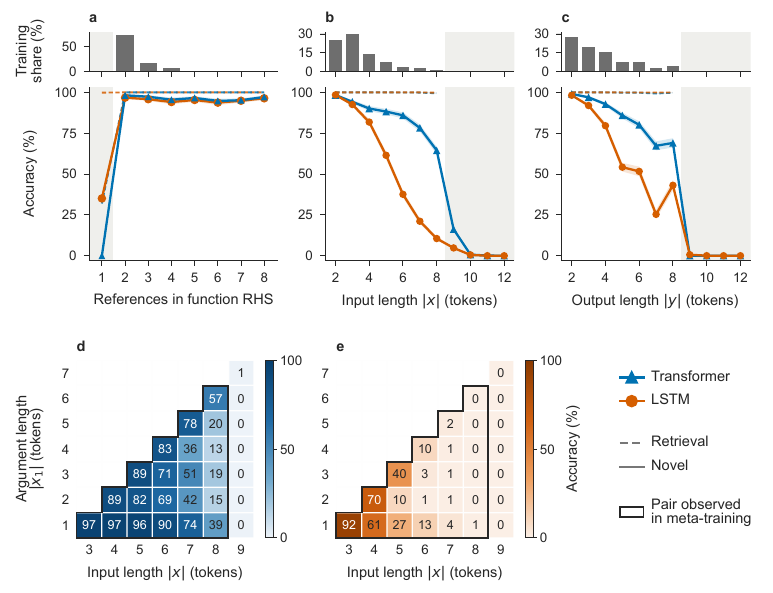}
\caption{\textbf{Accuracy under shifts in rule length, sequence length and argument width.} Exact-match accuracy for the replicated Transformer (blue) and LSTM (orange), both with context. In \textbf{a--c}, upper bars show meta-training shares, solid lines show novel-query accuracy, and dashed lines retrieval of support items. Shaded areas around the curves indicate $95\%$ episode-clustered bootstrap confidence intervals. \textbf{a,} Accuracy by the number of references in a rule's right-hand side. Novel queries apply one unary or binary rule to single-token arguments. Rules with one reference, which output a single argument, do not occur in meta-training. \textbf{b,} Accuracy by input length, with query outputs limited to eight tokens. \textbf{c,} Accuracy by output length, with query inputs limited to eight tokens. In both panels, dashed curves are grouped by support-item length. \textbf{d,e,} Novel-query accuracy for the injected reversal rule $x_1\,F\,x_2\rightarrow[x_2]\,[x_1]$, by input length and first-argument width $|x_1|$, for the Transformer (\textbf{d}) and LSTM (\textbf{e}). We score only queries whose support set includes an example of the reversal rule. The grey background (\textbf{a--c}) marks values outside the meta-training range, and black outlines (\textbf{d,e}) mark pairs within it.}\label{fig:ood_testing}
\end{figure}
Because benchmark evaluation largely preserves meta-training structure, high accuracy does not establish transfer under structural change. 
To obtain stronger behavioural evidence (Level~2), we therefore systematically varied grammar and sequence parameters within and beyond their meta-training ranges (Fig.~\ref{fig:ood_testing}; see Methods Sec.~\ref{sec:procOOTEval}, ``Procedural out-of-training (OOT) evaluation'').

For grammar-level manipulations, expanding the primitive inventory does not reduce accuracy 
(Supplementary Fig.~\ref{fig:supp_inventory_grid}). Both models exceed $93\%$ accuracy across single applications of function rules with right-hand sides containing two to eight references, including rarely sampled longer rules (Fig.~\ref{fig:ood_testing}a). These queries substitute at most two new atoms into a demonstrated pattern, so additional references lengthen the output without increasing compositional demands. For unseen single-reference function rules, accuracy drops for both models, with the Transformer collapsing to $0\%$ (Fig.~\ref{fig:ood_testing}a). 

At the sequence level, meta-training is dominated by short sequences within eight-token inputs and outputs. As input length increases from two to eight, novel-query accuracy falls from $98.4\%$ to $64.5\%$ for the Transformer and from $98.5\%$ to $10.5\%$ for the LSTM. Accuracy declines similarly with output length, while support-item accuracy remains $\geq99\%$ (Fig.~\ref{fig:ood_testing}b,c). The decline with input length persists when grammar and support set are held fixed (Supplementary Section~\ref{sec:C_within}), and for queries that combine several primitives without function applications (Supplementary Section~\ref{sec:C_primitives}).
Long inputs are handled more reliably when the output is short, whereas at $(|x|,|y|)=(8,8)$ accuracy is only $58\%$ for the Transformer and $4\%$ for the LSTM (Supplementary Fig.~\ref{fig:supp_length_grid}). Even among length-matched queries, performance is sensitive to binding structure, inconsistent with systematic transfer. For a demonstrated reversal rule at $|x|=7$, the Transformer's accuracy is $74$--$78\%$ when $|x_1|=1$ or $|x_2|=1$ but $36$--$51\%$ when both bindings are multi-token (Fig.~\ref{fig:ood_testing}d,e).

Beyond the eight-token meta-training length, performance collapses. Conditional on the correct eight-token prefix of a longer target, both models terminate after eight tokens with probability $\geq 0.98$, indicating a learned emission limit. Accordingly, the models produce almost no targets longer than eight tokens, for support items and novel queries alike (Fig.~\ref{fig:ood_testing}c and Supplementary Sections~\ref{sec:C_plane} and~\ref{sec:C_window}). By contrast, the Transformer reproduces at least $85\%$ of support items at input lengths nine and ten (Supplementary Section~\ref{sec:C_plane}). Yet even for novel queries with $|y|\leq8$, extending the input by a single token beyond meta-training lowers accuracy from $64.5\%$ at $|x|=8$ to $16.3\%$ at $|x|=9$, and below $1\%$ from $|x|=10$ onwards (Fig.~\ref{fig:ood_testing}b). 
The collapse beyond eight tokens confirms, for the grammar task, the limits of extrapolation reported in \citep{lake2023human} on the SCAN length split and in \citep{goodale2026fodor} for longer outputs, and is expected under Lake and Baroni’s own in-distribution condition. The declines within the meta-training range, by input length at fixed grammar and by binding width at matched length, are not covered by that condition, because every such query is drawn from the trained bounds. This shows that transfer depends on the practised sequence and binding statistics even where the episode is in-distribution.

\subsection{Rule forms outside the meta-grammar fail to transfer}\label{sec:rule_forms}
\begin{figure}[!t]
    \centering
    \includegraphics[width=1.0\textwidth]{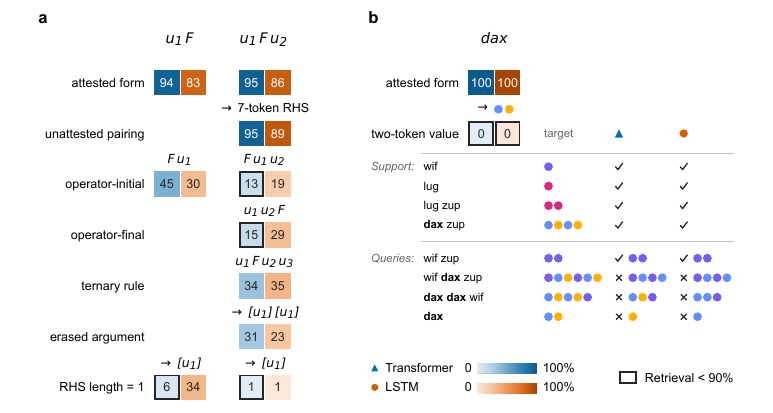}
    \caption{\textbf{Transfer of demonstrated rules fails beyond the meta-grammar.} Novel-query exact-match accuracy for the replicated Transformer (blue, left cells) and LSTM (orange, right cells), both with context. Black outlines mark retrieval accuracy below $90\%$. \textbf{a,} Injected function rules with atomic arguments. The top row shows controls with unary and binary forms observed in meta-training. The rows below change the rule, with the changed left- or right-hand side shown above the cells. The second row pools $19$ pairings of the binary form with seven-reference right-hand sides seen only with other rule forms in meta-training. We score only queries that apply the injected rule. \textbf{b,} Bare-query accuracy for a primitive whose bare demonstration is withheld, with a one-token (attested) or two-token value. Below, an example based on the episode of Fig.~\ref{fig:overview}, in which $\mathtt{dax}$ has the value $\mathtt{BLUE\ YELLOW}$: four of the $14$ support examples and four novel queries with their targets and the responses of both models. Both models reproduce all support examples.}\label{fig:rule_forms}
\end{figure}

To test whether demonstrated rules transfer beyond the rule forms of the meta-grammar, we evaluated rule forms absent from meta-training using atomic arguments, so that one support example fully specifies the argument mapping (Fig.~\ref{fig:rule_forms}a and Supplementary Fig.~\ref{fig:supp_battery_overview}). As a control, we combined the binary form $u_1\,F\,u_2$ with $19$ seven-reference right-hand sides that occur only with other rule forms in meta-training. The Transformer and LSTM reach $95\%$ and $89\%$ accuracy, close to $96\%$ and $90\%$ on attested seven-reference controls. Because the eight-token output bound largely restricts seven-reference rules to single-token arguments, these rules cannot be distinguished from attested string-slot rules by their examples. Only the hidden grammar is new. Accuracy falls when a demonstrated rule instead changes a property fixed by the meta-grammar, despite equally frequent support demonstrations. For operator positions unseen for their arity, ternary function rules and binary rules omitting one input argument, novel-query accuracy is only $13$--$45\%$. For altered binary operator positions, Transformer support-item accuracy is $68$--$69\%$, so failure already occurs on the demonstrations and these particular conditions do not isolate transfer. However, novel-query accuracy also remains low when support reproduction exceeds $90\%$.

The meta-grammar maps each primitive to one token, whereas function right-hand sides contain at least two argument references and thus produce at least two tokens for atomic arguments. Unary identity and binary first-argument projection violate this distinction by requiring one-token function outputs (Fig.~\ref{fig:rule_forms}a and Supplementary Fig.~\ref{fig:supp_battery_overview}). On these rules, the models reach only $1$--$34\%$ accuracy and mostly output two tokens instead of one (Supplementary Table~\ref{tab:supp_predicted_answers}). The Transformer reproduces only $24$--$38\%$ of the support examples of these rules, although their outputs are given in the support. When bare primitive mappings are withheld, both models recover every one-token mapping from composed supports, but never a two-token mapping, emitting a single token instead (Fig.~\ref{fig:rule_forms}b). Poor support reproduction cannot explain every failure. For example, both models exactly reproduce supports jointly implying $\mathtt{dax}\rightarrow\mathtt{BLUE\ YELLOW}$ yet answer the bare query $\mathtt{dax}$ with a single token, matching no single consistent episode grammar (Fig.~\ref{fig:rule_forms}b). In the evaluation episodes, both models likewise answer the bare query with a single token whenever they reproduce all support examples (Supplementary Section~\ref{sec:G_overview}). 
Failure on rule forms absent from meta-training is, by itself, what \citep{lake2023human} anticipates for forms the model has not practised. However, our matched-frequency control demonstrates that unattested pairings of attested components are solved at attested-level accuracy, whereas a change to one property of the rule form, demonstrated equally often, is not, which is the pattern expected if in-context generalization proceeds by substitution into demonstrated structures rather than by rule induction \citep{chan2022transformers,chang2025coverage}.
\subsection{Function composition fails within the meta-training range}\label{sec:composition}
\begin{figure}[!t]
    \centering
    \includegraphics[width=1.0\textwidth]{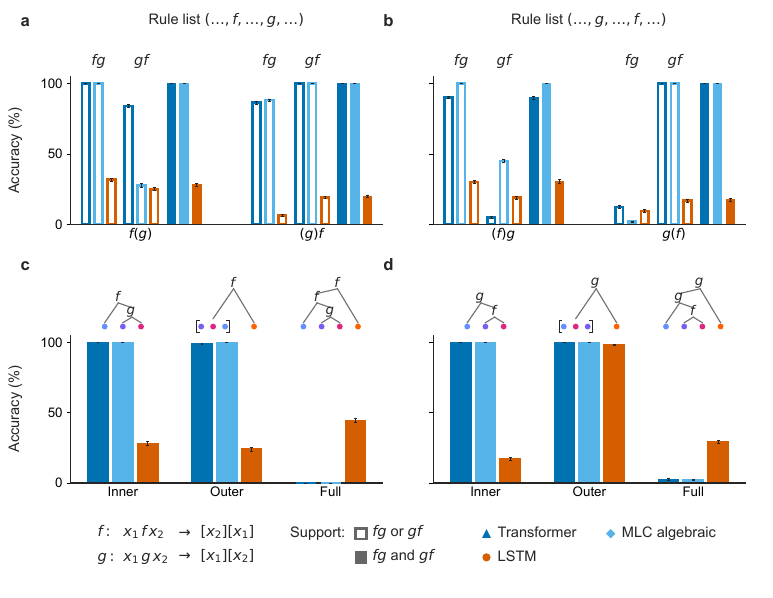}
    \caption{\textbf{Composition of demonstrated function rules fails within the meta-training range.}
    Novel-query exact-match accuracy for the replicated Transformer (blue), the released \emph{MLC(algebraic only)} checkpoint (light blue) and the replicated LSTM (orange), with a reversing function $f$ and an order-preserving function $g$ demonstrated in the same support set. \textbf{a,b,} Mixed two-application queries with $f$ (\textbf{a}) or $g$ (\textbf{b}) earlier in the rule list, labelled by their bracketing. Support demonstrates one mixed form (open bars, form given above) or both (filled bars). \textbf{c,d,} Inner expression, outer application and full query (trees above the bars) for $fgf$ with $f$ earlier (\textbf{c}) and $gfg$ with $g$ earlier (\textbf{d}), with both mixed forms demonstrated. In the outer query, the inner expression is replaced by primitive words with its correct output (bracketed in the trees; colours identify primitives). The three queries are decoded independently. Bars show accuracy over $500$ grammar pairs with ten queries per form, and error bars show $95\%$ bootstrap confidence intervals from resampling grammar pairs.}\label{fig:composition}
\end{figure}
To determine whether the observed limitations extend to function composition within the trained range, we demonstrated two attested two-reference rules, a reversing function $f$ and an order-preserving function $g$, in the same support set (Fig.~\ref{fig:composition}; see Methods, ``Arrangements of two rules''). We tested queries with up to three applications of these functions, such as $fg$ or $fgf$, for both orders of $f$ and $g$ in the hidden rule list, because the bracketing of mixed forms depends on this order. The support demonstrates $fg$, $gf$ or both and unambiguously determines the query targets, and all checkpoints reproduce every support example. When a mixed form is demonstrated, the Transformer answers novel queries of that form with $89.9$--$100\%$ accuracy, whereas the LSTM remains below $32\%$ (Fig.~\ref{fig:composition}a,b). Transfer to the mixed form absent from support is unreliable, with $84.1$--$86.7\%$ for the Transformer when $f$ is earlier in the rule list but only $5.1$--$12.4\%$ when $g$ is earlier.

With both mixed forms demonstrated, the Transformer still fails on alternating three-application queries although it answers their parts. It answers $fff$ queries of the same length perfectly, but none of the $fgf$ queries with $f$ earlier and $2.6\%$ of the $gfg$ queries with $g$ earlier, and its $fgf$ outputs mostly match $fff$ ($80.5\%$; Supplementary Section~\ref{sec:G_arrangements}). When the inner expression and the outer application of these queries are posed separately under the same support, it answers both correctly in $98.8\%$ and $100\%$ of cases, whereas the LSTM does so in only $6.9\%$ and $17.1\%$ (Fig.~\ref{fig:composition}c,d). The released \emph{MLC(algebraic only)} checkpoint, meta-trained without heuristic targets, shows the same pattern (Fig.~\ref{fig:composition}c,d), so these failures cannot be attributed to such biases. The responses to the full queries are thus inconsistent with the responses to their parts, although the support is the same. Such inconsistencies may arise because each query is decoded separately, without access to other responses of the episode.

All three released checkpoints of ref.~\citep{lake2023human} show similar sequence- and rule-structure limitations (Supplementary Sections~\ref{secC} and \ref{secG}). 
Unlike the collapse beyond the meta-training length and the rule-form results, the composition failure falls outside the limitations acknowledged in \citep{lake2023human} and reported in \citep{woydt2025fodor, goodale2026fodor}: the episode, the sequence lengths, and the rule forms all lie within meta-training, both rules are demonstrated, every support item is reproduced, and each part of the query is answered correctly in isolation. It thereby tests systematicity in the sense of \citep{lake2023human} directly, rather than productivity or transfer to unpractised forms, and parallels the parts-correct, whole-wrong dissociation reported for Transformers on algorithmic composition \citep{thomm2024limits} and for large language models \citep{press2023measuring}. 
Collectively, failures to compose demonstrated rules and to transfer them beyond the fixed structural constraints of the meta-grammar show that the evaluated models do not apply the demonstrated rules systematically. The models handle new symbol assignments within familiar structures (Level~1 evidence) but are sensitive to sequence bounds, rule forms, and rule composition (Level~2 evidence). Therefore, high aggregate benchmark performance does not establish that the models systematically infer and apply the rules demonstrated in the support.

\section{Discussion}\label{sec3}
Fodor and Pylyshyn's \cite{FodorPylyshyn1988connectionism} assertion that associative statistical learning in NNs is not sufficient to achieve compositional generalization has recently been challenged by meta-learning on compositional episodes, which has been reported to induce compositional competence \citep{lake2019compositional,conklin2021meta}, most prominently the MLC framework by Lake and Baroni \cite{lake2023human}. Based on these results, MLC has been proposed as a general programme of ``incentives and practice'' to overcome classic challenges of NNs \citep{irie2025overcoming}, as a theory of human compositionality \citep{russin2024human,binz2024meta}, and has been extended beyond language to the domain of abstract spatial reasoning \citep{mondorf2026compositional} based on the Abstraction and Reasoning Corpus (ARC) \citep{chollet2019measure}. 
The evidence provided in all of this work is 
accuracy on held-out episodes drawn from the meta-training distribution, which, however, reflects performance under a particular evaluation distribution rather than competence \citep{firestone2020performance,mitchell2026six} and does not by itself establish compositional generalization \citep{hupkes2020compositionality}. 
We examined what this evidence shows in the setting where the claim was made most forcefully \citep{lake2023human}. 
In our hierarchy of evidence for compositional competence, neither of the two levels addressed here yields sufficient evidence of compositional generalization for MLC.

Benchmark evidence (Level~1) reduces to performance on predefined train-test splits, where the MLC framework shows high aggregate accuracy (Section \ref{sec:evaluating_comp}). However, even at this weakest criterion, high accuracy provides limited evidence of compositional generalization. Our analyses of the released episodes (Section \ref{sec:benchmark_interpolation}) show that validation largely reuses the rule-form combinations, short sequences, and few rule applications of meta-training, and its aggregate score is dominated by the easiest query structures, and that the output bound prevents long rules from ever binding multi-token arguments, so high accuracy does not establish transfer beyond them. The remaining benchmark errors are not capacity-limited, and accuracy tracks the match between meta-training and test support statistics rather than the quantity of evidence (Section \ref{sec:level1c}), consistent with in-context learning being shaped and bounded by the distributional properties of the training stream \citep{chan2022data} and with earlier reports that meta-learned models fit the meta-training task distribution rather than acquiring a compositional inductive bias \citep{mitchell2021challenges}. This characterization does not contradict Lake and Baroni \citep{lake2023human}, which reports test accuracy as evidence of systematicity relative to human performance, but clarifies what the benchmark can and cannot test.
 
Behavioural evidence (Level~2) shows that neither architecture within the MLC framework continues to apply the episode grammar under controlled structural shifts. 
Some of our findings fall within what Lake and Baroni anticipated \citep{lake2023human}, as they concluded that MLC succeeds when a new episode is in-distribution with respect to the meta-training episodes, and accordingly
novel-query accuracy 
collapses beyond the meta-training input and output length,  as previously shown for longer outputs \citep{goodale2026fodor} and for the SCAN length split in \citep{lake2023human}: both models terminate at the meta-training maximum, consistent with evidence of end-of-sequence prediction in length extrapolation \citep{newman2020eos}. Accuracy also drops when a demonstrated rule changes a property fixed by the meta-grammar, such as operator position or arity (Sections \ref{sec:behavioral_ood} and \ref{sec:rule_forms}). 
Three further findings are not covered by that condition, because the episodes, lengths, and rule forms involved all fall within the meta-training range. First, accuracy declines with input and output length and with argument width at matched length inside the trained range (Section \ref{sec:behavioral_ood}). Second, unattested pairings of attested rule components are solved at attested-level accuracy, whereas a single property change at identical demonstration frequency is not, and the models reproduce supports jointly implying a two-token mapping yet answer the bare query with one token (Section \ref{sec:rule_forms}). Third, with both mixed rule forms demonstrated and every support item reproduced, the Transformer answers the inner and outer applications of a two-rule chain correctly in isolation yet fails the composed query, defaulting to a single repeated rule, with and without human-like biases in meta-training (Section \ref{sec:composition}). This last dissociation tests systematicity in the sense of \citep{lake2023human} directly, rather than productivity or transfer to unpractised forms, and parallels the parts-correct, whole-wrong pattern reported for Transformers on algorithmic composition \citep{thomm2024limits} and for large language models \citep{press2023measuring}.
Yet both models remain accurate on single applications of rare long rules. 

Behavioural evaluation therefore provides evidence against systematic application of structure-sensitive operations. Success where familiar configurations can be reused with new symbols and failure where demonstrated rules must be composed or extended beyond the structural constraints of meta-training would be expected if the models interpolate within the meta-training constraints instead of abstracting 
the demonstrated rules \citep{dziri2024faith, deletang2023neural}. Such reuse would suffice for accurate prediction within the narrow meta-training distribution \citep{geirhos2020shortcut,bayat2025the}. 
This account has independent support. In-context generalization in Transformers has been found to be exemplar-based where in-weights generalization is rule-based \citep{chan2022transformers}, and the coverage principle \citep{chang2025coverage} holds that compositional generalization in Transformers largely reduces to substituting fragments observed in shared contexts, with coverage requirements that grow with the number of composed components rather than following from the components alone; the matched-frequency controls of Section \ref{sec:rule_forms} and the parts-correct, whole-wrong result of Section \ref{sec:composition} are what both predict. The convergent ablation in \citep{mondorf2026compositional}, in which removing demonstrated two-step compositions from meta-training reduces three-step accuracy from 87\% to 21\%, suggests the same dependence on practised compositions in a visual domain.

Grammar-level evidence (Level~3) requires recovering an explicit grammar from a model's internal states and showing that this grammar accounts for its predictions. 
A model may behave consistently with a grammar without an underlying compositional representation of the grammar \citep{FodorPylyshyn1988connectionism}, so such evidence cannot be obtained from behaviour alone. 
The behavioural evidence (Level~2) of systematic failures under controlled structural shifts, however, is sufficient to reject the claim that MLC as currently trained resolves Fodor and Pylyshyn's \citep{FodorPylyshyn1988connectionism} challenge, regardless of the models' internal representation. Analysing the representation directly \citep{wang2022interpretability, yang2025emergent, tang2026circuit} could nonetheless build on our behavioural findings to reveal how the models succeed within familiar structures and why this does not extend to rule composition. Such analyses have so far been limited to circuits for individual tasks, such as a single function application \citep{tang2026circuit}, rather than a whole grammar, and their interventions can succeed without revealing the actual mechanism a model uses \citep{makelov2024subspace}. 

These findings carry implications beyond the MLC setting. 
Because the approaches that share MLC’s premise \citep{lake2019compositional,conklin2021meta,irie2025overcoming,mondorf2026compositional} rest on the same form of evidence, the Level 1 limitation applies to them directly, and the Level 2 tests introduced here can be transferred to each. 
If simple compositional rules cannot be reliably induced under controlled meta-learning conditions, then larger foundation models trained on unstructured data may face an analogous problem. The scale and surface diversity of their training corpora may mask the absence of structure-sensitive behaviour, enabling high performance through memorization and statistical interpolation 
\citep{meszaros2024rule}. Convergent evidence supports this concern: symbolic perturbations of mathematical problems cause sharp performance drops in large language models 
\citep{mirzadeh2024gsmsymbolic}, 
arithmetic has been found to be solved through bags of heuristics rather than coherent algorithms \citep{nikankin2024arithmetic}, Transformer predictions 
are often well described by n-gram co-occurrence statistics \citep{nguyen2024ngram}, accuracy tracks pretraining frequency  
\citep{razeghi2022impact,mccoy2024embers}, and 
apparently competent models fail on structurally simple reasoning tasks \citep{gontier2020measuring, nezhurina2024alicewonderlandsimpletasks}. 

The present results can be seen as an additional example of the pervasive difficulty of disentangling performance and competence in AI and cognitive systems in general \citep{firestone2020performance,mitchell2026six}. Output accuracy can dissociate from abstract rule competence, with correct answers arising from shortcuts and incorrect answers despite identification of the intended rule \citep{beger2025ai}. 
The dissociation is sharpest in Section 2.6, where correct answers on each part of a query coexist with failures on their composition. 
More generally, the relationship among natural language, the language of thought, formal languages, and the implementation of these competencies in animals and humans remains largely unclear \citep{fedorenko2024language,mahowald2024dissociating, kean2026evidence}. 
Our evaluation targets algebraic competence relative to the episode grammar and does not test the comparative claim that MLC errs where humans err \citep{lake2023human}, particularly because human performance under the conditions of Sections 2.5 and 2.6 currently remains unknown. Proposals that treat MLC as a model of human compositional learning \citep{russin2024human,binz2024meta} therefore inherit the Level 1 limitation but are not refuted based on the current work. 
Our results suggest that aggregate benchmarks 
are insufficient to validate compositional abilities \citep{dziri2024faith}, and that systematic evaluation along multiple structural axes is needed to assess rule abstraction \citep{hupkes2020compositionality}. 
The recently proposed ``compositionality continuum'' \citep{riveland_compositionality_2026} illustrates this point. 
Its axes are defined by neural implementation, yet meta-learned Transformers are placed high on both axes on the basis of Lake and Baroni's benchmark results \citep{lake2023human}. However, such scores do not establish compositional competence, and thus cannot by themselves locate a system on the continuum.

Finally, the limitations observed here are consistent with the central argument of Fodor and Pylyshyn: pure connectionist learning, without mechanisms that explicitly represent and operate on constituent structure, has not, in the models evaluated here, reliably produced systematic compositionality.  Broader meta-training may extend the range of structures on which such models succeed, but would not by itself guarantee compositional generalization beyond that range. 
The results in Section 2.6 sharpen this point: the composition that fails is one both of whose constituents are practised, so extending practice to the composition itself would add one more attested structure rather than the ability to derive it, which is precisely the distinction between practised and guaranteed generalization that Fodor and Pylyshyn drew. Whether meta-training on a wider distribution leaves the composition failure in place, or merely moves it, is a direct test of the ``incentives and practice'' programme \citep{irie2025overcoming} that our protocol makes possible. 
This does not foreclose the possibility that neural architectures could support compositionality under different inductive biases or with explicit structural scaffolding. Neurosymbolic approaches that incorporate symbolic constraints, structured intermediate representations, or rule-verification objectives could, in principle, provide the compositional representations and structure-sensitive operations that current systems lack \citep{ellis2020dreamcodergrowinggeneralizableinterpretable, dinu2024symbolicaiframeworklogicbasedapproaches, wust2026synthesizing, shindo2025iclr}.

While much attention is still devoted to devising new NN architectures, training data sets, and benchmarks, progress on this front will require at least as much attention to devising evaluation protocols to assess competence beyond performance. Such protocols must separate 
productivity, systematicity, and inferential coherence \citep{Fodor2001, hupkes2020compositionality}, the three dimensions of compositionality whose conflation has obscured the limits of current systems. 
Our own results illustrate the value of this separation: the length-bound failures concern productivity, which Lake and Baroni \citep{lake2023human} explicitly set aside, whereas the composition failures concern systematicity, which they claimed. However, the two are indistinguishable in an aggregate score. 
Even evaluations explicitly designed to measure compositional generalization can leave important dimensions of difficulty uncontrolled, since disjoint train and test compositions can still vary substantially in their structural similarity \citep{camposampiero2026scalable}. Benchmark designs that independently control for structural complexity, compositional demand, and lexical overlap \citep{keysers2020measuring, kim2020cogs} offer a more principled path toward valid claims of compositional generalization than current aggregate evaluation methods permit. More fundamentally, whether NNs have acquired systematic compositionality cannot be settled by benchmark accuracy alone, but requires behavioural and ultimately grammar-level evidence.

\section{Methods}\label{sec:methods}

\subsection{Meta-learning seq2seq grammar tasks}
\label{sec:metaLearnSeq2Seq}
Building on ref.~\cite{lake2023human}, we cast compositional sequence transduction as episodic meta-learning over latent interpretation grammars. 
Each episode $e=(G,S,Q)$ is defined by a hidden interpretation grammar $G$, a support set $S=\{(x_i,y_i)\}$ of input--output pairs and query inputs $Q=\{x_j\}$ sampled from the same grammar. The learner must predict the withheld outputs for $Q$ given $S$. This setting does not require the learner to output an explicit rule description for the given episode. Although intended to reward implicit rule extraction, the task may admit simple queries that can be answered by recombining demonstrated examples without learning the rules of $G$.

\subsubsection{Interpretation grammar} 
An interpretation grammar $G$ is a finite ordered set of compositional production rules, i.e., primitives and functions, over an input vocabulary $\mathcal{I}$ of pseudolanguage tokens and an output vocabulary $\mathcal{O}$ of colour tokens. 
Primitive rules map individual input symbols $i\in\mathcal{I}$ to terminal output tokens $o\in\mathcal{O}$.
Function rules map a matching left-hand-side (LHS) pattern to a right-hand-side (RHS) template: $v_1\,F\rightarrow f_F(v_1)$ for unary postfix rules or $v_1\,F\,v_2\rightarrow f_F(v_1,v_2)$ for binary infix rules. Here, $F$ is the operator and each argument $v_k$ binds a single input token or a non-empty input string.
Interpretation of a well-formed input $x\in\mathcal{I}^*$ proceeds by recursive rule application until only terminal output tokens remain, yielding $y\in\mathcal{O}^*$. At each step, the first matching rule in the grammar's ordered list is applied.

\subsubsection{Meta-grammar}
We sample $G\sim M_{\theta}$ from the meta-grammar of ref.~\cite{lake2023human}, parameterized by $\theta=(\mathcal{I},\mathcal{O},n_p,n_f,\min_f,\max_f)$ (Extended Data Table~\ref{tab:unified_parameters}). These parameters specify the input and output vocabularies, the numbers of primitives $n_p$ and functions $n_f$ (with $n_p\leq|\mathcal{O}|$ and $n_p+n_f\leq|\mathcal{I}|$), and the allowed RHS reference count $|\mathrm{RHS}|\in[\min_f,\max_f]$. The released RHS range is $[2,8]$.

Primitives are constructed by pairing distinct $i \in \mathcal{I}$ with distinct $o \in  \mathcal{O}$ without replacement.
Each function is sampled in two steps (Supplementary Sections~\ref{sec:B_meta_grammar} and~\ref{secD}). First, the LHS is sampled by choosing the function to take one argument (unary) or two (binary) with equal probability. Each argument independently uses an atom variable $u$, binding a single token, or a string variable $x$, binding a non-empty string, also with equal probability. These choices yield the unary forms $u \hspace{0.4em} F$ and $x \hspace{0.4em} F$ and the binary forms $u \hspace{0.4em} F \hspace{0.4em} u$, $u \hspace{0.4em} F \hspace{0.4em} x$, $x \hspace{0.4em} F \hspace{0.4em} u$ and $x \hspace{0.4em} F\hspace{0.4em} x$. Second, the RHS is sampled starting with one reference to each argument to enforce non-erasure. Randomly selected argument references are added until $\min_f$ is reached, continuing with sampling each additional reference with probability $0.4$, up to the maximum $\max_f$. The references on the RHS are then shuffled.
This construction yields bounded repetitions for unary operations and combinations of repetitions, permutation and concatenation for binary operations.
A fixed final production rule $u_1\,x_1 \rightarrow [u_1]\,[x_1]$ is appended to every sampled $G$, implementing left-to-right iconic concatenation so that outputs of adjacent constituents are simply concatenated left to right.
\subsubsection{Constrained episode generation} 
Grammar episodes are generated by sampling an interpretation grammar $G$ from the meta-grammar, and then subsequently sampling unique, well-formed examples $(x,y)$ from $G$ under sequence-level constraints (Supplementary Section \ref{secD}). The released builder of ref.~\cite{lake2023human} bounds input and output lengths by $|x|, |y| \leq 8$ and draws one pool of $n_s + n_q$ examples per grammar, which is partitioned at random into support and query sets (Extended Data Table \ref{tab:unified_parameters}). For controlled evaluations, we extended this procedure to sample support and queries separately under constraint configurations $C_s$ and $C_q$. The support set is sampled first and every query input must differ from every support input. 

Both sets share grammar parameters $\theta$, but can differ in their sequence-level constraints, permitting controlled distribution shifts between support and query. 
Each configuration, $C_s$ and $C_q$, specifies lower and upper bounds on input length $|x|$, output length $|y|$, and nesting degree $d$. The nesting degree $d$ is the depth of the application tree  used to interpret $x$. Here, every applied rule adds one level, including primitive lookups and concatenation, so a bare primitive has $d=1$ and $d \leq |x|$. Where required, we additionally constrain the number of function applications $a$, concatenations $c$, or distinct function symbols. Grammar parameters can additionally restrict grammars by supplementing unary $n_u$ or binary $n_b$ rule counts or admissible LHS forms $\mathcal{T}$.
Unless stated otherwise, generated episodes use the released defaults $n_s = 14$, $n_q = 10$, $|x|, |y| \in [1, 8]$ and $d \in [0, 8]$ (Extended Data Table \ref{tab:unified_parameters}). 

We adopt a search-based procedural content generation strategy that constructs candidate inputs by recursively expanding non-terminals under $G$, guided by the input-length and nesting bounds. Each candidate is interpreted under $G$ and retained only if its output contains only terminal symbols and all specified constraints are satisfied. Duplicate inputs are discarded. Distinct accepted input--output pairs are collected until the set is complete or $1{,}000$ attempts per requested example have been exhausted. The support set must contain exactly $n_s$ items and the query set exactly $n_q$ items unless a minimum $n_q^{\min} < n_q$ is specified. If sampling fails to assemble the required sets, the episode is resampled, with up to $500$ restarts ($2{,}000$ for the $(2,7)$ and $(2,8)$ cells of the length plane (Supplementary Section~\ref{sec:C_plane}). Grammars equivalent to the human few-shot task of ref.~\cite{lake2023human} or encountered during meta-training are excluded from generated evaluations. The full episode generation procedure is summarized in Supplementary Algorithm \ref{alg:generate-episode}.

\subsubsection{Rule and item injection}
For controlled manipulations, we extended the generation procedure by i) rule injection, which replaces selected function rules or primitive mappings of a sampled grammar $G$, and ii) item injection, which constructs support and query items of specified forms following the episode's grammar instead of sampling them.
Rule injection modifies a sampled grammar $G$ before the support and query sets are sampled. An injected function rule is specified by a LHS template and a RHS that is either specified or sampled within given reference-count bounds. An eligible rule of $G$ is selected at random without replacement , using the same argument count unless a different count is specified for the condition. Replacement preserves the original operator symbol and position in the grammar's rule list. At least one sampled function rule and one primitive remain unchanged as in-episode control, except for the arrangements of two rules, where the third function rule is fixed (see Methods, ``Arrangements of two rules''). The replacement and resampling procedures are detailed in Supplementary Algorithm~\ref{alg:inject-rule}.

Support and query sets are sampled from the modified $G$ under $C_s$ and $C_q$.
Queries of a specified form are obtained either by filtering query sampling with the occurrence of the specified template or replacing the query set with constructed inputs (Supplementary Algorithm~\ref{alg:pose-queries}).
Construction replaces the sampled query set by $n_q$ constructed inputs. Each input is a specified template with the injected operators and randomly chosen primitives demonstrated in the support set, or, where specified, applications of other rules of $G$. An input is accepted if it is fully interpreted under $G$, applies the specified rules where required, and differs from every support input and every other constructed input. For every query, we record whether an injected rule is applied in its derivation.

To withhold the bare demonstration of an injected primitive, we sample $n_s + 1$ support items and $n_q - 1$ queries. The support must contain the bare item and at least two other examples containing that primitive. The bare item is moved from the support set to the query set, so that the primitive is only demonstrated in combination with function rules or concatenation. 

\subsection{Meta-learning for compositionality (MLC) framework}
\label{sec:metaLearnCompositionalityMLC}
\subsubsection{Model architectures \& inputs}
\paragraph{Meta-learning seq2seq Transformer}
The meta-learning seq2seq Transformer is reimplemented from ref.~\cite{lake2023human}. Its replication against the released model is documented in Supplementary Section \ref{secA}. The model follows the architecture of a standard seq2seq encoder--decoder Transformer with multi-head self-attention in the encoder, masked self-attention in the decoder, and encoder--decoder cross-attention. Since the support set is not natively included in the Transformer architecture, a concatenation of query input and the complete set of support input--output pairs (\emph{with context}) is passed through the encoder network, enabling each token in this input sequence to attend to all other tokens via the Transformer's self-attention mechanism. We use explicit delimiters to mark different examples (``$|$'') and the input--output boundary (``$\rightarrow$'') for support pairs. 
Positions are encoded by fixed sinusoidal embeddings. The default configuration uses $3$ encoder/decoder layers, model dimension $128$, feed-forward multiplier $4$, GELU activations, dropout $0.1$, and $8$ attention heads (Extended Data Table \ref{tab:parameters_model}). 
\paragraph{Meta-learning seq2seq LSTM}
The meta-learning seq2seq LSTM is based on ref.~\cite{lake2019compositional}, replicated within a unified framework that matches the Transformer training setup in ref.~\cite{lake2023human} (Supplementary Section \ref{secA}).
The model is composed of three biLSTM encoders (support-input, support-output, and query) with encoders for support input and query input tied 
and a single attention-based RNN output decoder. A soft key--value memory uses the final hidden states of the support-input encoder as keys, and those of the support-output encoder as values. The query encoder attends to this memory conditioned on the query encoding. By default, the query encoder receives only the query input string (\emph{without context}) since the support set is represented in the separate support encoders. For the benchmark, and out-of-training evaluations, the query encoder instead receives the query concatenated with all support input--output pairs
(\emph{with context}) in addition to its key--value memory (Supplementary Section~\ref{secA}). 
The default architecture uses $2$ encoder layers, a $2$ layer decoder, hidden size $200$, and dropout $0.5$ (Extended Data Table \ref{tab:parameters_model}).
\subsubsection{Meta-training procedure}
We train on episodic tasks sampled from the Grammar meta-distribution
in batches of $25$ episodes following the meta-training procedure in ref.~\cite{lake2023human}. Each episode comprises a support set of size $n_s$ and a query set of size $n_q$. By default $n_s=14$ and $n_q=10$, yielding an effective sequence-level batch size of $(n_q+n_s)\times25=600$. We optimize the token-level cross-entropy with AdamW ($\beta_1=0.9$, $\beta_2=0.95$, weight decay $=0.01$). Models are trained for $25$ epochs, unless otherwise stated. The default Transformer with context replication was trained for $50$ epochs following ref.~\citep{lake2023human}. The learning rate is warmed up linearly to $0.001$ during the first epoch and then decayed linearly to $5\times10^{-5}$. A model-specific dropout (for the Transformer $p=0.1$ and for the LSTM $p=0.5$) is applied to the input embeddings and models.
We include an auxiliary support loss to train the models to reproduce support outputs alongside query outputs. Gradients for the LSTM are clipped to an $\ell_2$ norm of $50$. We select checkpoints by best token-level loss on the $200$ released validation episodes, evaluated every $100$ steps. No checkpoint is saved during the first $20\%$ of epochs. The same validation episodes provide the benchmark accuracy underlying Level~1 evidence. Each training configuration was run once.

\subsubsection{Evaluation procedure}
We evaluated five checkpoints: our Transformer and LSTM replications and the three checkpoints released with ref.~\cite{lake2023human}, \emph{MLC(algebraic+biases)}, \emph{MLC(algebraic only)} and \emph{MLC(joint)} (see Methods, ``Datasets''). Unless stated otherwise, reported results refer to our two checkpoints, with the results for the released checkpoints from ref.~\cite{lake2023human} in the Supplementary. At test time, models are evaluated without weight updates. Each query is decoded separately given the support set, without access to its responses to other queries of an episode. Training and validation curves report negative log-likelihood per token.

We report exact-match accuracy of greedy decoding with a fixed output-length cap that accommodates every target in the respective evaluation ($12$ tokens unless stated otherwise). Accuracy is reported separately for \emph{retrieval} (support items) and \emph{novel} (query items). For retrieval, each support input is presented as a query together with the unchanged support set, so that its target output is contained in the context. We report $95\%$ percentile bootstrap confidence intervals from $10{,}000$ resamples of episodes to measure variation across evaluation episodes. Unless stated otherwise, the released MLC Grammar validation datasets in ref.~\cite{lake2023human} are used for scoring. Generated episodes are sampled once per evaluation and scored by all checkpoints on the identical items. 
\subsubsection{Datasets}
\paragraph{MLC Grammar} 
The MLC Grammar datasets \citep{lake2023human} instantiate the meta-learning seq2seq grammar task (see Methods, ``Meta-learning seq2seq grammar tasks''). We use the $100{,}000$ training episodes and $200$ validation episodes provided in ref.~\cite{lake2023human}. Episodes use a fixed meta-grammar and sequence-length bounds $|x|, |y| \in [1, 8]$, with no explicit constraints on nesting degree $d$ (Extended Data Table \ref{tab:unified_parameters}). To assess grammar-level novelty, we compare each validation grammar's three function-rule forms as unordered and ordered combinations with those of the meta-training, with primitive tokens and operator symbols factored out (Supplementary Section~\ref{sec:B_grammar_structure}).

Three MLC Grammar protocols were introduced in ref.~\cite{lake2023human}, that differ in how query targets are altered during meta-training. In \emph{MLC(algebraic only)}, each query target is the grammar-consistent output. In \emph{MLC(algebraic+biases)}, a fraction of query targets ($p=0.2$) was replaced by heuristic, ``human-like'' outputs (e.g., one-to-one translations, misapplied rules). In both protocols, each query target token is additionally replaced at random with a probability $p=0.01$ during training. For the third released checkpoint, \emph{MLC(joint)}, \emph{algebraic+biases} episodes with fixed and variable support-set sizes are mixed with the human open-ended responses collected in ref.~\cite{lake2023human}. Validation is always \emph{algebraic only}, so all input--output sequences are grammar-consistent. 
\paragraph{SCAN}
SCAN \citep{lake2018generalization} is an instruction-following benchmark that evaluates systematic compositional generalization abilities by requiring models to 
translate natural-language-like commands (e.g., ``jump twice and walk right'') into action sequences (e.g., ``JUMP JUMP RTURN WALK''). As in our grammar tasks, commands combine primitives (e.g., walk, run, jump), functional modifiers (e.g., twice, around right), and conjunctions that concatenate in input order (and) or in reverse order (after).

We use the \emph{add primitive} split as a replication control for our meta-learning seq2seq LSTM. Training follows the original protocol of ref.~\cite{lake2019compositional}, with one episode per batch, $20$ support items and no warm-up, except that we use linear learning-rate schedule. We compare our replication with a model trained using the original code (Supplementary Section \ref{secA}).

\subsection{Episode-level re-evaluation protocol} 
We re-evaluated all five checkpoints \emph{with context} on the $200$ validation episodes at the level of individual episodes (Supplementary Section~\ref{sec:supp_episode_qualitative}). 
To assess decoding variability, we sampled $10$ independent outputs per query from the model's output distribution using three seeds. For each episode, we compute exact-match accuracy across all queries and sampled outputs, then report the mean and standard deviation across seeds. We use the mean episode accuracies to identify low-performing episodes for qualitative analysis (Supplementary Section~\ref{sec:supp_episode_qualitative}).

\subsection{Ablation methods}
\subsubsection{Meta-training ablations}
We conducted controlled ablations on the meta-training procedure for both models trained on \emph{MLC(algebraic+biases)}, keeping their architectures at the default settings.
\paragraph{Support set representation} 
To assess the sensitivity of how episodic evidence is presented, we train and evaluate each architecture varying (i) auxiliary support loss (enabled/disabled) trained with $15$ support items, (ii) query input serialization, i.e., \emph{input with context}, for which the query is concatenated with all support pairs and \emph{input without context}, for which only the query is provided. All variants were scored on the $200$ released validation episodes.  
\paragraph{Support set size}We varied the number of support items at training time $n_s \in \{5, 10, 15, 20, 25\}$, using the Transformer with and the LSTM without context, and fixed $n_q = 10$ throughout. For each training size, we generated $100{,}000$ training episodes using the default grammar and sequence-level parameters (Extended Data Table \ref{tab:unified_parameters}). Each model was evaluated on $500$ generated episodes per test support size $n_s\in\{5,10,\ldots,40\}$. Our reported checkpoints and the released checkpoints were additionally evaluated at the same test sizes and the default $n_s=14$ (Supplementary Section~\ref{secA}).
\subsubsection{Model ablations}  
We trained one run per architecture setting on \emph{MLC(algebraic+biases)} and recorded training and validation loss logged every $100$ steps over the first $100{,}000$ steps, the length of a $25$-epoch run. For the shared parameter evaluations, we additionally measured exact-match accuracy on novel queries and support items from the same $200$ validation episodes.
\paragraph{Shared parameters (LSTM \& Transformer)}We varied (i) the \emph{hidden size} $\in \{16, 32, 64, 128, 256\}$ at two layers and dropout $0.5$ for the LSTM and three layers and dropout $0.1$ for the Transformer, (ii) the \emph{number of layers} $\in \{1, 2, 3, 4\}$ of encoder and decoder alike at hidden size $256$ and dropout $0.5$ for the LSTM and hidden size $128$ and dropout $0.1$ for the Transformer, and (iii) \emph{dropout} $\in \{0.1, 0.3, 0.5\}$ at three layers and hidden size $256$ for the LSTM and two layers and hidden size $128$ for the Transformer. The complete grid over hidden sizes $\{64, 128, 256\}$, layers $\{1, \ldots, 4\}$ and the three dropout rates, and validation accuracy  of the Transformer across hidden sizes $8$ to $256$ are given in Supplementary Section~\ref{secE}.

\paragraph{Transformer-specific parameters} To examine the contribution of multi-head attention and feed-forward capacity, we conducted two separate Transformer sweeps. For the \emph{attention heads} sweep we varied the number of heads $\in \{2, 4, 8, 12\}$ at $16$ hidden units per head, so that the hidden size follows the head count as $32$, $64$, $128$, $192$, with the feed-forward multiplier fixed at $4$. For the \emph{feed-forward multiplier} sweep, we varied the multiplier $\in \{2, 4, 8, 12\}$ at four heads and hidden size $64$.

\subsection{Procedural out-of-training (OOT) evaluation}
\label{sec:procOOTEval}
We assess OOT generalization by procedurally generating episodes under controlled changes to grammar or query constraints, while the support distribution remains at default bounds except where stated below (see Methods, ``Constrained episode generation''). We evaluate all five checkpoints, using the \emph{with context} representation for both replications (Methods, ``Evaluation procedure''). Unless stated otherwise, novel-query accuracy is computed over all query items within each condition.

\subsubsection{Axes of variation}
Each evaluation configuration $C=(\theta,C_s,C_q)$ specifies the shared grammar parameters $\theta$, support constraints $C_s$ and query constraints $C_q$. We varied selected components of $C$, keeping all others at their defaults unless stated otherwise. The selected ranges were informed by the MLC Grammar dataset statistics (Supplementary Section~\ref{secB}).

Unless stated otherwise, we used $500$ episodes with ten queries per condition, using grammars not encountered during meta-training. On the input-length, output-length and fixed-input-length nesting axes, episodes were accepted with at least four queries. For the length plane, five queries were requested and at least two required. Supplementary Section~\ref{secC} reports the realized distributions of unconstrained properties and the episode and item counts.

\paragraph{Symbol inventory size} We varied the number of primitives $n_p \in \{2, \ldots, 6\}$ and functions $n_f \in \{0, \ldots, |\mathcal{I}| - n_p\}$, covering all $30$ admissible inventories with at least two primitives ($n_p \leq |\mathcal{O}|$, $n_p + n_f \leq |\mathcal{I}|$). Sequence-level constraints remained at their defaults for both support and queries (Supplementary Section~\ref{sec:C_inventory}).

\paragraph{RHS length} We varied the number of references $|\mathrm{RHS}|$ of the functions across $|\mathrm{RHS}| \in \{1, \ldots, 8\}$ by fixing $\min_f = \max_f = |\mathrm{RHS}|$, keeping input and output lengths within the meta-training bounds. Function LHS forms were restricted to $u\,F$ and $u\,F\,u$ with only the unary form at $|\mathrm{RHS}| = 1$. Queries had $|x| \in \{2, 3\}$ for $C_q$ and $|y| \leq 8$. The support set remained at $|x|, |y| \leq 8$. Accuracy was computed only over queries consisting of a single function application, for which $|y|=|\mathrm{RHS}|$ (Supplementary Section~\ref{sec:C_rhs}). 

\paragraph{Input length} We fixed the input length $|x|$ of the queries across $|x| \in \{2, \ldots, 12\}$, with $|y|\leq8$, keeping targets within the meta-training output range. Support inputs and outputs remained bounded by eight, and nesting unbounded for both sets. For $|x| \in \{11, 12\}$ the RHS window of the grammar was set to $\min_f = 2$ and $\max_f = 3$  to obtain queries within the output bound (Supplementary Section~\ref{sec:C_input}).

\paragraph{Output length} We fixed the output length $|y|$ of the queries across $|y| \in \{2, \ldots, 12\}$ by setting $\min |y| = \max |y| = |y|$ in $C_q$, with $|x| \leq 8$ for the queries, $|x|, |y| \leq 8$ for the support set and $d$ unbounded for both sets (Supplementary Section~\ref{sec:C_output}).

\paragraph{Length plane} We fixed input and output lengths jointly over $|x| \in \{1, \ldots, 11\}$ and $|y| \in \{1, \ldots, 12\}$. Both sets had $d \leq 8$, while support input and output lengths were bounded by $\max(8, |x|, |y|)$ to permit demonstrations at the shifted lengths. The full grid, including cells obtained using modified grammar or sampling settings, is reported in Supplementary Section~\ref{sec:C_plane}.

\paragraph{Nesting degree} We varied query nesting degree $d$ in two sweeps, both with $\min_f = \max_f = 2$ to keep queries within the output bound. The first varied $d\in\{2,\ldots,12\}$ with $|x|\leq d+1$ and $|y|\leq d+6$. Support input and output lengths were bounded by $\max(8,d+6)$ and support nesting by $\max(8,d)$. The second fixed query input length at $|x|\in\{6,7,8\}$ and varied $d\in\{2,\ldots,|x|\}$ with $|y|\leq8$. Here, support lengths remained bounded by eight and support nesting was unbounded (Supplementary Section~\ref{sec:C_nesting}).

\paragraph{Primitive-only control} We constructed queries from primitives shown in the support, keeping the default grammar and support settings. Queries repeated one primitive ($|x|\in\{2,\ldots,10\}$) or combined several, with at least one repetition ($|x|\in\{3,\ldots,10\}$). Targets were the corresponding colours in input order, so $|y|=|x|$. Repeated queries and matches to support inputs were allowed (Supplementary Section~\ref{sec:C_primitives}).

\paragraph{Within-grammar verification} We sampled $500$ unseen grammars at the default $\theta$ and fixed one $14$-item support set per grammar at the default bounds. Using the same grammar and support set across lengths and checkpoints, query sets were sampled using $n_q=10$ at each exact input length length $|x|\in\{2,\ldots,12\}$, with $|y|\leq12$ and $d$ unbounded. A grammar was omitted at a length if fewer than ten queries could be generated. For a matched comparison, we retained $293$ grammars, with at least one function-application query with $|y|\leq8$ at every input length from three to eight, and scored only those query subsets (Supplementary Section~\ref{sec:C_within}).

\paragraph{End-of-sequence probabilities} We fixed support set at $|x|, |y| \leq 8$, $|\mathrm{RHS}| \in [2, 6]$ and $d \leq 6$, and raised the query output bound to $14$ tokens. For each query, we supplied the correct output prefix at each position and recorded each model's probability of emitting the end-of-sequence token (Supplementary Section~\ref{sec:C_window}).

\subsection{OOT evaluation with injected rules}
Beyond the controlled axes of the OOT evaluation, we generated episodes with injected rules (see Methods, ``Rule and item injection''). For the argument-width and single-manipulation conditions, support sets and filtered queries were sampled at the released bounds, whereas constructed queries are constrained by their template with $d$ unbounded. For each condition, we generated $500$ episodes, or $1{,}000$ where a primitive value is held out. For filtered query sets, each episode was sampled with $n_q=10$ and accepted with at least $n_q^{\min}=1$. Constructed query sets contained ten items, with episodes excluded that do not demonstrate the injected operator. Novel accuracy is computed over queries that apply the injected rule, and retrieval accuracy over the support items that apply an injected rule. For held-out primitive conditions, novel accuracy is computed over the bare query of that primitive and retrieval accuracy over the support items that contain the primitive. Supplementary Section~\ref{secG} reports the full battery of injected conditions for the five checkpoints.

\subsubsection{Argument widths} We injected a reversing binary function $x_1\,F\,x_2 \rightarrow [x_2]\,[x_1]$ in place of a sampled binary function rule and constructed the queries $u_1\ldots u_{w_1}\,F\,u_{w_1+1}\ldots u_{w_1+w_2}$ for $w_1,w_2\geq1$ and $w_1+w_2\leq8$. The left and right arguments contain $w_1$ and $w_2$ tokens, respectively, so that $|x|$ ranges from three to nine tokens. All width pairs share the same grammars and support sets, drawn with the same seed and only the constructed queries differ (Supplementary Section ~\ref{sec:C_widths}).

\subsubsection{Arrangements of two rules} We injected a reversing function $f$, $x_1\,f\,x_2 \rightarrow [x_2]\,[x_1]$ and an order-preserving function $g$ with $x_1\,g\,x_2 \rightarrow [x_1]\,[x_2]$. The third function was fixed to the unary rule $u_1\,h \rightarrow [u_1]\,[u_1]$. Each grammar was paired with a version in which the positions of rules $f$ and $g$ in the rule list were exchanged, changing the bracketing in ambiguous cases. Depending on rule order, a query with three function operators may be derived either as a chain of nested applications or a branching derivation in which the middle operator combines two arguments that each contain one application.

We compared support demonstrating $fg$, $gf$, or both, always including all primitive mappings, single applications of the three functions, and four examples in which $f$ or $g$ takes an argument of more than one token ($a\,h\,f\,c$, $c\,f\,a\,h$, $a\,h\,g\,c$ and $c\,g\,a\,h$). Support size and total input--output token count were matched across conditions ($14$ examples, $72$ tokens). Within each condition, paired episodes used identical support inputs and presentation order, with targets interpreted under the corresponding grammar. For each of the $3{,}000$ supports, we enumerated all grammars of the meta-grammar that match its input--output examples and checked that the query targets are unambiguously determined by the support.

For each of the $14$ arrangements with one to three function applications, ten query inputs were constructed, including both chain and branching derivations (Supplementary Section~\ref{sec:G_arrangements}). Query inputs were identical across rule lists, and neither inputs nor targets appeared in support. Support demonstrating only $fg$ or only $gf$ was tested on the $fg$ and $gf$ queries only. For $fgf$ with $f$ earlier and $gfg$ with $g$ earlier, we additionally posed the inner expression and the outer application as separate queries, supplying the correct inner output as primitive words in the latter. We computed confidence intervals by resampling grammar pairs.

Full results and response classifications are reported in Supplementary Section~\ref{sec:G_arrangements}.

\subsubsection{Controlled manipulations} 
We manipulated one property of a single injected function rule or primitive. As controls, we injected a unary function $u_1\,F$ and a binary function $u_1\,F\,u_2$, each with two to four RHS references, and a primitive mapping to one output token. RHS not specified are sampled with two to four references, as for the controls. Queries are filtered for the injected form, except for the held-out values, whose bare query is posed (Supplementary Section~\ref{sec:G_overview}).
\paragraph{Seven-reference pairings} We tested combinations of individually attested rule components without violating the meta-grammar, pairing the binary LHS $u_1\,F\,u_2$ with seven-reference RHS patterns. Seven is the shortest RHS length at which some admissible pairings are absent from the meta-training grammars, while remaining within the eight-token meta-training bound. Of the $126$ non-erasing RHS patterns, $19$ never occurred with the atomic-only binary LHS, although each occurred with another binary LHS. We injected each of the $19$ absent pairings as one condition and, as controls, $19$ of the $107$ attested pairings drawn at random. Accuracy is reported over the queries of all conditions of a group. Supplementary Section~\ref{sec:G_rhs7} reports every pattern. 
\paragraph{Number of arguments} We injected a ternary function $u_1\,F\,u_2\,u_3$ in place of a binary rule, whereas the meta-grammar only admits unary and binary function rules. To ensure non-erasure, its RHS has at least three references.
\paragraph{Operator position} We injected functions with the operator in first position, $F\,u_1$ and $F\,u_1\,u_2$, or in last position, $u_1\,u_2\,F$, whereas the meta-grammar places it after the first argument.
\paragraph{References} We set $|\mathrm{RHS}|=1$, below the released range of two to eight references. For the injected unary function rule, $u_1\,F \rightarrow [u_1]$, this corresponds to the identity mapping, which is attested in the meta-grammar for binary functions only. For the injected binary function rule, $u_1\,F\,u_2 \rightarrow [u_1]$, this additionally erases its second argument, breaking the non-erasure requirement of the meta-grammar.
\paragraph{Non-erasure} We injected $u_1\,F\,u_2\rightarrow[u_1]\,[u_1]$, erasing the second argument while retaining two RHS references. With two references, the RHS stays within the window of the meta-grammar, so that, unlike the binary function with a single reference above, only the non-erasure requirement of the meta-grammar is violated. 
\paragraph{Primitive value} We injected a primitive with a two-token value and withheld its bare demonstration. At least two support items contained the primitive within function applications or concatenations, and only its bare input was scored as a novel query. Widths of one to twelve tokens are reported as the $|x| = 1$ column of the full length plane in Supplementary Section~\ref{sec:C_plane}. The example in Fig.~\ref{fig:rule_forms}b uses the episode of Fig.~\ref{fig:overview} with the injected value $\mathtt{dax}\rightarrow\mathtt{BLUE\ YELLOW}$. The withheld bare $\mathtt{dax}$ example is replaced by $\mathtt{wif\ kiki}$ to keep $14$ support examples.
\paragraph{Responses to bare queries} For each rule condition, we examined the \emph{bare} queries, in which the injected rule is applied once to primitives, such as $a\,F\,b$. We used only queries with non-identical primitives ($643$--$2{,}014$ per condition), so that each output token can be assigned to one argument. For each rule, we report the most frequent incorrect response with the share of model predictions matching it (Supplementary Table~\ref{tab:supp_predicted_answers}).

\section*{Data availability}
The evaluation data and model checkpoints are available on \href{https://zenodo.org/records/23917882?preview=1&token=eyJhbGciOiJIUzUxMiJ9.eyJpZCI6IjUzZjkwZWIxLWVmMjAtNGQ4YS1iYjEzLTZiODE4YWM2ZDEwZiIsImRhdGEiOnt9LCJyYW5kb20iOiJlMTkzNjI1NTJiOTZiOGU2NjFjZDZkMGY5MDBmNjhmNCJ9.nxEBKX9RNo3Xsxgi43ttSTuUj5vDGfv7g6H9LipVD392JcJSOzcwqV3WHje5twhShqL6KKAy7j5crkSDrFib0w}{Zenodo}. Download links for the MLC data and pretrained models from ref.~\cite{lake2023human} are provided at \url{https://github.com/brendenlake/MLC}. The SCAN  dataset~\citep{lake2018generalization} used for the replication
analyses is available at \url{https://github.com/brendenlake/SCAN}.

\section*{Code availability}
The code used for training, evaluation and analysis is available at
\url{https://github.com/RothkopfLab/meta-learning-limits}.

\bibliography{sn-bibliography}

\section*{Acknowledgements}
We would like to thank Dorothea Ballmann and Michelle Geisler for discussions. 

\section*{Author contributions}

\section*{Funding}
K.K. and C.A.R. acknowledge the German Research Foundation under Germany’s Excellence Strategy (EXC 3066/1 “The Adaptive Mind”) and (EXC 3057/1 "Reasonable Artificial Intelligence"). 
\section*{Competing interests}
The authors declare no competing interests.

\clearpage
\setcounter{figure}{0}\renewcommand{\figurename}{Extended Data Fig.}
\setcounter{table}{0}\renewcommand{\tablename}{Extended Data Table}
\renewcommand{\theHfigure}{ED.\arabic{figure}}\renewcommand{\theHtable}{ED.\arabic{table}}

\begin{figure}[!htbp]
    \centering
    \includegraphics[width=\linewidth]{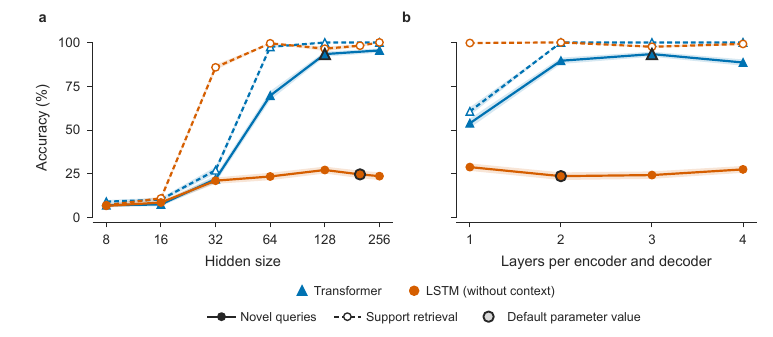}
    \caption{\textbf{Validation accuracy across model width and depth.}
    \textbf{a,} Hidden size at three layers per encoder and decoder
    for the Transformer (blue) and two for the LSTM (orange).
    \textbf{b,} Depth at hidden size $128$ for the Transformer
    and $256$ for the LSTM.
    The Transformer receives serialized support context; the LSTM
    accesses its key--value memory without serialized context.
    Dropout is $0.1$ for the Transformer and $0.5$ for the LSTM.
    All checkpoints are evaluated on the same $200$ released
    validation episodes. Filled markers and solid lines show
    novel-query accuracy and open markers and dashed lines
    support retrieval. Bands show $95\%$ episode-clustered bootstrap
    confidence intervals. Black outlines mark default
    architectural parameter values.}
    \label{fig:supp_hidden_size_accuracy}
\end{figure}

\begin{sidewaystable}[!htbp]
\caption{Parameters of episode generation}
\label{tab:unified_parameters}
\begin{tabular*}{\textwidth}{@{\extracolsep\fill} l p{0.5\textwidth} c}
\toprule
Parameter & Description & Default \\
\midrule
\multicolumn{3}{@{}l}{\textit{Meta-grammar parameters $\theta$}}\\
$\mathcal{I}$ & Input symbols (pseudowords). & \emph{dax, lug, wif, zup, fep, blicket, kiki, tufa, gazzer} \\
$\mathcal{O}$ & Output symbols (colours). & \colordot{myblue}, \colordot{myyellow}, \colordot{mygreen}, \colordot{myred}, \colordot{mypink}, \colordot{mypurple} \\
$n_p$ & Number of primitives. & 4 \\
$n_f$ & Number of functions. & 3 \\
$[\min_f, \max_f]$ & Bounds on the number of argument references $|\mathrm{RHS}|$ of a function. & [2, 8] \\
$p_{\mathrm{onearg}}$ & Probability that a function takes one argument. & 0.5 (fixed) \\
$p_{\mathrm{stop}}$ & Probability of ending the right-hand side before each further reference. & 0.6 (fixed) \\
$n_u$, $n_b$ & Numbers of unary and binary functions among the $n_f$\footnotemark[1]. & unconstrained \\
$\mathcal{T}$ & Admissible left-hand-side forms, with $F$ the operator symbol, $u$ an atom variable and $x$ a string variable: $u\,F$, $x\,F$, $u\,F\,u$, $u\,F\,x$, $x\,F\,u$, $x\,F\,x$. Each function's form is drawn from $\mathcal{T}$\footnotemark[1]. & all six \\
\addlinespace
\multicolumn{3}{@{}l}{\textit{Episode parameters}}\\
$n_s$ & Number of support items. & 14 \\
$n_q$ & Number of query items. & 10 \\
$n_q^{\min}$ & Minimum number of query items accepted\footnotemark[1]. & $n_q$ \\
\addlinespace
\multicolumn{3}{@{}l}{\textit{Sequence-level constraints, per configuration $C_s$ and $C_q$}}\\
$|x|$ & Bounds on the number of input tokens. & [1, 8] \\
$|y|$ & Bounds on the number of output tokens. & [1, 8] \\
$d$ & Bounds on the nesting degree, the depth of the application tree; unconstrained in the released procedure, and inert for $|x| \leq 8$ since $d \leq |x|$. & [0, 8] \\
$a$ & Number of function applications in $x$, counted as operator tokens\footnotemark[1]. & unconstrained \\
$c$ & Number of concatenation applications in $x$\footnotemark[1]. & unconstrained \\
$k$, $k_u$, $k_b$ & Numbers of distinct function symbols, unary function symbols and binary function symbols in $x$\footnotemark[1]. & unconstrained \\
\botrule
\end{tabular*}
\footnotetext{Parameters of the grammar sampler, the episode and the example sampler. Defaults are those of ref.~\cite{lake2023human} and every generated episode uses them unless the evaluation states otherwise.}
\footnotetext[1]{Not available in the released procedure; used only by the evaluations that state it.}
\end{sidewaystable}

\begin{sidewaystable}[!htbp]
\centering
\caption{Hyperparameters of the sequence-to-sequence architectures and their defaults}
\label{tab:parameters_model}
\begin{tabular*}{\textwidth}{@{\extracolsep\fill} l p{0.45\textwidth} c c}
\toprule
\textbf{Parameter} & \textbf{Description} & \textbf{LSTM} & \textbf{Transformer}\\
\midrule
\multicolumn{4}{@{}l}{\textit{Architecture}}\\
Number of layers (encoder) & Stacked layers in the encoder. & 2 & 3 \\
Number of layers (decoder) & Stacked layers in the decoder. & 2 & 3 \\
Hidden size & Dimensionality of token embeddings and model states. & 200 & 128 \\
Dropout & Probability applied to embeddings and intermediate activations. & 0.5 & 0.1 \\
Activation & Activation of the feed-forward network. & & GELU \\
Number of heads & Attention heads in multi-head self-attention. & & 8 \\
Feed-forward multiplier & Feed-forward width relative to the hidden size. & & 4 \\
Query input & Input to the query encoder\footnotemark[1]. & without context & with context \\
\addlinespace
\multicolumn{4}{@{}l}{\textit{Meta-training}}\\
Epochs & Passes over the $100{,}000$ training episodes. & 25 & 50 \\
Gradient clipping & Maximum $\ell_2$ norm of the gradient. & 50 & none \\
Support loss & Support outputs included in the training loss. & yes & yes \\
\botrule
\end{tabular*}
\footnotetext{Defaults of the two reported models. For the benchmark and out-of-training evaluations, the LSTM \emph{with} context is used. An empty cell means the parameter does not apply to that architecture. Batch size ($25$ episodes), optimiser and learning-rate schedule are shared (see Methods, ``Meta-training procedure'').}
\footnotetext[1]{\emph{With context}: the query concatenated with all support input--output pairs; \emph{without context}: the query alone. The LSTM reads the support set through its memory in both cases.}
\end{sidewaystable}

\clearpage
\renewcommand{\figurename}{Fig.}
\renewcommand{\tablename}{Table}
\renewcommand{\theHfigure}{SI.\thefigure}\renewcommand{\theHtable}{SI.\thetable}
\begin{appendices}
\input{sections/supplementary_material}
\newpage
\end{appendices}
\clearpage
\end{document}

%% file: sections/supplementary_material.tex
\section{Replication of meta-learning neural networks} \label{secA}
We re-implemented the meta-learning seq2seq LSTM and Transformer architectures in a unified training and evaluation framework, so that the two models can be compared in terms of their architectural differences rather than their training procedures. The replication (i) reproduces the published behaviour of each model on its native benchmark and (ii) provides a common baseline for the subsequent analyses. 

\subsection{Alignment of the training protocols} 
The LSTM of ref.~\cite{lake2019compositional} was developed for SCAN, whereas the Transformer of ref.~\cite{lake2023human} was developed for MLC Grammar. We used MLC Grammar for its parametric control over structure and complexity and aligned the LSTM's training protocol with the Transformer's. This changed the number of primitives from $20$ to $4$, batch size from $1$ to $25$, and support size from $20$ to $14$, and introduced epoch-based training with linear learning-rate decay. On MLC Grammar, we trained the Transformer for $50$ epochs of $4{,}000$ steps each and the LSTM for $25$ epochs, where learning curves had converged. We selected each by its lowest validation loss (see Methods, ``Meta-training procedure'').

\begin{figure}[!htbp]
    \centering
    \includegraphics[width=1\textwidth]{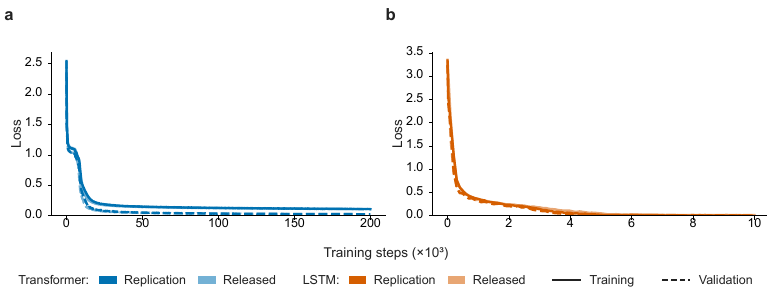}
    \caption{\textbf{Replication of the reported training dynamics.} Mean token cross-entropy on training (solid) and validation (dashed) episodes for our implementations (dark) and comparison runs (light). \textbf{a,} Transformer on MLC Grammar, compared with the released run: $50$ epochs of $4{,}000$ steps. \textbf{b,} LSTM on SCAN, compared with a model trained here using the original code of ref.~\cite{lake2019compositional}.}
    \label{fig:supp_replication_loss_curves}
\end{figure}

To check the LSTM implementation on SCAN, we trained it for $10{,}000$ episodes with one episode per batch, $20$ support and query items, and no warm-up. We followed the original protocol except for linear learning-rate decay in place of mid-training halving. As no checkpoint was released, we compared against a model trained with the original code of ref.~\cite{lake2019compositional}. The learning curves closely followed this reference (Fig.~\ref{fig:supp_replication_loss_curves}b).

Default architectural parameters follow prior work (Extended Data Table~\ref{tab:parameters_model}). By default, the LSTM query encoder receives the query alone (\emph{without context}), whereas the Transformer receives the query concatenated with all support input--output pairs (\emph{with context}), separated by the delimiter token ``$|$''. The LSTM retains access to its key--value memory in both configurations. Our reported LSTM was trained and evaluated with context, including in the OOT analyses and the test-support-size evaluation below. The architecture grids and the experiments varying support size during training retain the default configurations: Transformer with context and LSTM without context (see Methods, ``Ablation methods'').

\subsection{Correctness of the replication} 
\begin{figure}[!htbp]
    \centering
    \includegraphics[width=1\textwidth]{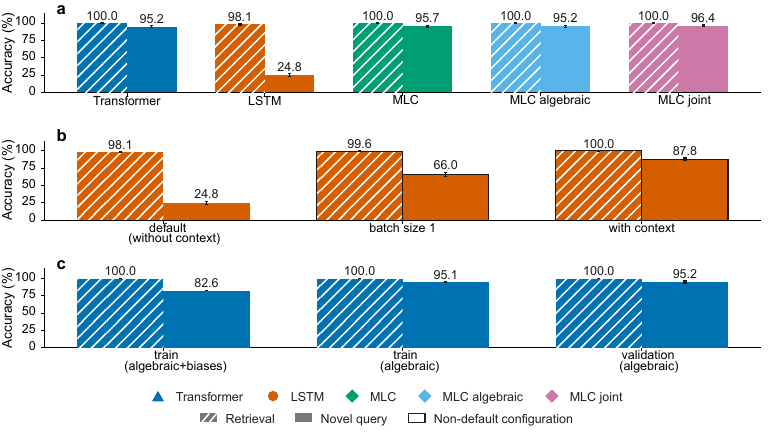}
    \caption{\textbf{Baseline accuracies and the effects of training protocol and target definition.} Bars show exact-match accuracy (\%) for support retrieval (hatched) and novel queries (unhatched), with $95\%$ episode-clustered bootstrap intervals. \textbf{a,} Our Transformer with context, LSTM without context and three released MLC checkpoints on the $200$ validation episodes. \textbf{b,} LSTM accuracy on the same episodes without context at batch sizes $25$ (left) and $1$ (middle), and with context at batch size $25$ (right). Black outlines mark the two alternative configurations. \textbf{c,} The fixed Transformer on $100{,}000$ training episodes with algebraic+biases training episodes (left), algebraic training episodes (middle), validation episodes (right). Support and validation targets are algebraic.}
\label{fig:supp_rep_anomalies}
\end{figure}
Both re-implementations reproduce the reference training dynamics on their native datasets (Fig.~\ref{fig:supp_replication_loss_curves}). Our Transformer reaches $95.2\%$ novel-query accuracy on the validation episodes, compared with $95.2$--$96.4\%$ for the released MLC checkpoints (Fig.~\ref{fig:supp_rep_anomalies}a). The LSTM without context reaches $24.8\%$. Reducing its batch size from $25$ to the original $1$ raises accuracy to $66\%$, but the LSTM with context performs better at $87.8\%$ (Fig.~\ref{fig:supp_rep_anomalies}b). We therefore use the LSTM with context for all OOT analyses.

For both Transformer runs trained on MLC(algebraic+biases), validation loss is lower than training loss (Fig.~\ref{fig:supp_replication_loss_curves}a). Training targets include heuristic substitutions ($p = 0.2$) and perturbations ($p = 0.01$), whereas validation targets are algebraic (see Methods, ``Datasets''). Resampling these alterations for the $1{,}000{,}000$ training queries changes $133{,}815$ targets ($13.4\%$). On these queries, the fixed Transformer checkpoint produces the algebraic output in $93.3\%$ of cases and matches the
altered target in only $328$ cases. Its accuracy across all training queries is $82.6\%$ against the altered targets and $95.1\%$ against the algebraic targets,
compared with $95.2\%$ on the algebraic validation queries
(Fig.~\ref{fig:supp_rep_anomalies}c). The training--validation accuracy gap thus largely disappears when both splits are scored against algebraic targets.

\subsection{Support-set size at test time for the checkpoints} 
\begin{figure}[!htbp]
    \centering
    \includegraphics[width=\textwidth]{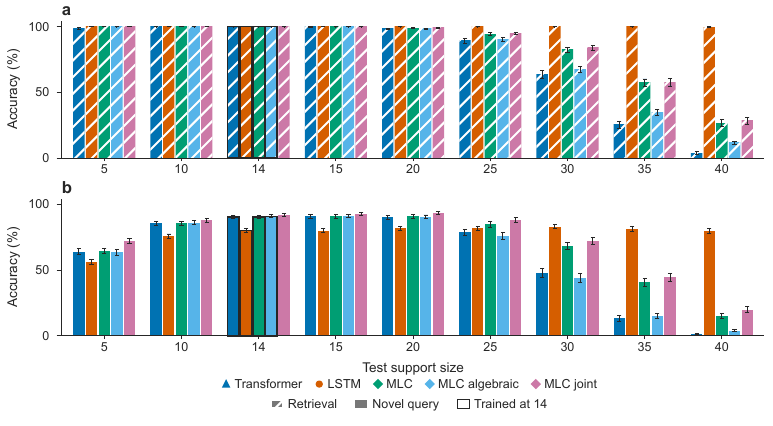}
    \caption{\textbf{Effects of test support size for the five checkpoints.}
    \textbf{a,} Retrieval accuracy.
    \textbf{b,} Novel-query accuracy.
    At each test support size, the five checkpoints are evaluated on the same
    $500$ generated episodes with ten novel queries each. Bars are grouped by
    support size. Black outlines mark
    $n_s=14$ for the four checkpoints trained with a fixed support size;
    \emph{MLC(joint)} also used variable support sizes. Error bars are $95\%$
    episode-clustered bootstrap confidence intervals.}
\label{fig:supp_nsupport_default_models}
\end{figure}

All checkpoints used a fixed training support size of $14$,
except \emph{MLC(joint)}, which used variable support
sizes (see Methods, ``Datasets''). At each test size $n_s \in \{5, 10, 14, 15, 20, \ldots, 40\}$, we evaluated all five checkpoints on the same $500$ generated episodes (Fig.~\ref{fig:supp_nsupport_default_models}).
At $14$ examples, novel-query accuracy is $90$--$92\%$ for the four Transformer checkpoints and $80\%$ for the LSTM with context. Larger support sets reduce Transformer accuracy. At $40$, our Transformer and the released MLC checkpoint reach only $2\%$ and $15\%$ novel-query accuracy, with retrieval at $4\%$ and $27\%$, respectively. The LSTM with context maintains $80$--$83\%$ novel-query accuracy and near-perfect retrieval. At five examples, novel-query accuracy falls to $56$--$72\%$ across models. This decline may reflect reduced evidence about the grammar as much as a support set size change: with fewer support items, the primitives and rules of $G$ appear less often and in fewer combinations, and a query may require an undemonstrated rule.

\clearpage
\section{MLC Grammar dataset analysis} \label{secB}
We quantified the grammar-level and item-level statistics of the released MLC Grammar episodes, over all $100{,}000$ training episodes ($2.4$ million items) and the $200$ validation episodes, to identify where the models receive most exposure and to inform the constraints of the OOT evaluation. Quantifying the training statistics relates a model's failures to the frequency of the configurations that fail, so that a failure can be read as a departure from frequent configurations rather than as a failure of rule composition alone. The analysis reveals a strong bias towards simple and shallow episodes, with narrow ranges of every generation parameter. Testing compositional generalization therefore requires extending the evaluation beyond these ranges, in a systematic OOT evaluation that respects the dependencies among the parameters. 

\subsection{Episode parameters} \label{sec:B_meta_grammar}
\begin{figure}[!htbp]
    \centering
    \includegraphics[width=1\textwidth, keepaspectratio]{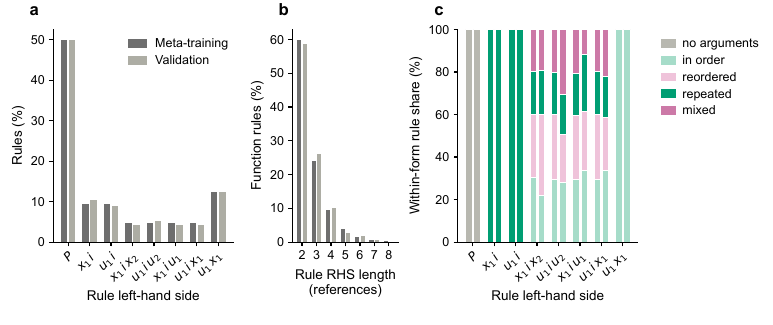}
    \caption{\textbf{Rule statistics of meta-training and validation.} Paired bars summarise $100{,}000$ meta-training grammars (left) and $200$ validation grammars (right).\textbf{a,} Share of each left-hand-side form among all rules, including primitives ($P$) and concatenation ($u_1\,x_1$).\textbf{b,} Right-hand-side (RHS) lengths in argument references, among function rules.\textbf{c,} RHS argument use by form: each argument once in order or reordered, repeated without reordering, or repeated and reordered (mixed). Primitives have no argument references.}
    \label{fig:supp_meta_grammar_stats}
\end{figure}

The fixed meta-grammar restricts the symbolic and structural
range of generated sequences through its design and
hyperparameters (Extended Data Table~\ref{tab:unified_parameters}). Each grammar contains four primitives, three function rules and the fixed concatenation rule. Function rules are equally likely to be unary or binary, and each argument is independently an atom or a string with equal probability. This gives six possible left-hand-side forms, all with the operator following the first argument (Fig.~\ref{fig:supp_meta_grammar_stats}a). The vocabularies are also small, with $9$ input symbols, of which each grammar uses seven, and $6$ output symbols. Each right-hand side (RHS) includes every argument and contains between two and eight references. Further references are added with probability $0.4$, and the sampled references are then shuffled. Short RHS dominate this distribution. Of the function rules, $60\%$ contain two references, $24\%$ contain three, $10\%$ contain four and only $6\%$ contain five to eight
(Fig.~\ref{fig:supp_meta_grammar_stats}b). The two-reference minimum means that unary rules always repeat their argument. Among binary rules, $30\%$ preserve argument order without repetition and $30\%$ reverse it. A further $20\%$ repeat at least one argument without reordering, and $20\%$ combine repetition with reordering (Fig.~\ref{fig:supp_meta_grammar_stats}c). Apart from symbol assignments, structural variation across grammars is limited to the forms of the three function rules.

\begin{figure}[!htbp]
    \centering
    \includegraphics[width=1\textwidth,keepaspectratio]{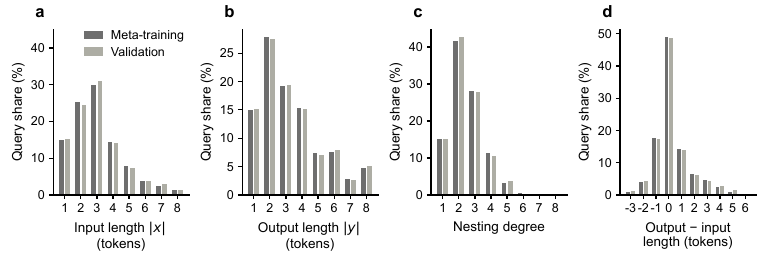}
    \caption{\textbf{Query statistics of the meta-training and validation splits.} Share of the $1{,}000{,}000$ meta-training queries (dark) and the $2{,}000$ validation queries (light) by \textbf{a,} input length, \textbf{b,} output length, \textbf{c,} nesting degree and \textbf{d,} the difference between output and input length.}
    \label{fig:supp_example_stats}
\end{figure}

The sequences generated from these grammars mirror this low complexity. Inputs and outputs are bounded at $1$--$8$ tokens (Extended Data Table~\ref{tab:unified_parameters}). Within this range the queries are short: $69\%$ of the training queries have two to four input tokens, with a peak at three, and $8\%$ have six or more; $62\%$ have two to four output tokens (Fig.~\ref{fig:supp_example_stats}a,b).
The nesting degree $d$, the depth of the application tree, with a bare primitive with $d=1$, is not constrained in the released generation. It ranges from $1$ to $8$, with $42\%$ of the queries at depth two and $4\%$ at depth five or more (Fig.~\ref{fig:supp_example_stats}c). The validation split matches all four distributions within $1.2$ percentage points at every value. 

\subsection{Interaction of episode parameters}
\begin{figure}[!htbp]
    \centering
    \includegraphics[width=1\textwidth,keepaspectratio]{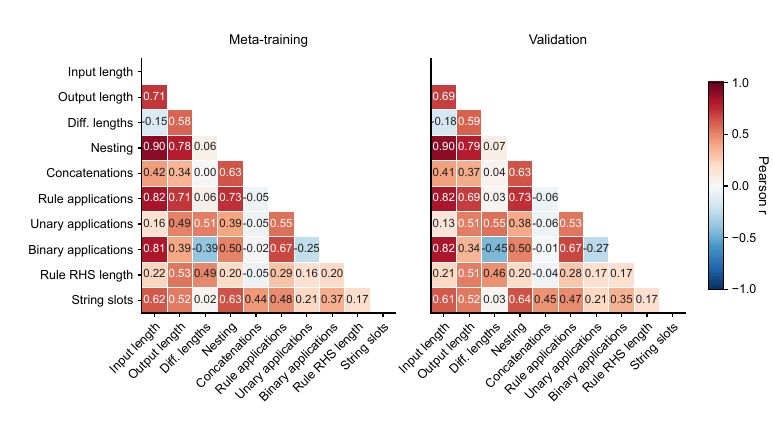}
    \caption{\textbf{Dependencies among item parameters.} Pearson correlations over the $1{,}000{,}000$ meta-training queries (left) and the $2{,}000$ validation queries (right) between selected sequence parameters. The length difference is $|y|-|x|$. RHS length and string-slot count
refer to the outermost applied rule.} 
    \label{fig:supp_correlation_data}
\end{figure}

Item parameters are not independent. Input and output lengths coincide for $49\%$ of queries, while outputs are more often longer ($29\%$) than shorter ($23\%$)  (Fig.~\ref{fig:supp_example_stats}d). Argument repetition increases output length, whereas operator tokens contribute only to input length. The strongest dependence links length and nesting degree ($r = 0.90$ with input length and $r = 0.78$ with output length). Each level of nesting requires input tokens, so input length bounds the possible depth. Input length also correlates strongly with the number of rule applications ($r = 0.82$), including binary applications ($r = 0.81$), which contribute operator and argument tokens (Fig.~\ref{fig:supp_correlation_data}).

Comparisons within the released episodes therefore do not isolate the effects of individual parameters. Our OOT evaluations vary one or two axes at a time while keeping other generation constraints at their defaults. For nesting, we report sweeps both with input length growing alongside depth and with input length held fixed (Section~\ref{sec:C_nesting}).

\subsection{Overlap between training and validation on the level of grammar structure}
\label{sec:B_grammar_structure}
Beyond marginal statistics, we asked whether validation grammars introduce new rule combinations, refining our comparison in ref.~\cite{woydt2025fodor}.
We represent each function rule by its left-hand-side pattern, with the operator symbol replaced by a placeholder, and its right-hand-side reference pattern, for example $u_1\,i\,u_2 \rightarrow [u_2]\,[u_1]$. Primitives and the fixed concatenation rule are omitted because their forms do not vary across grammars. Initially ignoring rule order, we represent each grammar by the combination of its three function-rule forms. A validation grammar is attested if this combination occurs in meta-training, a recombination if its three forms occur individually but not together, and novel if it contains an unseen form. The $100{,}000$ meta-training grammars contain $1{,}322$ distinct rule forms and $23{,}197$ combinations. Of the $200$ validation grammars, $178$ are attested, $21$ are recombinations and one is novel, containing an unseen binary form with seven references. Also accounting for rule order, which determines the interpretation of ambiguous queries, $147$ validation grammars still match a meta-training grammar.

Therefore, most validation grammars use rule combinations already seen in meta-training, even when rule order is taken into account. Repetition is also common within meta-training, where $84\%$ of grammars have the same three rule forms as at least one other grammar when order is ignored. High overall accuracy on this split therefore provides limited evidence for generalization to new rule combinations.
\newpage
\section{Episode-level evidence for non-compositional behaviour}
\label{sec:supp_episode_qualitative}

The MLC Grammar benchmark of ref.~\citep{lake2023human} is intended to test compositional generalization to novel combinations of operations inferred from an episode's support examples. Most validation grammars reuse rule-form combinations
already seen during meta-training (Section~\ref{sec:B_grammar_structure}). This limits what validation accuracy can establish about generalization to new grammar structures.

\subsection{Repeated decoding across checkpoints}\label{sec:C_repeated}

\begin{wraptable}{r}{0.54\textwidth}
\centering
\caption{\textbf{Accuracy on the $2{,}000$ validation queries.} Greedy decoding is deterministic; repeated
decoding samples ten outputs per query, as the mean $\pm$ s.d.\ over three seeds.}
\label{tab:supp_repeated_decoding}
\footnotesize
\begin{tabular*}{\linewidth}{@{\extracolsep{\fill}} l r r @{}}
\toprule
Checkpoint & Greedy & Repeated \\
\midrule
Transformer (ours) & $95.2\%$ & $83.6 \pm 0.7\%$ \\
LSTM (ours) & $87.8\%$ & $72.5 \pm 1.0\%$ \\
\emph{MLC(algebraic+biases)} & $95.7\%$ & $83.8 \pm 0.7\%$ \\
\emph{MLC(algebraic only)} & $95.2\%$ & $92.6 \pm 0.5\%$ \\
\emph{MLC(joint)} & $96.4\%$ & $84.8 \pm 0.7\%$ \\
\botrule
\end{tabular*}
\end{wraptable}

Using greedy decoding (see Methods, ``Evaluation procedure''), our replicated Transformer produces exact matches for $95.2\%$ of the $2{,}000$ validation queries. In $96$ of the $97$ errors, the prediction contains only colours from the target but orders or repeats them incorrectly. In $32$ errors, every colour count is correct and only the order is wrong.

To determine whether average performance conceals systematic episode-level errors, we re-evaluated all five checkpoints on the MLC Grammar validation episodes using repeated decodes. For each episode, we decoded every query with the full support set as context (\emph{with context}) and sampled ten outputs per query, repeating the protocol with three seeds. We calculated exact-match accuracy across all queries and all ten outputs in each episode (Tables~\ref{tab:supp_repeated_decoding} and~\ref{tab:supp_worst_episodes}).

We extend the protocol reported in ref.~\cite{woydt2025fodor} to our LSTM and Transformer replications and all three reported checkpoints in ref.~\cite{lake2023human}. Episodes~\#133 and~\#32 are among the three lowest-scoring episodes for every checkpoint, and~\#133 is the lowest for four of the five (Table~\ref{tab:supp_worst_episodes}). For complete model responses, see Section~\ref{sec:supp_extended_outputs}.

\begin{table}[!b]
\centering
\caption{\textbf{The three lowest-scoring validation episodes of each checkpoint.} Episode-level
exact-match accuracy under repeated decoding, as the mean $\pm$ s.d.\ over three seeds, ranked from the
lowest of the $200$ validation episodes.}
\label{tab:supp_worst_episodes}
\footnotesize
\begin{tabular*}{\textwidth}{@{\extracolsep{\fill}} l c c c @{}}
\toprule
Checkpoint & 1st & 2nd & 3rd \\
\midrule
Transformer (ours)
& \#133\quad $41.3 \pm 1.2\%$
& \#32\quad $47.3 \pm 0.6\%$
& \#122\quad $51.7 \pm 2.3\%$ \\
LSTM (ours)
& \#32\quad $32.0 \pm 1.0\%$
& \#133\quad $37.0 \pm 2.0\%$
& \#99\quad $40.0 \pm 4.4\%$ \\
\emph{MLC(algebraic+biases)}
& \#133\quad $40.0 \pm 1.0\%$
& \#32\quad $56.7 \pm 2.5\%$
& \#122\quad $58.3 \pm 3.5\%$ \\
\emph{MLC(algebraic only)}
& \#133\quad $42.0 \pm 1.0\%$
& \#32\quad $56.0 \pm 1.0\%$
& \#122\quad $60.3 \pm 3.1\%$ \\
\emph{MLC(joint)}
& \#133\quad $42.3 \pm 1.5\%$
& \#32\quad $46.3 \pm 1.5\%$
& \#99\quad $49.0 \pm 6.1\%$ \\
\botrule
\end{tabular*}
\end{table}

\subsection{Failure in rule extraction}

In Episode \#133 (Table~\ref{tab:ep133}), the model frequently maps the token intended as $\mathtt{twice}$ (here $\langle\mathrm{fep}\rangle$) to outputs consistent with $\mathtt{thrice}$ (here $\langle\mathrm{gazzer}\rangle$), yielding repetition counts that are systematically offset. This persists despite a support item that applies $\mathtt{twice}$, suggesting that the inferred operation semantics are not applied consistently across queries. Considering the examples $\langle\mysquare[myblue] \rightarrow \mycirc[myblue]\rangle$, $\langle\mysquare[myred] \rightarrow \mycirc[myred]\rangle$, $\langle\mysquare[myblue]\ \mathtt{thrice} \rightarrow \mycirc[myblue] \mycirc[myblue] \mycirc[myblue]\rangle$, a systematic learner would infer that $\langle\mathtt{twice}\rangle$ and $\langle\mathtt{thrice}\rangle$ carry different semantics.

\begin{table*}[!hb]\footnotesize
\centering
\caption{\textbf{Responses of our Transformer validation episode \#133 ref.~\citep{lake2023human}.}
Grammar, decoding and support set (left) and model responses for the query set (right), shown as counts over 10 stochastic decodes per query. Expected outputs are highlighted in green.}
\label{tab:ep133}

\begin{tabularx}{\textwidth}{@{}p{0.43\textwidth} p{0.54\textwidth}@{}}

\begin{minipage}[t]{\linewidth}
\vspace{0pt}
\centering
\begin{tabular}{c}
\cellcolor{gray!20}\textbf{GRAMMAR \#133} (Lake and Baroni) \\
$\mathrm{wif} \rightarrow \mycirc[myblue],\ 
\mathrm{tufa} \rightarrow \mycirc[myred],\ 
\mathrm{kiki} \rightarrow \mycirc[mypink],\ 
\mathrm{lug} \rightarrow \mycirc[mypurple],$ \\
$u_1\ \mathrm{zup}\ x_1 \rightarrow u_1\ x_1,\qquad
x_1\ \mathrm{gazzer}\ \rightarrow x_1\ x_1\ x_1,$\\
$x_1\ \mathrm{fep} \rightarrow x_1\ x_1$ \\
\cellcolor{gray!20}\textbf{DECODING} (this paper; for readability) \\
$\mathrm{wif} : \mysquare[myblue],\ 
\mathrm{tufa} : \mysquare[myred],\ 
\mathrm{kiki} : \mysquare[mypink],\ 
\mathrm{lug} : \mysquare[mypurple],$ \\
$\mathrm{zup} : \mathtt{before},\ 
\mathrm{gazzer} : \mathtt{thrice},\ 
\mathrm{fep} : \mathtt{twice}$ \\
\cellcolor{gray!20}\textbf{SUPPORT} (Lake and Baroni) \\
$\mysquare[myblue] \rightarrow \mycirc[myblue],\ 
\mysquare[myred] \rightarrow \mycirc[myred],\ 
\mysquare[mypink] \rightarrow \mycirc[mypink],\ 
\mysquare[mypurple] \rightarrow \mycirc[mypurple],$ \\
$\mysquare[myblue]\ \mysquare[myblue] \rightarrow \mycirc[myblue] \mycirc[myblue],\ 
\mysquare[mypink]\ \mysquare[myred] \rightarrow \mycirc[mypink] \mycirc[myred],$ \\
$\mysquare[myblue]\ \mathtt{thrice} \rightarrow \mycirc[myblue] \mycirc[myblue] \mycirc[myblue],\ 
\mysquare[mypink]\ \mysquare[myred]\ \mathtt{thrice} \rightarrow \mycirc[mypink] \mycirc[myred] \mycirc[mypink] \mycirc[myred] \mycirc[mypink] \mycirc[myred],$ \\
$\mysquare[myred]\ \mysquare[myred]\ \mathtt{before}\ \mysquare[myred] \rightarrow \mycirc[myred] \mycirc[myred] \mycirc[myred],\ 
\mysquare[myred]\ \mysquare[myblue]\ \mathtt{before}\ \mysquare[myblue] \rightarrow \mycirc[myred] \mycirc[myblue] \mycirc[myblue],$ \\
$\mysquare[myblue]\ \mathtt{before}\ \mysquare[myred]\ \mathtt{thrice} \rightarrow \mycirc[myblue] \mycirc[myred] \mycirc[myred] \mycirc[myred],$ \\
$\mysquare[mypurple]\ \mathtt{before}\ \mysquare[mypurple]\ \mathtt{before}\ \mysquare[mypurple] \rightarrow \mycirc[mypurple] \mycirc[mypurple] \mycirc[mypurple],$ \\
$\mysquare[mypink]\ \mysquare[mypink]\ \mysquare[mypink]\ \mathtt{before}\ \mysquare[mypink]\ \mathtt{before}\ \mysquare[myred] \rightarrow \mycirc[mypink] \mycirc[mypink] \mycirc[mypink] \mycirc[mypink] \mycirc[myred],$ \\
$\mysquare[mypurple]\ \mathtt{thrice\ twice} \rightarrow \mycirc[mypurple] \mycirc[mypurple] \mycirc[mypurple] \mycirc[mypurple] \mycirc[mypurple] \mycirc[mypurple]$ \\
\end{tabular}
\end{minipage}
&
\begin{minipage}[t]{\linewidth}
\vspace{0pt}
\centering
\begin{tabular}{c}
\cellcolor{gray!20}\textbf{QUERY} (Lake and Baroni) \\
\begin{tabular}{c c c}
\textbf{IN} & \textbf{OUT} & \textbf{\#} \\
\hline
\begin{tabular}{c}
$\mysquare[mypink]\ \mathtt{before}\ \mysquare[myblue]\ \mathtt{before}$\\
$\mysquare[myred]\ \mathtt{twice}$
\end{tabular}
&
\begin{tabular}{c}
$\mycirc[mypink] \mycirc[myblue] \mycirc[myred] \mycirc[myred] \mycirc[myred]$ \\
$\mycirc[mypink] \mycirc[myblue] \mycirc[myred] \mycirc[mypink] \mycirc[myblue] \mycirc[myred]$ \\
$\mycirc[mypink] \mycirc[mypink] \mycirc[myblue] \mycirc[myblue] \mycirc[myred] \mycirc[mygreen]$ \\
$\mycirc[mypink] \mycirc[mypink] \mycirc[myblue] \mycirc[mypurple] \mycirc[myred] \mycirc[myred]$ \\
$\mycirc[mypink] \mycirc[mypink] \mycirc[myblue] \mycirc[myred] \mycirc[myblue] \mycirc[myyellow]$ \\
\cellcolor{green!20}$\mycirc[mypink] \mycirc[myblue] \mycirc[myred] \mycirc[myred]$ \\
\end{tabular}
&
\begin{tabular}{c}
$5$ \\$2$ \\$1$ \\$1$ \\$1$ \\$-$
\end{tabular}
\\
\hline
$\mysquare[myblue]\ \mathtt{twice}$
&
\begin{tabular}{c}
$\mycirc[myblue] \mycirc[myblue] \mycirc[myblue]$ \\
\cellcolor{green!20}$\mycirc[myblue] \mycirc[myblue]$ \\
\end{tabular}
&
\begin{tabular}{c}
$9$ \\$1$
\end{tabular}
\\
\hline
$\mysquare[mypurple]\ \mysquare[myblue]\ \mathtt{twice}$
&
\begin{tabular}{c}
$\mycirc[mypurple] \mycirc[myblue] \mycirc[mypurple] \mycirc[myblue] \mycirc[mypurple] \mycirc[myblue]$ \\
\cellcolor{green!20}$\mycirc[mypurple] \mycirc[myblue] \mycirc[mypurple] \mycirc[myblue]$ \\
$\mycirc[mypurple] \mycirc[myblue] \mycirc[mygreen]$ \\
$\mycirc[mypurple] \mycirc[myblue] \mycirc[mypink]$ \\
$\mycirc[mypurple] \mycirc[myblue] \mycirc[mypurple] \mycirc[mygreen] \mycirc[mypurple] \mycirc[myblue]$ \\
\end{tabular}
&
\begin{tabular}{c}
$5$ \\$2$ \\$1$ \\$1$ \\$1$
\end{tabular}
\\
\hline
$\mysquare[myred]\ \mathtt{twice}$
&
\begin{tabular}{c}
$\mycirc[myred] \mycirc[myred] \mycirc[myred]$ \\
$\mycirc[myred] \mycirc[myred] \mycirc[mypink]$ \\
\cellcolor{green!20}$\mycirc[myred] \mycirc[myred]$ \\
\end{tabular}
&
\begin{tabular}{c}
$9$ \\$1$ \\$-$
\end{tabular}
\\
\hline
\begin{tabular}{c}
$\mysquare[myred]\ \mathtt{before}\ \mysquare[mypurple]\ \mathtt{before}$\\
$\mysquare[myblue]\ \mathtt{twice}$
\end{tabular}
&
\begin{tabular}{c}
$\mycirc[myred] \mycirc[mypurple] \mycirc[myblue] \mycirc[myred] \mycirc[mypurple] \mycirc[myblue]$ \\
$\mycirc[myred] \mycirc[mypurple] \mycirc[myblue] \mycirc[myblue] \mycirc[myblue]$ \\
$\mycirc[myred] \mycirc[myblue] \mycirc[mypurple] \mycirc[myblue] \mycirc[myblue] \mycirc[myblue]$ \\
$\mycirc[myred] \mycirc[mypurple] \mycirc[myblue] \mycirc[myblue] \mycirc[myblue] \mycirc[myblue]$ \\
$\mycirc[myred] \mycirc[myred] \mycirc[mypurple] \mycirc[myred] \mycirc[myblue] \mycirc[mypink]$ \\
\cellcolor{green!20}$\mycirc[myred] \mycirc[mypurple] \mycirc[myblue] \mycirc[myblue]$ \\
\end{tabular}
&
\begin{tabular}{c}
$4$ \\$3$ \\$1$ \\$1$ \\$1$ \\$-$
\end{tabular}
\\
\hline
$\mysquare[myblue]\ \mathtt{before}\ \mysquare[mypink]\ \mathtt{twice}$
&
\begin{tabular}{c}
$\mycirc[myblue] \mycirc[mypink] \mycirc[myblue] \mycirc[mypink] \mycirc[myblue] \mycirc[mypink]$ \\
$\mycirc[myblue] \mycirc[mypink] \mycirc[mypink] \mycirc[mypink]$ \\
$\mycirc[mypink] \mycirc[myblue] \mycirc[mypink] \mycirc[myblue] \mycirc[mypink] \mycirc[myblue]$ \\
\cellcolor{green!20}$\mycirc[myblue] \mycirc[mypink] \mycirc[mypink]$ \\
\end{tabular}
&
\begin{tabular}{c}
$7$ \\$2$ \\$1$ \\$-$
\end{tabular}
\\
\end{tabular}
\end{tabular}
\end{minipage}

\\
\end{tabularx}
\end{table*}

\subsection{Non-systematic parsing}

The hidden grammar allows for an ambiguous interpretation of nested transduction queries, which would normally be a challenge for a systematic reasoner. For example, the query $\langle\mysquare[myblue]\ \mathtt{before}\ \mysquare[mypink]\ \mathtt{twice}\rangle$ could be parsed as either $\langle\mysquare[myblue]\ \mathtt{before}\ (\mysquare[mypink]\ \mathtt{twice})\rangle$ (marked as the target in ref.~\citep{lake2023human}) or $\langle(\mysquare[myblue]\ \mathtt{before}\ \mysquare[mypink])\ \mathtt{twice}\rangle$, and similarly for a query with $\langle\mathtt{thrice}\rangle$.
Although the support set, e.g., $\langle\mysquare[myblue]\ \mathtt{before}\ \mysquare[myred]\ \mathtt{thrice} \rightarrow \mycirc[myblue] \mycirc[myred] \mycirc[myred] \mycirc[myred]\rangle$, provides evidence favouring the intended interpretation, repeated decoding reveals multiple competing output patterns that cannot be explained by a single, stable composition strategy (Table~\ref{tab:ep133}), indicating fragile parse selection. While the common mistakes $\langle\mysquare[mypink]\ \mathtt{before}\ \mysquare[myblue]\ \mathtt{before}\ \mysquare[myred]\ \mathtt{twice} \rightarrow \mycirc[mypink] \mycirc[myblue] \mycirc[myred] \mycirc[myred] \mycirc[myred]\rangle$ and $\langle\mysquare[myred]\ \mathtt{before}\ \mysquare[mypurple]\ \mathtt{before}\ \mysquare[myblue]\ \mathtt{twice} \rightarrow \mycirc[myred] \mycirc[mypurple] \mycirc[myblue] \mycirc[myblue] \mycirc[myblue]\rangle$ could be explained by the parse $\langle u_1\ \mathtt{before}\ (u_2\ \mathtt{before}\ (u_3\ \underline{\mathtt{thrice}}))\rangle$, the error $\langle\mysquare[myblue]\ \mathtt{before}\ \mysquare[mypink]\ \mathtt{twice}\ \rightarrow \mycirc[myblue] \mycirc[mypink] \mycirc[myblue] \mycirc[mypink] \mycirc[myblue] \mycirc[mypink]\rangle$ would instead require the parse $\langle(u_1\ \mathtt{before}\ u_2)\ \underline{\mathtt{thrice}}\rangle$.

\begin{table*}[!h]\footnotesize
\centering
\caption{\textbf{Responses of our Transformer for validation episode \#1 in ref.~\citep{lake2023human} with additional queries.}
Grammar, decoding and support set (left) and selected model responses for the new query set (right), shown as counts over 10 stochastic decodes per query. Expected outputs are highlighted in green.}
\label{tab:ep1}

\begin{tabularx}{\textwidth}{@{}p{0.43\textwidth} p{0.54\textwidth}@{}}

\begin{minipage}[t]{\linewidth}
\vspace{0pt}
\centering
\begin{tabular}{c}
\cellcolor{gray!20}\textbf{GRAMMAR \#1} (Lake and Baroni) \\
$\mathrm{tufa} \rightarrow \mycirc[myyellow],\ 
\mathrm{wif} \rightarrow \mycirc[myblue],$\\
$\mathrm{lug} \rightarrow \mycirc[mypurple],\ 
\mathrm{fep} \rightarrow \mycirc[myred],$ \\
$x_1\ \mathrm{gazzer}\ \rightarrow x_1\ x_1\ x_1,$\\
$x_1\ \mathrm{kiki}\ u_1 \rightarrow x_1\ u_1\ x_1,$\\
$x_1\ \mathrm{zup} \rightarrow x_1\ x_1$ \\
\cellcolor{gray!20}\textbf{DECODING} (this paper; for readability) \\
$\mathrm{tufa} : \mysquare[myyellow],\ 
\mathrm{wif} : \mysquare[myblue],$\\
$\mathrm{lug} : \mysquare[mypurple],\ 
\mathrm{fep} : \mysquare[myred],$ \\
$\mathrm{gazzer} : \mathtt{thrice},\ 
\mathrm{kiki} : \mathtt{around},\ 
\mathrm{zup} : \mathtt{twice}$ \\
\cellcolor{gray!20}\textbf{SUPPORT} (Lake and Baroni) \\
$\mysquare[myblue] \rightarrow \mycirc[myblue],\ 
\mysquare[mypurple] \rightarrow \mycirc[mypurple],\ 
\mysquare[myyellow]\ \mysquare[myyellow] \rightarrow \mycirc[myyellow] \mycirc[myyellow],$ \\
$\mysquare[myred]\ \mathtt{twice} \rightarrow \mycirc[myred] \mycirc[myred],\ 
\mysquare[myred]\ \mathtt{thrice} \rightarrow \mycirc[myred] \mycirc[myred] \mycirc[myred],$ \\
$\mysquare[myred]\ \mysquare[myblue]\ \mathtt{twice} \rightarrow \mycirc[myred] \mycirc[myblue] \mycirc[myred] \mycirc[myblue],$ \\
$\mysquare[myred]\ \mysquare[myred]\ \mathtt{thrice} \rightarrow \mycirc[myred] \mycirc[myred] \mycirc[myred] \mycirc[myred] \mycirc[myred] \mycirc[myred],$ \\
$\mysquare[mypurple]\ \mysquare[myred]\ \mathtt{twice} \rightarrow \mycirc[mypurple] \mycirc[myred] \mycirc[mypurple] \mycirc[myred],$ \\
$\mysquare[myyellow]\ \mysquare[myred]\ \mathtt{thrice} \rightarrow \mycirc[myyellow] \mycirc[myred] \mycirc[myyellow] \mycirc[myred] \mycirc[myyellow] \mycirc[myred],$ \\
$\mysquare[myblue]\ \mathtt{around}\ \mysquare[mypurple] \rightarrow \mycirc[myblue] \mycirc[mypurple] \mycirc[myblue],\ 
\mysquare[myyellow]\ \mathtt{around}\ \mysquare[myyellow] \rightarrow \mycirc[myyellow] \mycirc[myyellow] \mycirc[myyellow],$ \\
$\mysquare[myred]\ \mathtt{around}\ \mysquare[myred]\ \mathtt{around}\ \mysquare[myblue] \rightarrow \mycirc[myred] \mycirc[myred] \mycirc[myred] \mycirc[myblue] \mycirc[myred] \mycirc[myred] \mycirc[myred],$ \\
$\mysquare[myyellow]\ \mathtt{around}\ \mysquare[myblue]\ \mathtt{twice} \rightarrow \mycirc[myyellow] \mycirc[myblue] \mycirc[myyellow] \mycirc[myyellow] \mycirc[myblue] \mycirc[myyellow],$ \\
$\mysquare[myblue]\ \mathtt{thrice}\ \mathtt{around}\ \mysquare[mypurple] \rightarrow \mycirc[myblue] \mycirc[myblue] \mycirc[myblue] \mycirc[mypurple] \mycirc[myblue] \mycirc[myblue] \mycirc[myblue]$ \\
\end{tabular}
\end{minipage}
&
\begin{minipage}[t]{\linewidth}
\vspace{0pt}
\centering
\begin{tabular}{c}
\cellcolor{gray!20}\textbf{QUERY} (new; this paper) \\
\begin{tabular}{c c c}
\textbf{IN} & \textbf{OUT} & \textbf{\#} \\
\hline
$\mysquare[myred]\ \mathtt{thrice}\ \mathtt{around}\ \mysquare[myblue]\ \mysquare[mypurple]$
&
\begin{tabular}{c}
$\mycirc[myred] \mycirc[myred] \mycirc[myred] \mycirc[myblue] \mycirc[mypurple] \mycirc[myred] \mycirc[myred] \mycirc[myred]$ \\
\cellcolor{green!20}$\mycirc[myred] \mycirc[myred] \mycirc[myred] \mycirc[myblue] \mycirc[myred] \mycirc[myred] \mycirc[myred] \mycirc[mypurple]$ \\
\end{tabular}
&
\begin{tabular}{c}
$10$\\$-$
\end{tabular}
\\
\hline
$\mysquare[myred]\ \mathtt{thrice}\ \mathtt{around}\ \mysquare[myblue]\ \mysquare[myyellow]$
&
\begin{tabular}{c}
$\mycirc[myred] \mycirc[myred] \mycirc[myred] \mycirc[myblue] \mycirc[myyellow] \mycirc[myred] \mycirc[myred] \mycirc[myred]$ \\
\cellcolor{green!20}$\mycirc[myred] \mycirc[myred] \mycirc[myred] \mycirc[myblue] \mycirc[myred] \mycirc[myred] \mycirc[myred] \mycirc[myyellow]$ \\
\end{tabular}
&
\begin{tabular}{c}
$8$\\$-$
\end{tabular}
\\
\hline
$\mysquare[myred]\ \mathtt{thrice}\ \mathtt{around}\ \mysquare[myblue]\ \mysquare[myblue]$
&
\begin{tabular}{c}
$\mycirc[myred] \mycirc[myred] \mycirc[myred] \mycirc[myblue] \mycirc[myblue] \mycirc[myred] \mycirc[myred] \mycirc[myred]$ \\
\cellcolor{green!20}$\mycirc[myred] \mycirc[myred] \mycirc[myred] \mycirc[myblue] \mycirc[myred] \mycirc[myred] \mycirc[myred] \mycirc[myblue]$ \\
\end{tabular}
&
\begin{tabular}{c}
$8$\\$-$
\end{tabular}
\\
\hline
$\mysquare[myblue]\ \mathtt{thrice}\ \mathtt{around}\ \mysquare[myyellow]\ \mysquare[myred]$
&
\begin{tabular}{c}
$\mycirc[myblue] \mycirc[myblue] \mycirc[myblue] \mycirc[myyellow] \mycirc[myred] \mycirc[myblue] \mycirc[myblue] \mycirc[myblue]$ \\
\cellcolor{green!20}$\mycirc[myblue] \mycirc[myblue] \mycirc[myblue] \mycirc[myyellow] \mycirc[myblue] \mycirc[myblue] \mycirc[myblue] \mycirc[myred]$ \\
\end{tabular}
&
\begin{tabular}{c}
$9$\\$-$
\end{tabular}
\\
\hline
$\mysquare[myblue]\ \mathtt{thrice}\ \mathtt{around}\ \mysquare[mypurple]\ \mysquare[myblue]$
&
\begin{tabular}{c}
$\mycirc[myblue] \mycirc[myblue] \mycirc[myblue] \mycirc[mypurple] \mycirc[myblue] \mycirc[myblue] \mycirc[myblue]$ \\
\cellcolor{green!20}$\mycirc[myblue] \mycirc[myblue] \mycirc[myblue] \mycirc[mypurple] \mycirc[myblue] \mycirc[myblue] \mycirc[myblue] \mycirc[myblue]$ \\
\end{tabular}
&
\begin{tabular}{c}
$8$\\$-$
\end{tabular}
\\
\hline
$\mysquare[myblue]\ \mathtt{thrice}\ \mathtt{around}\ \mysquare[myblue]\ \mysquare[myblue]$
&
\begin{tabular}{c}
$\mycirc[myblue] \mycirc[myblue] \mycirc[myblue] \mycirc[myblue] \mycirc[myblue] \mycirc[myblue] \mycirc[myblue]$ \\
\cellcolor{green!20}$\mycirc[myblue] \mycirc[myblue] \mycirc[myblue] \mycirc[myblue] \mycirc[myblue] \mycirc[myblue] \mycirc[myblue] \mycirc[myblue]$ \\
\end{tabular}
&
\begin{tabular}{c}
$9$\\$-$
\end{tabular}
\\
\hline
$\mysquare[myred]\ \mathtt{around}\ \mysquare[myblue]\ \mysquare[mypurple]\ \mathtt{twice}$
&
\begin{tabular}{c}
$\mycirc[myred] \mycirc[myblue] \mycirc[mypurple] \mycirc[myred] \mycirc[myred] \mycirc[myblue] \mycirc[mypurple] \mycirc[myred]$ \\
\cellcolor{green!20}$\mycirc[myred] \mycirc[myblue] \mycirc[myred] \mycirc[mypurple] \mycirc[myred] \mycirc[myblue] \mycirc[myred] \mycirc[mypurple]$ \\
\end{tabular}
&
\begin{tabular}{c}
$8$\\$-$
\end{tabular}
\\
\hline
$\mysquare[myred]\ \mathtt{around}\ \mysquare[myblue]\ \mysquare[myblue]\ \mathtt{twice}$
&
\begin{tabular}{c}
$\mycirc[myred] \mycirc[myblue] \mycirc[myblue] \mycirc[myred] \mycirc[myred] \mycirc[myblue] \mycirc[myblue] \mycirc[myred]$ \\
\cellcolor{green!20}$\mycirc[myred] \mycirc[myblue] \mycirc[myred] \mycirc[myblue] \mycirc[myred] \mycirc[myblue] \mycirc[myred] \mycirc[myblue]$ \\
\end{tabular}
&
\begin{tabular}{c}
$8$\\$-$
\end{tabular}
\\
\hline
$\mysquare[myblue]\ \mathtt{around}\ \mysquare[myred]\ \mysquare[myblue]\ \mathtt{twice}$
&
\begin{tabular}{c}
\cellcolor{green!20}$\mycirc[myblue] \mycirc[myred] \mycirc[myblue] \mycirc[myblue] \mycirc[myblue] \mycirc[myred] \mycirc[myblue] \mycirc[myblue]$ \\
\end{tabular}
&
\begin{tabular}{c}
$6$
\end{tabular}
\\
\hline
$\mysquare[myblue]\ \mathtt{around}\ \mysquare[myblue]\ \mysquare[myred]\ \mathtt{twice}$
&
\begin{tabular}{c}
$\mycirc[myblue] \mycirc[myblue] \mycirc[myred] \mycirc[myblue] \mycirc[myblue] \mycirc[myblue] \mycirc[myred] \mycirc[myblue]$ \\
\cellcolor{green!20}$\mycirc[myblue] \mycirc[myblue] \mycirc[myblue] \mycirc[myred] \mycirc[myblue] \mycirc[myblue] \mycirc[myblue] \mycirc[myred]$ \\
\end{tabular}
&
\begin{tabular}{c}
$7$\\$-$
\end{tabular}
\\
\hline
$\mysquare[myblue]\ \mathtt{around}\ \mysquare[myblue]\ \mysquare[myblue]\ \mathtt{twice}$
&
\begin{tabular}{c}
$\mycirc[myblue] \mycirc[myblue] \mycirc[myblue] \mycirc[myblue] \mycirc[myblue] \mycirc[myblue]$ \\
\cellcolor{green!20}$\mycirc[myblue] \mycirc[myblue] \mycirc[myblue] \mycirc[myblue] \mycirc[myblue] \mycirc[myblue] \mycirc[myblue] \mycirc[myblue]$ \\
\end{tabular}
&
\begin{tabular}{c}
$5$\\$2$
\end{tabular}
\\
\end{tabular}
\end{tabular}
\end{minipage}

\\
\end{tabularx}
\end{table*}

\subsection{Structure-sensitivity tests based on Episode \#1}

To isolate parse-related behaviour, we additionally evaluated Episode~\#1 using the original grammar and support set but with novel constructed query patterns (Table~\ref{tab:ep1}). We consider queries of the form $\langle u_1\ \mathtt{thrice}\ \mathtt{around}\ u_2\ u_3\rangle$ and $\langle u_1\ \mathtt{around}\ u_2\ u_3\ \mathtt{twice}\rangle$, comparing the intended bracketings
$\langle ((u_1\ \mathtt{thrice})\ \mathtt{around}\ u_2)\ u_3\rangle$ and $\langle ((u_1\ \mathtt{around}\ u_2)\ u_3)\ \mathtt{twice}\rangle$
with the alternative bracketings
$\langle (u_1\ \mathtt{thrice})\ \mathtt{around}\ (u_2\ u_3)\rangle$ and $\langle (u_1\ \mathtt{around}\ (u_2\ u_3))\ \mathtt{twice}\rangle$.

Across repeated decoding, the model's most frequent outputs are consistent with the alternative bracketing of $\mathtt{around}$ for inputs where the two bracketings yield different outputs (Table~\ref{tab:ep1}). For some inputs, both bracketings yield identical outputs (e.g., $\langle\mysquare[myblue]\ \mathtt{thrice}\ \mathtt{around}\ \mysquare[mypurple]\ \mysquare[myblue]\rangle$ and $\langle\mysquare[myblue]\ \mathtt{around}\ \mysquare[myblue]\ \mysquare[myblue]\ \mathtt{twice}\rangle$), so correct responses do not distinguish between them. Other sampled responses contain errors in colour or repetition, so the alternative bracketing does not explain all errors (Table~\ref{tab:ep1extSUPPORT}).

\subsection{Extended outputs for episode-level re-evaluation}
\label{sec:supp_extended_outputs}

We provide the full grammars, support sets and query sets for selected episodes \#133, \#32 and \#122 and \#1 together with output distributions under repeated decoding (Tables~\ref{tab:apx_133}--\ref{tab:ep1extSUPPORT}).

\begin{sidewaystable*}[p]\footnotesize
\centering
\caption{\textbf{Complete responses of our Transformer for  meta-learning validation episode \#133 in ref.~\citep{lake2023human} (full query set).}
Grammar, decoding and support set (left) and model responses for the query set (right), shown as counts over 10 stochastic decodes per query. Expected outputs are highlighted in green.}
\label{tab:apx_133}

\begin{tabularx}{\textwidth}{@{}p{0.43\textwidth} p{0.54\textwidth}@{}}

\begin{minipage}[t]{\linewidth}
\vspace{0pt}
\centering
\begin{tabular}{c}
\cellcolor{gray!20}\textbf{GRAMMAR \#133} (Lake and Baroni) \\
$\mathrm{wif} \rightarrow \mycirc[myblue],\ 
\mathrm{tufa} \rightarrow \mycirc[myred],\ 
\mathrm{kiki} \rightarrow \mycirc[mypink],\ 
\mathrm{lug} \rightarrow \mycirc[mypurple],$ \\
$u_1\ \mathrm{zup}\ x_1 \rightarrow u_1\ x_1,$ \qquad
$x_1\ \mathrm{gazzer}\ \rightarrow x_1\ x_1\ x_1,$\\
$x_1\ \mathrm{fep} \rightarrow x_1\ x_1$ \\
\cellcolor{gray!20}\textbf{DECODING} (this paper; for readability) \\
$\mathrm{wif} : \mysquare[myblue],\ 
\mathrm{tufa} : \mysquare[myred],\ 
\mathrm{kiki} : \mysquare[mypink],\ 
\mathrm{lug} : \mysquare[mypurple],$ \\
$\mathrm{zup} : \mathtt{before},\ 
\mathrm{gazzer} : \mathtt{thrice},\ 
\mathrm{fep} : \mathtt{twice}$ \\
\cellcolor{gray!20}\textbf{SUPPORT} (Lake and Baroni) \\
$\mysquare[myblue] \rightarrow \mycirc[myblue],\ 
\mathtt{\mysquare[myred]} \rightarrow \mycirc[myred],\ 
\mathtt{\mysquare[mypink]} \rightarrow \mycirc[mypink],\ 
\mathtt{\mysquare[mypurple]} \rightarrow \mycirc[mypurple],$ \\
$\mathtt{\mysquare[myblue]\ \mysquare[myblue]} \rightarrow \mycirc[myblue] \mycirc[myblue],\ 
\mathtt{\mysquare[mypink]\ \mysquare[myred]} \rightarrow \mycirc[mypink] \mycirc[myred],$ \\
$\mysquare[myblue]\ \mathtt{thrice} \rightarrow \mycirc[myblue] \mycirc[myblue] \mycirc[myblue],\ 
\mysquare[mypink]\ \mysquare[myred]\ \mathtt{thrice} \rightarrow \mycirc[mypink] \mycirc[myred] \mycirc[mypink] \mycirc[myred] \mycirc[mypink] \mycirc[myred],$ \\
$\mysquare[myred]\ \mysquare[myred]\ \mathtt{before}\ \mysquare[myred] \rightarrow \mycirc[myred] \mycirc[myred] \mycirc[myred],\ 
\mathtt{\mysquare[myred]\ \mysquare[myblue]\ before\ \mysquare[myblue]} \rightarrow \mycirc[myred] \mycirc[myblue] \mycirc[myblue],$ \\
$\mysquare[myblue]\ \mathtt{before}\ \mysquare[myred]\ \mathtt{thrice} \rightarrow \mycirc[myblue] \mycirc[myred] \mycirc[myred] \mycirc[myred],$ \\
$\mathtt{\mysquare[mypurple]\ before\ \mysquare[mypurple]\ before\ \mysquare[mypurple]} \rightarrow \mycirc[mypurple] \mycirc[mypurple] \mycirc[mypurple],$ \\
$\mathtt{\mysquare[mypink]\ \mysquare[mypink]\ \mysquare[mypink]\ before\ \mysquare[mypink]\ before\ \mysquare[myred]} \rightarrow \mycirc[mypink] \mycirc[mypink] \mycirc[mypink] \mycirc[mypink] \mycirc[myred],$ \\
$\mathtt{\mysquare[mypurple]\ thrice\ twice} \rightarrow \mycirc[mypurple] \mycirc[mypurple] \mycirc[mypurple] \mycirc[mypurple] \mycirc[mypurple] \mycirc[mypurple]$ \\
\cellcolor{gray!20}\textbf{QUERY} (Lake and Baroni) \\
\begin{tabular}{c c c}
\textbf{IN} & \textbf{OUT} & \textbf{\#} \\
\hline
$\mysquare[myred]\ \mathtt{twice}$
&
\begin{tabular}{c}
$\mycirc[myred] \mycirc[myred] \mycirc[myred]$ \\
$\mycirc[myred] \mycirc[myred] \mycirc[mypink]$ \\
\cellcolor{green!20}$\mycirc[myred] \mycirc[myred]$ \\
\end{tabular}
&
\begin{tabular}{c}
$9$ \\$1$ \\$-$
\end{tabular}
\\
\hline
$\mysquare[myblue]\ \mathtt{twice}$
&
\begin{tabular}{c}
$\mycirc[myblue] \mycirc[myblue] \mycirc[myblue]$ \\
\cellcolor{green!20}$\mycirc[myblue] \mycirc[myblue]$ \\
\end{tabular}
&
\begin{tabular}{c}
$9$ \\$1$
\end{tabular}
\\
\end{tabular}
\end{tabular}
\end{minipage}
&
\begin{minipage}[t]{\linewidth}
\vspace{0pt}
\centering
\begin{tabular}{c}
\cellcolor{gray!20}\textbf{QUERY} (Lake and Baroni) \\
\begin{tabular}{c c c}
\textbf{IN} & \textbf{OUT} & \textbf{\#} \\
\hline
\begin{tabular}{c}
$\mysquare[mypink]\ \mathtt{before}\ \mysquare[myblue]\ \mathtt{before}$\\
$\mysquare[myred]\ \mathtt{twice}$
\end{tabular}
&
\begin{tabular}{c}
$\mycirc[mypink] \mycirc[myblue] \mycirc[myred] \mycirc[myred] \mycirc[myred]$ \\
$\mycirc[mypink] \mycirc[myblue] \mycirc[myred] \mycirc[mypink] \mycirc[myblue] \mycirc[myred]$ \\
$\mycirc[mypink] \mycirc[mypink] \mycirc[myblue] \mycirc[myblue] \mycirc[myred] \mycirc[mygreen]$ \\
$\mycirc[mypink] \mycirc[mypink] \mycirc[myblue] \mycirc[mypurple] \mycirc[myred] \mycirc[myred]$ \\
$\mycirc[mypink] \mycirc[mypink] \mycirc[myblue] \mycirc[myred] \mycirc[myblue] \mycirc[myyellow]$ \\
\cellcolor{green!20}$\mycirc[mypink] \mycirc[myblue] \mycirc[myred] \mycirc[myred]$ \\
\end{tabular}
&
\begin{tabular}{c}
$5$ \\$2$ \\$1$ \\$1$ \\$1$ \\$-$
\end{tabular}
\\
\hline
$\mysquare[mypurple]\ \mysquare[myblue]\ \mathtt{twice}$
&
\begin{tabular}{c}
$\mycirc[mypurple] \mycirc[myblue] \mycirc[mypurple] \mycirc[myblue] \mycirc[mypurple] \mycirc[myblue]$ \\
\cellcolor{green!20}$\mycirc[mypurple] \mycirc[myblue] \mycirc[mypurple] \mycirc[myblue]$ \\
$\mycirc[mypurple] \mycirc[myblue] \mycirc[mygreen]$ \\
$\mycirc[mypurple] \mycirc[myblue] \mycirc[mypink]$ \\
$\mycirc[mypurple] \mycirc[myblue] \mycirc[mypurple] \mycirc[mygreen] \mycirc[mypurple] \mycirc[myblue]$ \\
\end{tabular}
&
\begin{tabular}{c}
$5$ \\$2$ \\$1$ \\$1$ \\$1$
\end{tabular}
\\
\hline
\begin{tabular}{c}
$\mysquare[myred]\ \mathtt{before}\ \mysquare[mypurple]\ \mathtt{before}$\\
$\mysquare[myblue]\ \mathtt{twice}$
\end{tabular}
&
\begin{tabular}{c}
$\mycirc[myred] \mycirc[mypurple] \mycirc[myblue] \mycirc[myred] \mycirc[mypurple] \mycirc[myblue]$ \\
$\mycirc[myred] \mycirc[mypurple] \mycirc[myblue] \mycirc[myblue] \mycirc[myblue]$ \\
$\mycirc[myred] \mycirc[myblue] \mycirc[mypurple] \mycirc[myblue] \mycirc[myblue] \mycirc[myblue]$ \\
$\mycirc[myred] \mycirc[mypurple] \mycirc[myblue] \mycirc[myblue] \mycirc[myblue] \mycirc[myblue]$ \\
$\mycirc[myred] \mycirc[myred] \mycirc[mypurple] \mycirc[myred] \mycirc[myblue] \mycirc[mypink]$ \\
\cellcolor{green!20}$\mycirc[myred] \mycirc[mypurple] \mycirc[myblue] \mycirc[myblue]$ \\
\end{tabular}
&
\begin{tabular}{c}
$4$ \\$3$ \\$1$ \\$1$ \\$1$ \\$-$
\end{tabular}
\\
\hline
$\mysquare[myblue]\ \mathtt{before}\ \mysquare[mypink]\ \mathtt{twice}$
&
\begin{tabular}{c}
$\mycirc[myblue] \mycirc[mypink] \mycirc[myblue] \mycirc[mypink] \mycirc[myblue] \mycirc[mypink]$ \\
$\mycirc[myblue] \mycirc[mypink] \mycirc[mypink] \mycirc[mypink]$ \\
$\mycirc[mypink] \mycirc[myblue] \mycirc[mypink] \mycirc[myblue] \mycirc[mypink] \mycirc[myblue]$ \\
\cellcolor{green!20}$\mycirc[myblue] \mycirc[mypink] \mycirc[mypink]$ \\
\end{tabular}
&
\begin{tabular}{c}
$7$ \\$2$ \\$1$ \\$-$
\end{tabular}
\\
\hline
$\mysquare[myblue]\ \mysquare[mypink]$
&
\begin{tabular}{c}
\cellcolor{green!20}$\mycirc[myblue] \mycirc[mypink]$ \\
\end{tabular}
&
\begin{tabular}{c}
$10$
\end{tabular}
\\
\hline
$\mysquare[mypurple]\ \mathtt{before}\ \mysquare[mypurple]$
&
\begin{tabular}{c}
\cellcolor{green!20}$\mycirc[mypurple] \mycirc[mypurple]$ \\
$\mycirc[mypurple] \mycirc[mypink] \mycirc[mypurple]$ \\
\end{tabular}
&
\begin{tabular}{c}
$9$ \\$1$
\end{tabular}
\\
\hline
$\mysquare[myblue]\ \mathtt{before}\ \mysquare[myred]$
&
\begin{tabular}{c}
\cellcolor{green!20}$\mycirc[myblue] \mycirc[myred]$ \\
\end{tabular}
&
\begin{tabular}{c}
$10$
\end{tabular}
\\
\hline
$\mysquare[myblue]\ \mathtt{before}\ \mysquare[myblue]$
&
\begin{tabular}{c}
\cellcolor{green!20}$\mycirc[myblue] \mycirc[myblue]$ \\
\end{tabular}
&
\begin{tabular}{c}
$10$
\end{tabular}
\\
\end{tabular}
\end{tabular}
\end{minipage}

\\
\end{tabularx}
\end{sidewaystable*}

\begin{sidewaystable*}[p]\footnotesize
\centering
\caption{\textbf{Complete responses of our Transformer for meta-learning validation episode \#32 in ref.~\citep{lake2023human}.}
Grammar, decoding and support set (left) and model responses for the query set (right), shown as counts over 10 stochastic decodes per query. Expected outputs are highlighted in green.}
\label{tab:apx_32}

\begin{tabularx}{\textwidth}{@{}p{0.43\textwidth} p{0.54\textwidth}@{}}

\begin{minipage}[t]{\linewidth}
\vspace{0pt}
\centering
\begin{tabular}{c}
\cellcolor{gray!20}\textbf{GRAMMAR \#32} (Lake and Baroni) \\
$\mathrm{tufa} \rightarrow \mycirc[myred],\ 
\mathrm{zup} \rightarrow \mycirc[myyellow],\ 
\mathrm{kiki} \rightarrow \mycirc[mygreen],\ 
\mathrm{lug} \rightarrow \mycirc[mypurple],$ \\
$x_1\ \mathrm{dax}\ u_1 \rightarrow u_1\ x_1,$ \\
$x_1\ \mathrm{gazzer}\ x_2\ \rightarrow x_1\ x_2,$\\
$x_1\ \mathrm{wif}\ x_2 \rightarrow x_1\ x_1\ x_2\ x_2\ x_2\ x_1$ \\
\cellcolor{gray!20}\textbf{DECODING} (this paper; for readability) \\
$\mathrm{tufa} : \mysquare[myred],\ 
\mathrm{zup} : \mysquare[myyellow],\ 
\mathrm{kiki} : \mysquare[mygreen],\ 
\mathrm{lug} : \mysquare[mypurple],$ \\
$\mathrm{dax} : \mathtt{after},\ 
\mathrm{gazzer} : \mathtt{before},\ 
\mathrm{wif} : \mathtt{twice\ before\ and\ once\ after}$\\
$\mathtt{three\ times}$ \\
\cellcolor{gray!20}\textbf{SUPPORT} (Lake and Baroni) \\
$\mysquare[myred] \rightarrow \mycirc[myred],\ 
\mysquare[myyellow] \rightarrow \mycirc[myyellow],\ 
\mysquare[mypurple] \rightarrow \mycirc[mypurple],$\\
$\mysquare[myred]\ \mysquare[myred] \rightarrow \mycirc[myred] \mycirc[myred],$\\
$\mysquare[mygreen]\ \mysquare[mygreen] \rightarrow \mycirc[mygreen] \mycirc[mygreen],$\\
$\mysquare[mypurple]\ \mysquare[mypurple] \rightarrow \mycirc[mypurple] \mycirc[mypurple],$\\
$\mysquare[myred]\ \mysquare[mypurple] \rightarrow \mycirc[myred] \mycirc[mypurple],$\\
$\mysquare[myred]\ \mysquare[mygreen]\ \mysquare[mypurple] \rightarrow \mycirc[myred] \mycirc[mygreen] \mycirc[mypurple],$\\
$\mysquare[mygreen]\ \mathtt{after}\ \mysquare[mygreen] \rightarrow \mycirc[mygreen] \mycirc[mygreen],$\\
$\mysquare[mypurple]\ \mathtt{before}\ \mysquare[mypurple] \rightarrow \mycirc[mypurple] \mycirc[mypurple],$\\
$\mysquare[mypurple]\ \mysquare[myyellow]\ \mathtt{after}\ \mysquare[myyellow] \rightarrow \mycirc[myyellow] \mycirc[mypurple] \mycirc[myyellow],$\\
$\mysquare[myred]\ \mathtt{after}\ \mysquare[mygreen]\ \mathtt{after}\ \mysquare[myyellow] \rightarrow \mycirc[myyellow] \mycirc[mygreen] \mycirc[myred],$\\
$\mysquare[myred]\ \mathtt{after}\ \mysquare[mypurple]\ \mathtt{after}\ \mysquare[mypurple] \rightarrow \mycirc[mypurple] \mycirc[mypurple] \mycirc[myred],$\\
$\mysquare[mypurple]\ \mysquare[myyellow]\ \mathtt{after}\ \mysquare[myred]\ \mathtt{after}\ \mysquare[myyellow] \rightarrow \mycirc[myyellow] \mycirc[myred] \mycirc[mypurple] \mycirc[myyellow]$\\

\cellcolor{gray!20}\textbf{QUERY} (Lake and Baroni) \\
\begin{tabular}{c c c}
\textbf{IN} & \textbf{OUT} & \textbf{\#} \\
\hline
$\mysquare[mygreen]$
&
\begin{tabular}{c}
\cellcolor{green!20}$\mycirc[mygreen]$ \\
$\mycirc[myblue]$ \\
\end{tabular}
&
\begin{tabular}{c}
$9$ \\$1$
\end{tabular}
\\
\hline
$\mysquare[mygreen]\ \mathtt{after}\ \mysquare[myyellow]$
&
\begin{tabular}{c}
\cellcolor{green!20}$\mycirc[myyellow] \mycirc[mygreen]$ \\
\end{tabular}
&
\begin{tabular}{c}
$10$
\end{tabular}
\\
\hline
$\mysquare[myred]\ \mysquare[myyellow]\ \mathtt{before}\ \mysquare[myyellow]$
&
\begin{tabular}{c}
\cellcolor{green!20}$\mycirc[myred] \mycirc[myyellow] \mycirc[myyellow]$ \\
$\mycirc[myred] \mycirc[myyellow] \mycirc[mygreen] \mycirc[myyellow]$ \\
$\mycirc[myyellow] \mycirc[myred] \mycirc[myyellow]$ \\
\end{tabular}
&
\begin{tabular}{c}
$8$ \\$1$ \\$1$
\end{tabular}
\\
\hline
$\mysquare[myyellow]\ \mysquare[myyellow]$
&
\begin{tabular}{c}
\cellcolor{green!20}$\mycirc[myyellow] \mycirc[myyellow]$ \\
\end{tabular}
&
\begin{tabular}{c}
$10$
\end{tabular}
\\
\end{tabular}
\end{tabular}
\end{minipage}
&
\begin{minipage}[t]{\linewidth}
\vspace{0pt}
\centering
\begin{tabular}{c}
\cellcolor{gray!20}\textbf{QUERY} (Lake and Baroni) \\
\begin{tabular}{c c c}
\textbf{IN} & \textbf{OUT} & \textbf{\#} \\
\hline
\begin{tabular}{c}
$\mysquare[mygreen]\ \mathtt{after}\ \mysquare[myyellow]\ \mathtt{after}\ \mysquare[mygreen]$\\
$\mathtt{before}\ \mysquare[myyellow]$
\end{tabular}
&
\begin{tabular}{c}
$\mycirc[myyellow] \mycirc[mygreen] \mycirc[myyellow] \mycirc[mygreen]$ \\
$\mycirc[mygreen] \mycirc[mygreen] \mycirc[myyellow] \mycirc[myyellow]$ \\
$\mycirc[myblue] \mycirc[mypink] \mycirc[myyellow] \mycirc[mygreen] \mycirc[mypurple] \mycirc[myred] \mycirc[myyellow]$ \\
$\mycirc[mygreen] \mycirc[mygreen] \mycirc[myyellow] \mycirc[mygreen]$ \\
$\mycirc[mygreen] \mycirc[myyellow] \mycirc[myyellow] \mycirc[mygreen]$ \\
\cellcolor{green!20}$\mycirc[mygreen] \mycirc[myyellow] \mycirc[mygreen] \mycirc[myyellow]$ \\
\end{tabular}
&
\begin{tabular}{c}
$5$ \\$2$ \\$1$ \\$1$ \\$1$ \\$-$
\end{tabular}
\\
\hline
$\mysquare[mygreen]\ \mysquare[mypurple]\ \mathtt{after}\ \mysquare[myred]\ \mathtt{after}\ \mysquare[myyellow]$
&
\begin{tabular}{c}
\cellcolor{green!20}$\mycirc[myyellow] \mycirc[myred] \mycirc[mygreen] \mycirc[mypurple]$ \\
$\mycirc[mygreen] \mycirc[myyellow] \mycirc[myred] \mycirc[mypurple]$ \\
$\mycirc[mypurple] \mycirc[mypurple] \mycirc[myblue] \mycirc[myred] \mycirc[myred] \mycirc[myyellow]$ \\
\end{tabular}
&
\begin{tabular}{c}
$8$ \\$1$ \\$1$
\end{tabular}
\\
\hline
\begin{tabular}{c}
$\mysquare[mygreen]\ \mysquare[myred]\ \mysquare[mypurple]\ \mathtt{before}\ \mysquare[mygreen]$\\
$\mathtt{before}\ \mysquare[mypurple]$
\end{tabular}
&
\begin{tabular}{c}
$\mycirc[mypurple] \mycirc[mygreen] \mycirc[mygreen] \mycirc[myred] \mycirc[mypurple]$ \\
\cellcolor{green!20}$\mycirc[mygreen] \mycirc[myred] \mycirc[mypurple] \mycirc[mygreen] \mycirc[mypurple]$ \\
$\mycirc[mygreen] \mycirc[myred] \mycirc[mygreen] \mycirc[mypurple] \mycirc[mypurple]$ \\
$\mycirc[mypurple] \mycirc[mygreen] \mycirc[mypink] \mycirc[myred] \mycirc[mypurple]$ \\
\end{tabular}
&
\begin{tabular}{c}
$6$ \\$2$ \\$1$ \\$1$
\end{tabular}
\\
\hline
\begin{tabular}{c}
$\mysquare[mypurple]\ \mathtt{twice\ before}$\\
$\mathtt{and\ once\ after}$\\
$\mathtt{three\ times}\ \mysquare[mygreen]$
\end{tabular}
&
\begin{tabular}{c}
$\mycirc[mygreen] \mycirc[mygreen] \mycirc[mypurple]$ \\
$\mycirc[mygreen] \mycirc[mypurple]$ \\
$\mycirc[mygreen] \mycirc[mypurple] \mycirc[mygreen] \mycirc[mypurple] \mycirc[mygreen] \mycirc[mygreen]$ \\
$\mycirc[mygreen] \mycirc[mypurple] \mycirc[mypurple] \mycirc[mygreen] \mycirc[mypurple] \mycirc[mypurple]$ \\
$\mycirc[mypurple] \mycirc[mygreen]$ \\
$\mycirc[mypurple] \mycirc[mygreen] \mycirc[mygreen] \mycirc[mypurple] \mycirc[mypurple]$ \\
$\mycirc[mypurple] \mycirc[mygreen] \mycirc[mypurple]$ \\
$\mycirc[mypurple] \mycirc[mypurple] \mycirc[mypurple] \mycirc[mygreen] \mycirc[mygreen]$ \\
$\mycirc[mypurple] \mycirc[mypurple] \mycirc[mypurple] \mycirc[mygreen] \mycirc[mygreen] \mycirc[mypurple]$ \\
$\mycirc[mypurple] \mycirc[mypurple] \mycirc[mypurple] \mycirc[mypurple] \mycirc[mygreen] \mycirc[mygreen]$ \\
\cellcolor{green!20}$\mycirc[mypurple] \mycirc[mypurple] \mycirc[mygreen] \mycirc[mygreen] \mycirc[mygreen] \mycirc[mypurple]$ \\
\end{tabular}
&
\begin{tabular}{c}
$1$ \\$1$ \\$1$ \\$1$ \\$1$ \\$1$ \\$1$ \\$1$ \\$1$ \\$1$ \\$-$
\end{tabular}
\\
\hline
\begin{tabular}{c}
$\mysquare[mypurple]\ \mathtt{twice\ before}$\\
$\mathtt{and\ once\ after}$\\
$\mathtt{three\ times}\ \mysquare[myyellow]$
\end{tabular}
&
\begin{tabular}{c}
$\mycirc[mypurple] \mycirc[mypurple] \mycirc[mypurple] \mycirc[myyellow] \mycirc[mypurple]$ \\
$\mycirc[mypurple] \mycirc[mypurple] \mycirc[mypurple] \mycirc[myyellow] \mycirc[myyellow]$ \\
$\mycirc[mypurple] \mycirc[mypurple] \mycirc[myyellow]$ \\
$\mycirc[mypurple] \mycirc[myyellow] \mycirc[mypurple]$ \\
$\mycirc[myyellow] \mycirc[mypink]$ \\
$\mycirc[myyellow] \mycirc[mypurple]$ \\
$\mycirc[myyellow] \mycirc[mypurple] \mycirc[mypurple]$ \\
$\mycirc[myyellow] \mycirc[mypurple] \mycirc[mypurple] \mycirc[myyellow] \mycirc[myyellow] \mycirc[myyellow]$ \\
$\mycirc[myyellow] \mycirc[myyellow] \mycirc[mypurple] \mycirc[mypurple] \mycirc[mypurple]$ \\
$\mycirc[myyellow] \mycirc[myyellow] \mycirc[mypurple] \mycirc[mypurple] \mycirc[mypurple] \mycirc[mypurple]$ \\
\cellcolor{green!20}$\mycirc[mypurple] \mycirc[mypurple] \mycirc[myyellow] \mycirc[myyellow] \mycirc[myyellow] \mycirc[mypurple]$ \\
\end{tabular}
&
\begin{tabular}{c}
$1$ \\$1$ \\$1$ \\$1$ \\$1$ \\$1$ \\$1$ \\$1$ \\$1$ \\$1$ \\$-$
\end{tabular}
\\
\hline
\begin{tabular}{c}
$\mysquare[myyellow]\ \mysquare[myred]\ \mysquare[myred]\ \mathtt{after}\ \mysquare[mygreen]$\\
$\mathtt{before}\ \mysquare[mypurple]$
\end{tabular}
&
\begin{tabular}{c}
$\mycirc[mypurple] \mycirc[mygreen] \mycirc[myyellow] \mycirc[myred] \mycirc[myred]$ \\
$\mycirc[mypurple] \mycirc[myred] \mycirc[myred] \mycirc[mygreen] \mycirc[myblue] \mycirc[mypurple] \mycirc[mypurple]$ \\
$\mycirc[mypurple] \mycirc[myyellow] \mycirc[myred] \mycirc[mygreen] \mycirc[myred]$ \\
$\mycirc[myyellow] \mycirc[myred] \mycirc[mygreen] \mycirc[myred] \mycirc[mypurple]$ \\
$\mycirc[myyellow] \mycirc[myred] \mycirc[mygreen] \mycirc[myyellow] \mycirc[myred] \mycirc[mypink] \mycirc[mypurple]$ \\
$\mycirc[myyellow] \mycirc[myred] \mycirc[myred] \mycirc[mypink] \mycirc[myyellow] \mycirc[myyellow] \mycirc[mypurple]$ \\
$\mycirc[myyellow] \mycirc[myred] \mycirc[myred] \mycirc[mypurple] \mycirc[mygreen] \mycirc[mypink] \mycirc[mypurple]$ \\
\cellcolor{green!20}$\mycirc[mygreen] \mycirc[myyellow] \mycirc[myred] \mycirc[myred] \mycirc[mypurple]$ \\
\end{tabular}
&
\begin{tabular}{c}
$4$ \\$1$ \\$1$ \\$1$ \\$1$ \\$1$ \\$1$ \\$-$
\end{tabular}
\\
\end{tabular}
\end{tabular}
\end{minipage}

\\
\end{tabularx}
\end{sidewaystable*}

\begin{sidewaystable*}[p]\footnotesize
\centering
\caption{\textbf{Complete responses of our Transformer for meta-learning validation episode \#122 in ref.~\citep{lake2023human}.}
Grammar, decoding and support set (left) and model responses for the query set (right), shown as counts over 10 stochastic decodes per query. Expected outputs are highlighted in green.}
\label{tab:apx_122}

\begin{tabularx}{\textwidth}{@{}p{0.43\textwidth} p{0.54\textwidth}@{}}

\begin{minipage}[t]{\linewidth}
\vspace{0pt}
\centering
\begin{tabular}{c}
\cellcolor{gray!20}\textbf{GRAMMAR \#122} (Lake and Baroni) \\
$\mathrm{blicket} \rightarrow \mycirc[myred],\ 
\mathrm{kiki} \rightarrow \mycirc[mygreen],\ 
\mathrm{zup} \rightarrow \mycirc[myyellow],\ 
\mathrm{lug} \rightarrow \mycirc[myblue],$ \\
$x_1\ \mathrm{dax} \rightarrow x_1\ x_1\ x_1\ x_1,$\\
$u_1\ \mathrm{fep}\ x_1\ \rightarrow x_1\ u_1\ u_1\ x_1,$\\
$x_1\ \mathrm{gazzer} \rightarrow x_1\ x_1$ \\
\cellcolor{gray!20}\textbf{DECODING} (this paper; for readability) \\
$\mathrm{blicket} : \mysquare[myred],\ 
\mathrm{kiki} : \mysquare[mygreen],\ 
\mathrm{zup} : \mysquare[myyellow],\ 
\mathrm{lug} : \mysquare[myblue],$ \\
$\mathrm{dax} : \mathtt{four\ times},\ 
\mathrm{fep} : \mathtt{twice\ within},\ 
\mathrm{gazzer} : \mathtt{twice}$ \\
\cellcolor{gray!20}\textbf{SUPPORT} (Lake and Baroni) \\
$\mysquare[myblue] \rightarrow \mycirc[myblue],\ 
\mysquare[mygreen] \rightarrow \mycirc[mygreen],\ 
\mysquare[myred] \rightarrow \mycirc[myred],$\\
$\mysquare[mygreen]\ \mysquare[myyellow] \rightarrow \mycirc[mygreen] \mycirc[myyellow],$\\
$\mysquare[myyellow]\ \mysquare[mygreen] \rightarrow \mycirc[myyellow] \mycirc[mygreen],$\\
$\mysquare[myblue]\ \mathtt{twice} \rightarrow \mycirc[myblue] \mycirc[myblue],$\\
$\mysquare[mygreen]\ \mathtt{twice} \rightarrow \mycirc[mygreen] \mycirc[mygreen],$\\
$\mysquare[myred]\ \mathtt{four\ times} \rightarrow \mycirc[myred] \mycirc[myred] \mycirc[myred] \mycirc[myred],$\\
$\mysquare[mygreen]\ \mathtt{four\ times} \rightarrow \mycirc[mygreen] \mycirc[mygreen] \mycirc[mygreen] \mycirc[mygreen],$\\
$\mysquare[myblue]\ \mysquare[myyellow]\ \mathtt{twice} \rightarrow \mycirc[myblue] \mycirc[myyellow] \mycirc[myblue] \mycirc[myyellow],$\\
$\mysquare[mygreen]\ \mysquare[myyellow]\ \mathtt{twice} \rightarrow \mycirc[mygreen] \mycirc[myyellow] \mycirc[mygreen] \mycirc[myyellow],$\\
$\mysquare[mygreen]\ \mysquare[myblue]\ \mysquare[myblue]\ \mysquare[myyellow]\ \mathtt{twice} \rightarrow \mycirc[mygreen] \mycirc[myblue] \mycirc[myblue] \mycirc[myyellow] \mycirc[mygreen] \mycirc[myblue] \mycirc[myblue] \mycirc[myyellow],$\\
$\mysquare[myred]\ \mathtt{twice\ within}\ \mysquare[myred]\ \mathtt{twice} \rightarrow \mycirc[myred] \mycirc[myred] \mycirc[myred] \mycirc[myred] \mycirc[myred] \mycirc[myred],$\\
$\mysquare[myblue]\ \mysquare[myred]\ \mysquare[myblue]\ \mysquare[myblue]\ \mysquare[myred]\ \mysquare[myyellow] \rightarrow \mycirc[myblue] \mycirc[myred] \mycirc[myblue] \mycirc[myblue] \mycirc[myred] \mycirc[myyellow]$\\

\cellcolor{gray!20}\textbf{QUERY} (Lake and Baroni) \\
\begin{tabular}{c c c}
\textbf{IN} & \textbf{OUT} & \textbf{\#} \\
\hline
$\mysquare[myred]\ \mysquare[myblue]$
&
\begin{tabular}{c}
\cellcolor{green!20}$\mycirc[myred] \mycirc[myblue]$ \\
\end{tabular}
&
\begin{tabular}{c}
$10$
\end{tabular}
\\
\hline
$\mysquare[myblue]\ \mysquare[myblue]$
&
\begin{tabular}{c}
\cellcolor{green!20}$\mycirc[myblue] \mycirc[myblue]$ \\
\end{tabular}
&
\begin{tabular}{c}
$10$
\end{tabular}
\\
\hline
$\mysquare[myyellow]$
&
\begin{tabular}{c}
\cellcolor{green!20}$\mycirc[myyellow]$ \\
$\mycirc[myblue]$ \\
\end{tabular}
&
\begin{tabular}{c}
$9$ \\$1$
\end{tabular}
\\
\hline
$\mysquare[myyellow]\ \mathtt{four\ times}$
&
\begin{tabular}{c}
\cellcolor{green!20}$\mycirc[myyellow] \mycirc[myyellow] \mycirc[myyellow] \mycirc[myyellow]$ \\
$\mycirc[mypurple] \mycirc[myred]$ \\
\end{tabular}
&
\begin{tabular}{c}
$9$ \\$1$
\end{tabular}
\\
\hline
$\mysquare[myyellow]\ \mathtt{twice}$
&
\begin{tabular}{c}
\cellcolor{green!20}$\mycirc[myyellow] \mycirc[myyellow]$ \\
\end{tabular}
&
\begin{tabular}{c}
$10$
\end{tabular}
\\
\end{tabular}
\end{tabular}
\end{minipage}
&
\begin{minipage}[t]{\linewidth}
\vspace{0pt}
\centering
\begin{tabular}{c}
\cellcolor{gray!20}\textbf{QUERY} (Lake and Baroni) \\
\begin{tabular}{c c c}
\textbf{IN} & \textbf{OUT} & \textbf{\#} \\
\hline
$\mysquare[myred]\ \mysquare[myred]\ \mathtt{four\ times}$
&
\begin{tabular}{c}
$\mycirc[myred] \mycirc[myred] \mycirc[myred] \mycirc[myred] \mycirc[myred]$ \\
\cellcolor{green!20}$\mycirc[myred] \mycirc[myred] \mycirc[myred] \mycirc[myred] \mycirc[myred] \mycirc[myred] \mycirc[myred] \mycirc[myred]$ \\
\end{tabular}
&
\begin{tabular}{c}
$9$ \\$1$
\end{tabular}
\\
\hline
$\mysquare[myred]\ \mysquare[myred]\ \mathtt{twice\ within}\ \mysquare[myyellow]\ \mysquare[myblue]$
&
\begin{tabular}{c}
$\mycirc[myyellow] \mycirc[myblue] \mycirc[myyellow] \mycirc[myblue] \mycirc[myred] \mycirc[myred]$ \\
$\mycirc[myred] \mycirc[myred] \mycirc[mygreen] \mycirc[myyellow] \mycirc[myblue]$ \\
$\mycirc[myred] \mycirc[myred] \mycirc[mypink] \mycirc[mypurple] \mycirc[myblue]$ \\
$\mycirc[myred] \mycirc[myred] \mycirc[myyellow] \mycirc[myblue] \mycirc[myyellow] \mycirc[myblue]$ \\
$\mycirc[myred] \mycirc[myyellow] \mycirc[myblue] \mycirc[myyellow] \mycirc[myblue] \mycirc[myred] \mycirc[myred]$ \\
$\mycirc[myred] \mycirc[myyellow] \mycirc[myblue] \mycirc[myyellow] \mycirc[myblue] \mycirc[myred] \mycirc[myred] \mycirc[myred]$ \\
$\mycirc[myyellow] \mycirc[myblue] \mycirc[myred] \mycirc[myred] \mycirc[myred] \mycirc[myred]$ \\
$\mycirc[myyellow] \mycirc[myblue] \mycirc[myyellow] \mycirc[myblue] \mycirc[myred] \mycirc[myred] \mycirc[myred]$ \\
$\mycirc[myyellow] \mycirc[myblue] \mycirc[myyellow] \mycirc[myyellow] \mycirc[myblue] \mycirc[myblue] \mycirc[myred] \mycirc[myred]$ \\
\cellcolor{green!20}$\mycirc[myred] \mycirc[myyellow] \mycirc[myblue] \mycirc[myred] \mycirc[myred] \mycirc[myyellow] \mycirc[myblue]$ \\
\end{tabular}
&
\begin{tabular}{c}
$2$ \\$1$ \\$1$ \\$1$ \\$1$ \\$1$ \\$1$ \\$1$ \\$1$ \\$-$
\end{tabular}
\\
\hline
$\mysquare[mygreen]\ \mathtt{twice\ within}\ \mysquare[myblue]\ \mathtt{twice}$
&
\begin{tabular}{c}
$\mycirc[myblue] \mycirc[myblue] \mycirc[mygreen] \mycirc[myblue] \mycirc[myblue] \mycirc[myblue]$ \\
$\mycirc[myblue] \mycirc[mygreen] \mycirc[myblue] \mycirc[myblue] \mycirc[myblue] \mycirc[mygreen]$ \\
$\mycirc[myblue] \mycirc[mygreen] \mycirc[myblue] \mycirc[myblue] \mycirc[mygreen] \mycirc[myblue]$ \\
$\mycirc[myblue] \mycirc[mygreen] \mycirc[mygreen] \mycirc[myblue]$ \\
$\mycirc[mygreen] \mycirc[myblue] \mycirc[myblue] \mycirc[mygreen] \mycirc[myblue] \mycirc[myblue]$ \\
$\mycirc[mygreen] \mycirc[myblue] \mycirc[myblue] \mycirc[mygreen] \mycirc[mygreen] \mycirc[myblue]$ \\
$\mycirc[mygreen] \mycirc[mygreen] \mycirc[mygreen] \mycirc[myblue] \mycirc[mygreen]$ \\
$\mycirc[mygreen] \mycirc[mygreen] \mycirc[mygreen] \mycirc[myblue] \mycirc[mygreen] \mycirc[mygreen]$ \\
$\mycirc[mygreen] \mycirc[mypink] \mycirc[myblue] \mycirc[myblue]$ \\
$\mycirc[mygreen] \mycirc[mypink] \mycirc[myblue] \mycirc[mypurple]$ \\
\cellcolor{green!20}$\mycirc[myblue] \mycirc[myblue] \mycirc[mygreen] \mycirc[mygreen] \mycirc[myblue] \mycirc[myblue]$ \\
\end{tabular}
&
\begin{tabular}{c}
$1$ \\$1$ \\$1$ \\$1$ \\$1$ \\$1$ \\$1$ \\$1$ \\$1$ \\$1$ \\$-$
\end{tabular}
\\
\hline
$\mysquare[myblue]\ \mathtt{twice\ within}\ \mysquare[myred]$
&
\begin{tabular}{c}
$\mycirc[myblue] \mycirc[myblue] \mycirc[myblue] \mycirc[myred]$ \\
$\mycirc[myred] \mycirc[myred] \mycirc[myblue] \mycirc[myred]$ \\
$\mycirc[myblue] \mycirc[myblue] \mycirc[myblue] \mycirc[myred] \mycirc[myred]$ \\
$\mycirc[myblue] \mycirc[myblue] \mycirc[myred] \mycirc[myblue]$ \\
$\mycirc[myblue] \mycirc[myred]$ \\
$\mycirc[myblue] \mycirc[myyellow] \mycirc[myred]$ \\
$\mycirc[myred] \mycirc[myblue]$ \\
$\mycirc[myred] \mycirc[myblue] \mycirc[myred] \mycirc[myblue]$ \\
\cellcolor{green!20}$\mycirc[myred] \mycirc[myblue] \mycirc[myblue] \mycirc[myred]$ \\
\end{tabular}
&
\begin{tabular}{c}
$2$ \\$2$ \\$1$ \\$1$ \\$1$ \\$1$ \\$1$ \\$1$ \\$-$
\end{tabular}
\\
\hline
$\mysquare[myyellow]\ \mysquare[myyellow]\ \mathtt{four\ times}$
&
\begin{tabular}{c}
$\mycirc[myyellow] \mycirc[myyellow] \mycirc[myyellow] \mycirc[myyellow] \mycirc[myyellow]$ \\
\cellcolor{green!20}$\mycirc[myyellow] \mycirc[myyellow] \mycirc[myyellow] \mycirc[myyellow] \mycirc[myyellow] \mycirc[myyellow] \mycirc[myyellow] \mycirc[myyellow]$ \\
\end{tabular}
&
\begin{tabular}{c}
$6$ \\$4$
\end{tabular}
\\
\end{tabular}
\end{tabular}
\end{minipage}

\\
\end{tabularx}
\end{sidewaystable*}


\begin{sidewaystable*}[p]\footnotesize
\centering
\caption{\textbf{Responses of our Transformer to our own query examples for Episode \#1.}
Grammar, decoding and support set (left) and query evaluations (right). Expected outputs are highlighted in green.}
\label{tab:ep1extSUPPORT}

\begin{tabularx}{\textwidth}{@{}p{0.42\textwidth} p{0.56\textwidth}@{}}

\begin{minipage}[t]{\linewidth}
\vspace{0pt}
\centering
\begin{tabular}{c}
\cellcolor{gray!20}\textbf{GRAMMAR \#1} (Lake and Baroni) \\
$\mathrm{tufa} \rightarrow \mycirc[myyellow],\ 
\mathrm{wif} \rightarrow \mycirc[myblue],\ 
\mathrm{lug} \rightarrow \mycirc[mypurple],\ 
\mathrm{fep} \rightarrow \mycirc[myred],$ \\
$x_1\ \mathrm{gazzer}\ \rightarrow x_1\ x_1\ x_1,$\\
$x_1\ \mathrm{kiki}\ u_1 \rightarrow x_1\ u_1\ x_1,$\\
$x_1\ \mathrm{zup} \rightarrow x_1\ x_1$ \\
\cellcolor{gray!20}\textbf{DECODING} (this paper; for readability) \\
$\mathrm{tufa} : \mysquare[myyellow],\ 
\mathrm{wif} : \mysquare[myblue],\ 
\mathrm{lug} : \mysquare[mypurple],\ 
\mathrm{fep} : \mysquare[myred],$ \\
$\mathrm{gazzer} : \mathtt{thrice},\ 
\mathrm{kiki} : \mathtt{around},\ 
\mathrm{zup} : \mathtt{twice}$ \\
\cellcolor{gray!20}\textbf{SUPPORT} (Lake and Baroni) \\
$\mysquare[myblue] \rightarrow \mycirc[myblue],\ 
\mysquare[mypurple] \rightarrow \mycirc[mypurple],$\\
$\mysquare[myyellow]\ \mysquare[myyellow] \rightarrow \mycirc[myyellow] \mycirc[myyellow],$\\
$\mysquare[myred]\ \mathtt{twice} \rightarrow \mycirc[myred] \mycirc[myred],$\\
$\mysquare[myred]\ \mathtt{thrice} \rightarrow \mycirc[myred] \mycirc[myred] \mycirc[myred],$\\
$\mysquare[myred]\ \mysquare[myblue]\ \mathtt{twice} \rightarrow \mycirc[myred] \mycirc[myblue] \mycirc[myred] \mycirc[myblue],$\\
$\mysquare[myred]\ \mysquare[myred]\ \mathtt{thrice} \rightarrow \mycirc[myred] \mycirc[myred] \mycirc[myred] \mycirc[myred] \mycirc[myred] \mycirc[myred],$\\
$\mysquare[mypurple]\ \mysquare[myred]\ \mathtt{twice} \rightarrow \mycirc[mypurple] \mycirc[myred] \mycirc[mypurple] \mycirc[myred],$\\
$\mysquare[myyellow]\ \mysquare[myred]\ \mathtt{thrice} \rightarrow \mycirc[myyellow] \mycirc[myred] \mycirc[myyellow] \mycirc[myred] \mycirc[myyellow] \mycirc[myred],$\\
$\mysquare[myblue]\ \mathtt{around}\ \mysquare[mypurple] \rightarrow \mycirc[myblue] \mycirc[mypurple] \mycirc[myblue],$\\
$\mysquare[myyellow]\ \mathtt{around}\ \mysquare[myyellow] \rightarrow \mycirc[myyellow] \mycirc[myyellow] \mycirc[myyellow],$\\
$\mysquare[myred]\ \mathtt{around}\ \mysquare[myred]\ \mathtt{around}\ \mysquare[myblue] \rightarrow \mycirc[myred] \mycirc[myred] \mycirc[myred] \mycirc[myblue] \mycirc[myred] \mycirc[myred] \mycirc[myred],$\\
$\mysquare[myyellow]\ \mathtt{around}\ \mysquare[myblue]\ \mathtt{twice} \rightarrow \mycirc[myyellow] \mycirc[myblue] \mycirc[myyellow] \mycirc[myyellow] \mycirc[myblue] \mycirc[myyellow],$\\
$\mysquare[myblue]\ \mathtt{thrice}\ \mathtt{around}\ \mysquare[mypurple] \rightarrow \mycirc[myblue] \mycirc[myblue] \mycirc[myblue] \mycirc[mypurple] \mycirc[myblue] \mycirc[myblue] \mycirc[myblue]$\\
\end{tabular}
\end{minipage}
&
\begin{minipage}[t]{\linewidth}
\vspace{0pt}
\centering
\begin{tabular}{c}
\cellcolor{gray!20}\textbf{QUERY} (new; this paper) \\
\begin{tabular}{c c c}
\textbf{IN} & \textbf{OUT} & \textbf{\#} \\
\hline

$\mysquare[myred]\ \mathtt{thrice}\ \mathtt{around}\ \mysquare[myblue]\ \mysquare[mypurple]$
&
\begin{tabular}{c}
$\mycirc[myred] \mycirc[myred] \mycirc[myred] \mycirc[myblue] \mycirc[mypurple] \mycirc[myred] \mycirc[myred] \mycirc[myred]$\\
\cellcolor{green!20}$\mycirc[myred] \mycirc[myred] \mycirc[myred] \mycirc[myblue] \mycirc[myred] \mycirc[myred] \mycirc[myred] \mycirc[mypurple]$\\
\end{tabular}
&
\begin{tabular}{c}
$10$\\$-$
\end{tabular}
\\
\hline

$\mysquare[myred]\ \mathtt{thrice}\ \mathtt{around}\ \mysquare[myblue]\ \mysquare[myyellow]$
&
\begin{tabular}{c}
$\mycirc[myred] \mycirc[myred] \mycirc[myred] \mycirc[myblue] \mycirc[myyellow] \mycirc[myred] \mycirc[myred] \mycirc[myred]$\\
$\mycirc[mygreen] \mycirc[mygreen] \mycirc[myblue] \mycirc[myblue] \mycirc[myblue]$\\
$\mycirc[myred] \mycirc[myred]$\\
\cellcolor{green!20}$\mycirc[myred] \mycirc[myred] \mycirc[myred] \mycirc[myblue] \mycirc[myred] \mycirc[myred] \mycirc[myred] \mycirc[myyellow]$\\
\end{tabular}
&
\begin{tabular}{c}
$8$\\$1$\\$1$\\$-$
\end{tabular}
\\
\hline

$\mysquare[myred]\ \mathtt{thrice}\ \mathtt{around}\ \mysquare[myblue]\ \mysquare[myblue]$
&
\begin{tabular}{c}
$\mycirc[myred] \mycirc[myred] \mycirc[myred] \mycirc[myblue] \mycirc[myblue] \mycirc[myred] \mycirc[myred] \mycirc[myred]$\\
$\mycirc[myblue] \mycirc[myblue] \mycirc[myblue] \mycirc[myred] \mycirc[myblue] \mycirc[myblue] \mycirc[myblue]$\\
$\mycirc[mygreen] \mycirc[mypurple] \mycirc[mypurple] \mycirc[myblue] \mycirc[myblue]$\\
\cellcolor{green!20}$\mycirc[myred] \mycirc[myred] \mycirc[myred] \mycirc[myblue] \mycirc[myred] \mycirc[myred] \mycirc[myred] \mycirc[myblue]$\\
\end{tabular}
&
\begin{tabular}{c}
$8$\\$1$\\$1$\\$-$
\end{tabular}
\\
\hline

$\mysquare[myblue]\ \mathtt{thrice}\ \mathtt{around}\ \mysquare[myyellow]\ \mysquare[myred]$
&
\begin{tabular}{c}
$\mycirc[myblue] \mycirc[myblue] \mycirc[myblue] \mycirc[myyellow] \mycirc[myred] \mycirc[myblue] \mycirc[myblue] \mycirc[myblue]$\\
$\mycirc[myblue] \mycirc[myblue] \mycirc[myblue] \mycirc[myyellow] \mycirc[myblue] \mycirc[myblue] \mycirc[myblue]$\\
\cellcolor{green!20}$\mycirc[myblue] \mycirc[myblue] \mycirc[myblue] \mycirc[myyellow] \mycirc[myblue] \mycirc[myblue] \mycirc[myblue] \mycirc[myred]$\\
\end{tabular}
&
\begin{tabular}{c}
$9$\\$1$\\$-$
\end{tabular}
\\
\hline

$\mysquare[myblue]\ \mathtt{thrice}\ \mathtt{around}\ \mysquare[mypurple]\ \mysquare[myblue]$
&
\begin{tabular}{c}
$\mycirc[myblue] \mycirc[myblue] \mycirc[myblue] \mycirc[mypurple] \mycirc[myblue] \mycirc[myblue] \mycirc[myblue]$\\
$\mycirc[myblue] \mycirc[mygreen] \mycirc[mygreen] \mycirc[mypurple] \mycirc[myblue]$\\
$\mycirc[myblue] \mycirc[myyellow] \mycirc[myyellow] \mycirc[mypurple] \mycirc[myblue]$\\
\cellcolor{green!20}$\mycirc[myblue] \mycirc[myblue] \mycirc[myblue] \mycirc[mypurple] \mycirc[myblue] \mycirc[myblue] \mycirc[myblue] \mycirc[myblue]$\\
\end{tabular}
&
\begin{tabular}{c}
$8$\\$1$\\$1$\\$-$
\end{tabular}
\\
\hline

$\mysquare[myblue]\ \mathtt{thrice}\ \mathtt{around}\ \mysquare[myblue]\ \mysquare[myblue]$
&
\begin{tabular}{c}
$\mycirc[myblue] \mycirc[myblue] \mycirc[myblue] \mycirc[myblue] \mycirc[myblue] \mycirc[myblue] \mycirc[myblue]$\\
$\mycirc[myblue] \mycirc[myblue]$\\
\cellcolor{green!20}$\mycirc[myblue] \mycirc[myblue] \mycirc[myblue] \mycirc[myblue] \mycirc[myblue] \mycirc[myblue] \mycirc[myblue] \mycirc[myblue]$\\
\end{tabular}
&
\begin{tabular}{c}
$9$\\$1$\\$-$
\end{tabular}
\\
\hline

$\mysquare[myred]\ \mathtt{around}\ \mysquare[myblue]\ \mysquare[mypurple]\ \mathtt{twice}$
&
\begin{tabular}{c}
$\mycirc[myred] \mycirc[myblue] \mycirc[mypurple] \mycirc[myred] \mycirc[myred] \mycirc[myblue] \mycirc[mypurple] \mycirc[myred]$\\
$\mycirc[mypink] \mycirc[myred] \mycirc[myblue] \mycirc[mypurple] \mycirc[mypurple]$\\
$\mycirc[mypurple] \mycirc[myred] \mycirc[myblue] \mycirc[mypurple] \mycirc[mygreen]$\\
\cellcolor{green!20}$\mycirc[myred] \mycirc[myblue] \mycirc[myred] \mycirc[mypurple] \mycirc[myred] \mycirc[myblue] \mycirc[myred] \mycirc[mypurple]$\\
\end{tabular}
&
\begin{tabular}{c}
$8$\\$1$\\$1$\\$-$
\end{tabular}
\\
\hline

$\mysquare[myred]\ \mathtt{around}\ \mysquare[myblue]\ \mysquare[myblue]\ \mathtt{twice}$
&
\begin{tabular}{c}
$\mycirc[myred] \mycirc[myblue] \mycirc[myblue] \mycirc[myred] \mycirc[myred] \mycirc[myblue] \mycirc[myblue] \mycirc[myred]$\\
$\mycirc[myblue] \mycirc[myblue] \mycirc[myred] \mycirc[myblue] \mycirc[myblue] \mycirc[myred]$\\
$\mycirc[mypink] \mycirc[myred] \mycirc[myblue] \mycirc[myblue]$\\
\cellcolor{green!20}$\mycirc[myred] \mycirc[myblue] \mycirc[myred] \mycirc[myblue] \mycirc[myred] \mycirc[myblue] \mycirc[myred] \mycirc[myblue]$\\
\end{tabular}
&
\begin{tabular}{c}
$8$\\$1$\\$1$\\$-$
\end{tabular}
\\
\hline

$\mysquare[myblue]\ \mathtt{around}\ \mysquare[myred]\ \mysquare[myblue]\ \mathtt{twice}$
&
\begin{tabular}{c}
\cellcolor{green!20}$\mycirc[myblue] \mycirc[myred] \mycirc[myblue] \mycirc[myblue] \mycirc[myblue] \mycirc[myred] \mycirc[myblue] \mycirc[myblue]$\\
$\mycirc[myblue] \mycirc[myred] \mycirc[myblue] \mycirc[myblue] \mycirc[myblue]$\\
$\mycirc[myblue] \mycirc[myred] \mycirc[myblue] \mycirc[myblue] \mycirc[myred] \mycirc[myblue]$\\
$\mycirc[myblue] \mycirc[myyellow] \mycirc[mygreen] \mycirc[myblue] \mycirc[mypurple]$\\
$\mycirc[myred] \mycirc[myblue] \mycirc[myblue] \mycirc[myblue] \mycirc[myred] \mycirc[myblue] \mycirc[myblue] \mycirc[myblue]$\\
\end{tabular}
&
\begin{tabular}{c}
$6$\\$1$\\$1$\\$1$\\$1$
\end{tabular}
\\
\hline

$\mysquare[myblue]\ \mathtt{around}\ \mysquare[myblue]\ \mysquare[myred]\ \mathtt{twice}$
&
\begin{tabular}{c}
$\mycirc[myblue] \mycirc[myblue] \mycirc[myred] \mycirc[myblue] \mycirc[myblue] \mycirc[myblue] \mycirc[myred] \mycirc[myblue]$\\
$\mycirc[myblue] \mycirc[myred] \mycirc[myblue] \mycirc[mygreen] \mycirc[myblue]$\\
$\mycirc[myblue] \mycirc[myred] \mycirc[myblue] \mycirc[myred] \mycirc[myblue] \mycirc[myblue]$\\
$\mycirc[myblue] \mycirc[myyellow] \mycirc[myblue] \mycirc[mypink] \mycirc[mypurple]$\\
\cellcolor{green!20}$\mycirc[myblue] \mycirc[myblue] \mycirc[myblue] \mycirc[myred] \mycirc[myblue] \mycirc[myblue] \mycirc[myblue] \mycirc[myred]$\\
\end{tabular}
&
\begin{tabular}{c}
$7$\\$1$\\$1$\\$1$\\$-$
\end{tabular}
\\
\hline

$\mysquare[myblue]\ \mathtt{around}\ \mysquare[myblue]\ \mysquare[myblue]\ \mathtt{twice}$
&
\begin{tabular}{c}
$\mycirc[myblue] \mycirc[myblue] \mycirc[myblue] \mycirc[myblue] \mycirc[myblue] \mycirc[myblue]$\\
\cellcolor{green!20}$\mycirc[myblue] \mycirc[myblue] \mycirc[myblue] \mycirc[myblue] \mycirc[myblue] \mycirc[myblue] \mycirc[myblue] \mycirc[myblue]$\\
$\mycirc[myblue] \mycirc[myblue] \mycirc[myblue] \mycirc[myblue] \mycirc[myblue] \mycirc[myblue] \mycirc[myblue]$\\
$\mycirc[myblue] \mycirc[myblue] \mycirc[myblue] \mycirc[myblue] \mycirc[myyellow]$\\
$\mycirc[myblue] \mycirc[mypurple] \mycirc[myblue] \mycirc[myblue] \mycirc[mypurple]$\\
\end{tabular}
&
\begin{tabular}{c}
$5$\\$2$\\$1$\\$1$\\$1$
\end{tabular}
\\

\end{tabular}
\end{tabular}
\end{minipage}

\\
\end{tabularx}
\end{sidewaystable*}
\clearpage
\section{OOT testing using dataset-inherent statistics} \label{secC}

This section specifies the constraints under which out-of-training (OOT) episodes were generated and reports what the generator returned under each. Episodes are drawn by the search-based procedure of Section~\ref{secD} (Algorithm~\ref{alg:generate-episode}) from the meta-grammar class used in training with the values stated for each family. We draw grammars with no soft-constraint fallback and use them to generate support sets under constraints $C_s$ and query sets under $C_q$ over inputs absent from the support inputs. The grammar is unseen during training, and sampled $(x,y)$ pairs are unique within an episode and grammar-consistent. The primitive-only control retains repeated query draws and matches to support inputs. Unless stated otherwise, we request $500$ episodes per cell, each with $n_q=10$ query items, and accept an episode when it yields at least $n_q^{\min}$ of them.

All five checkpoints, our Transformer, our LSTM and the three released checkpoints of ref.~\cite{lake2023human} (\emph{MLC(algebraic+biases)}, \emph{MLC(algebraic only)} and \emph{MLC(joint)}), are scored on the same episodes by greedy decoding. Accuracy is computed over all query items of a cell and retrieval over all support items. Unless stated otherwise, accuracy is computed over all query items of a cell and retrieval over all support items. Section~\ref{sec:C_within} distinguishes the full within-grammar evaluation from the matched comparison restricted to function applications. By default, confidence intervals are $95\%$ percentile bootstraps over $10{,}000$ resamples of episodes. For the within-grammar verification, we resample grammars, retaining their paired lengths together in the matched comparison. 

\subsection{Constraint specification}
Each evaluation is specified by a configuration $C=(\theta, C_s, C_q)$. The meta-grammar parameters $\theta$ determine the admissible grammars. They comprise the input and output symbol inventories $\mathcal{I}$ and $\mathcal{O}$, the number of primitives $n_p$ and of function rules $n_f$, the bounds $[\min_f, \max_f]$ on the number of references a function places in its right-hand side, and the admissible left-hand-side forms $\mathcal{T}$. The support constraints $C_s$ and the query constraints $C_q$ determine the admissible items. They impose hard bounds on the input length $|x|$, the output length $|y|$ and the nesting degree $d$, and may in addition pin the number of function or concatenation applications, or the number of distinct function symbols (Extended Data Table~\ref{tab:unified_parameters}). The two sets of item constraints can be varied independently of each other, whereas $\theta$ applies to the episode as a whole, since support and query items are generated from the same grammar. Unless stated otherwise, the support bounds remain at the released defaults. Items outside the bounds are rejected. Under the exact-fill flag, the sampler additionally biases expansion of the input towards the pinned query length or range (Section~\ref{sec:F_constrained}). The symbol inventory, the nesting sweep and the within-grammar verification rely on rejection alone. We vary specified axes at a time and hold the remaining bounds at their defaults.

\begin{sidewaystable}[p]
\caption{\textbf{Episodes requested and obtained for parameter sweeps and associated controls.} Each family with the values pinned or varied
in it, the cells requested, the episodes requested per cell, the query items requested per episode ($n_q$)
and, in parentheses, the number of them at which an episode is accepted ($n_q^{\min}$); $n_q^{\min} = n_q$, the
released setting, accepts an episode only when all its query items are generated. Unless stated otherwise in the text, bounds not listed stay at the released defaults, $|x|, |y| \leq 8$, $d \leq 8$ and
$|\mathrm{RHS}| \in [2, 8]$. The last column splits the cells into those that returned every episode with all its
queries, those short of either, and those that returned none. Indented rows are variants of the family above them.}
\label{tab:supp_generation_summary}
\footnotesize
\begin{tabular*}{\textheight}{@{\extracolsep{\fill}} l l r r c c @{}}
\toprule
Family & Pinned or varied & Cells & Episodes & \shortstack{$n_q$\\($n_q^{\min}$)} & \shortstack{Complete /\\partial / none} \\
\midrule
Symbol inventory & $n_p\,{=}\,2$--$6$, $n_f \leq 9\!-\!n_p$ & 30 & 500 & 10 (10) & 30 / 0 / 0 \\
Input length & $|x|\,{=}\,2$--$12$; $|y| \leq 8$; $d$ free & 11 & 500 & 10 (4) & 7 / 4 / 0 \\
Output length & $|y|\,{=}\,2$--$12$; $|x| \leq 8$; $d$ free & 11 & 500 & 10 (4) & 3 / 8 / 0 \\
\addlinespace
Length plane & $|x|\,{=}\,1$--$11$ and $|y|\,{=}\,1$--$12$ jointly & 132 & 500 & 5 (2) & 50 / 36 / 46 \\
\quad below the output floor & $|\mathrm{RHS}|\,{\in}\,[1,2]$ & 36 & 500 & 10 (10) & 29 / 3 / 4 \\
\quad the remaining seven cells\footnotemark[1] & $n_q^{\min}\,{=}\,4$, then erasing RHS sampler & 7 & 500 & 10 (4) & 4 / 3 / 0 \\
\quad $(2,9)$--$(2,12)$, $(3,11)$\footnotemark[1] & atom-slot rule; $|\mathrm{RHS}|\,{=}\,|y|$ & 5 & 500 & 10 (4) & 1 / 4 / 0 \\
\quad $|x|\,{=}\,12$, $|y|\,{=}\,7$--$9$\footnotemark[1] & same joint pin, released support & 3 & 50 & 10 (10) & 3 / 0 / 0 \\
\quad $|x|\,{=}\,1$ & held-out primitive value, width $1$--$12$ & 12 & 1{,}000 & 1 (1) & 12 / 0 / 0 \\
\addlinespace
Argument widths & $w_1\!+\!w_2 \leq 8$; injected reversing rule & 28 & 500 & 10 (10) & 28 / 0 / 0 \\
\addlinespace
Nesting degree\footnotemark[1] & $d\,{=}\,2$--$12$; $|x| \leq d{+}1$, $|y| \leq d{+}6$; $|\mathrm{RHS}|\,{=}\,2$ & 11 & 500 & 10 (10) & 11 / 0 / 0 \\
\quad at fixed input length & $|x|\,{=}\,6$--$8$, $d\,{=}\,2$--$|x|$; $|\mathrm{RHS}|\,{=}\,2$ & 18 & 500 & 10 (4) & 10 / 4 / 4 \\
\addlinespace
RHS length & $\min_f\,{=}\,\max_f\,{=}\,r\,{=}\,1$--$8$; $|x|\,{\in}\,\{2,3\}$ & 8 & 500 & 10 (4) & 8 / 0 / 0 \\
\addlinespace
Primitive-only control & $n_f\,{=}\,3$, $n_p\,{=}\,4$; $|x|\,{=}\,2$--$10$ & 17 & 500 & 10 (10) & 17 / 0 / 0 \\
\addlinespace
Within-grammar verification & one grammar, fixed support; exact $|x|\,{=}\,2$--$12$ & 500 & 1 & 10 (10) & 320 / 180 / 0 \\
End-of-sequence probe & $|y| \leq 14$; $|\mathrm{RHS}|\,{\in}\,[2,6]$, $d \leq 6$; teacher forced & 1 & 500 & 10 (10) & 1 / 0 / 0 \\
\botrule
\end{tabular*}

\footnotetext{In the within-grammar row the $500$ entries are grammars rather than cells. A grammar counts as complete if it returned ten queries at every input length from $2$ to $12$, and as partial if it did so at some but not all lengths. The argument widths and primitive-only control construct queries. The latter retains repeated draws and support matches. The end-of-sequence probe is teacher forced and does not decode.}
\footnotetext[1]{Decoder cap raised from $12$ to $20$ tokens; on the within-grammar row it is $|x| + 16$, and on the $|x| = 12$ row it is recorded only at $|y| = 7$, no prediction there having exceeded eight tokens.}
\end{sidewaystable}

\subsection{Axes of variation}
\subsubsection{Symbol inventory size} \label{sec:C_inventory} 

\begin{figure}[!htbp]
  \centering
  \includegraphics[width=1\textwidth]{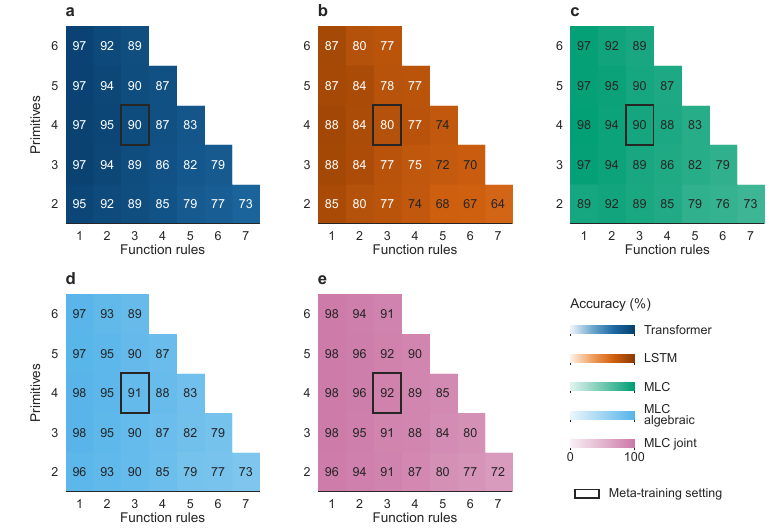}
   \caption{\textbf{Novel-query accuracy over the symbol inventory for the five checkpoints.} \textbf{a--e,} Mean accuracy on novel queries at every admissible $(n_p, n_f)$ with $n_f \geq 1$, pooled over the $5{,}000$ query items of each cell ($500$ episodes); the outlined cell is the meta-training setting. Grammars without functions, in which every query is a concatenation of primitives, are omitted.}
  \label{fig:supp_inventory_grid}
\end{figure}

We drew all $30$ combinations of $n_p \in \{2, \ldots, 6\}$ primitives and $n_f \in \{0, \ldots, 9 - n_p\}$ functions that the default symbol sets admit for inputs and outputs, at the released bounds for support and query. The symbol sets cannot be enlarged, since the architectures cannot process symbols they have not observed in training. Accuracy decreases as function rules are added to the inventory, but changes little with more primitives (Fig.~\ref{fig:supp_inventory_grid}). The support set is fixed at $14$ items, so each added rule leaves the others with fewer demonstrations. Across grammars at four primitives and queries in which a single rule fires, the median number of support items that apply that rule falls from $7$ with one function to $4$ with five functions. Moreover, at three functions, accuracy is $29\%$ when the support set never demonstrates the rule, $77\%$ with one demonstration and $89\%$ with two or more, over $59$, $92$ and $3{,}847$ queries. However, demonstrations do not account for the axis entirely. Even with two or more demonstrations, accuracy declines from $93\%$ at two functions to $84\%$ at five.

\subsubsection{RHS length}\label{sec:C_rhs} 

\begin{figure}[!htbp]
  \centering
  \includegraphics[width=1\textwidth]{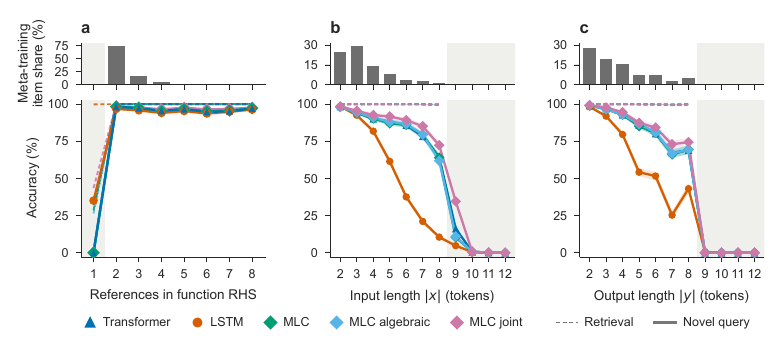}
   \caption{\textbf{The sampled axes of Fig.~\ref{fig:ood_testing}a--c for the five checkpoints.} \textbf{a,} References in the right-hand side of the function, over queries that apply it once, so that $|y| = r$. \textbf{b,} Input length. \textbf{c,} Output length. Novel queries are solid and retrieval dashed, $500$ episodes per cell. Bands are $95\%$ episode-clustered bootstrap intervals. The strip above each panel is the share of meta-training items at each value. Grey shading marks values beyond the meta-training range.}
  \label{fig:supp_sweeps_all_checkpoints}
\end{figure}

We set $\min_f = \max_f = r$ for $r \in \{1, \ldots, 8\}$ on both sets and pinned $|x| \in \{2, 3\}$ with $|y| \leq 8$, so that a single application yields $|y| = r$. The functions are drawn from the two atom forms $u\,F$ and $u\,F\,u$, whose arguments are single tokens ($u\,F$ alone at $r = 1$). Accuracy is computed over the single applications, $682$ to $897$ one-argument and $2{,}374$ to $2{,}809$ two-argument items per cell from $r = 2$ on and $1{,}821$ at $r = 1$, with concatenations excluded (Fig.~\ref{fig:supp_sweeps_all_checkpoints}a). From $r = 2$ to $8$ accuracy is flat at $94$ to $99\%$ for every checkpoint. Exposure in meta-training does not predict accuracy here: the share of meta-training items whose function has $r$ references falls from $74$ to $0.2\%$. At $r = 1$, where the output is shorter than the input, the Transformer and the released checkpoints solve no query and reproduce only $26$ to $44\%$ of the support items, whereas the LSTM solves $35\%$ while reproducing nearly all of them.

\subsubsection{Input and output length}\label{sec:C_input}\label{sec:C_output} 

\begin{figure}[!htbp]
  \centering
  \includegraphics[width=1\textwidth]{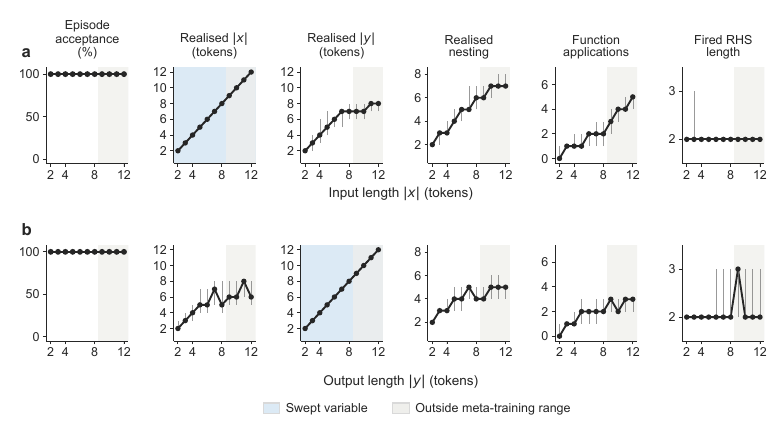}
  \caption{\textbf{Generation diagnostics for the one-dimensional length sweeps.} Rows are the input-length (\textbf{a}) and output-length (\textbf{b}) sweeps, $500$ episodes per grid cell. Columns give the acceptance rate (episodes returned per episode requested) and the realized query parameters. Markers give the median and error bars the interquartile range over the $4{,}762$ to $5{,}000$ query items of a cell. The tinted panel of each row is the swept parameter. Grey shading marks values beyond the meta-training range.}
  \label{fig:supp_generation_diagnostic}
\end{figure}

Query inputs and query outputs were each pinned to $\ell \in \{2, \ldots, 12\}$ in $C_q$. The other unpinned length was bounded by the released maximum of eight tokens and nesting degree was unbounded. The support set was held at $|x|, |y| \leq 8$, so that retrieval stays above $99\%$. On the input sweep, at $\ell \in \{11, 12\}$ the output must be shorter than the input. We bounded $|\mathrm{RHS}|$ to $[2, 3]$ for these two input lengths to reduce sampling time, since the sampler builds every candidate output in full before rejecting it. Among the $7{,}427$ items with $|y| \leq 8$ that the length plane returned at $|x| = 11$ under the released bounds, the function that fires took two arguments in every item and had two or three references in all but three.

Realized query parameters for both sweeps are given in Fig.~\ref{fig:supp_generation_diagnostic} and accuracy for all five checkpoints in Fig.~\ref{fig:supp_sweeps_all_checkpoints}. Every target on the input sweep lies inside the meta-training output range, so the collapse there is not explained by the emission window (Section~\ref{sec:C_window}), whereas on the output sweep every target beyond $\ell = 8$ lies outside that range.

\subsubsection{Length plane}\label{sec:C_plane}

\begin{figure}[!htbp]
  \centering
  \includegraphics[width=1\textwidth]{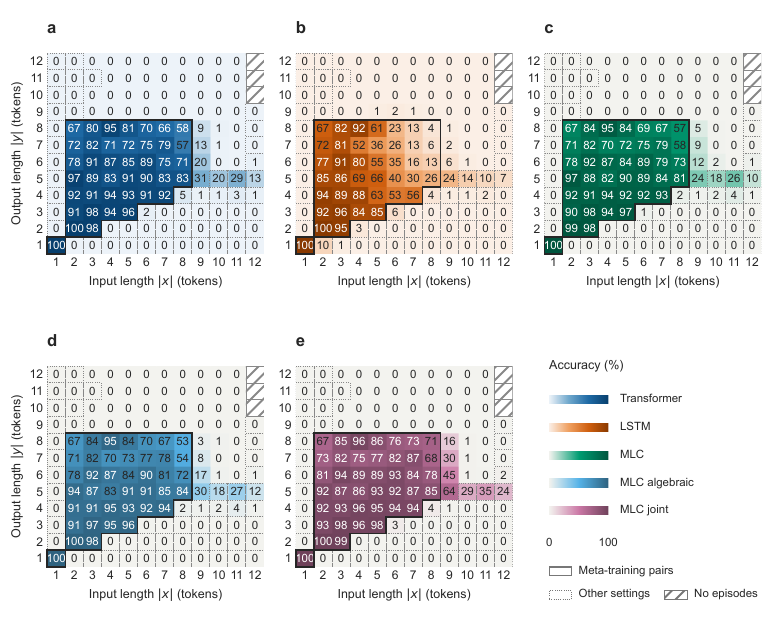}
  \caption{\textbf{Novel-query accuracy over the full length plane.} \textbf{a--e,} Mean accuracy on novel queries for the five checkpoints at every $(|x|, |y|)$ cell with $|x|, |y| \in \{1, \ldots, 12\}$, pooled over all query items of a cell. The outline traces the length pairs observed in meta-training. Cells with a dotted edge come from the four other generation designs. Hatched cells were not obtained.}
  \label{fig:supp_length_grid}
\end{figure}

We pinned the query input length $|x|$ to every value from $1$ to $11$ and the query output length $|y|$ to every value from $1$ to $12$, in all combinations, giving $132$ cells with $500$ episodes and five queries per episode requested for each. The nesting degree was bounded by $d \leq 8$ in both sets, and the support lengths by $\max(8, |x|, |y|)$, so that the support set can contain items as long as the queries. In cells with $|x| > 8$ or $|y| > 8$, $9$ to $15\%$ of the support items exceed eight tokens. However, those with more than eight output tokens are reproduced in $0\%$ of cases (LSTM $0.5\%$), and those whose input alone exceeds eight tokens in $74$ to $86\%$, whereas retrieval in every other cell is at least $99.8\%$. Forty-five cells of the plane cannot be generated under the released settings, and $|x| = 12$ lies beyond it. These cells were filled from four other designs, drawn with a dotted edge in Fig.~\ref{fig:supp_length_grid} and sized in Table~\ref{tab:supp_generation_summary}.
\begin{itemize}
\item \emph{Below the output floor}, where $|y| \leq \lfloor |x|/2 \rfloor$ ($30$ cells of the plane and six at $|x| = 12$). Every released function keeps all its arguments and has at least two references, so only a two-argument function can shorten the sequence, by at most one token per application, and at most $\lfloor (|x| - 1)/2 \rfloor$ such applications fit into $|x|$ tokens. We bounded $|\mathrm{RHS}|$ to $[1, 2]$, which filled $29$ of them. At $(2, 1)$ and $(11, 2)$, where this returned $13$ episodes and one, $n_q^{\min}$ was lowered from ten to four. The remaining five, $(9, 1)$ to $(12, 1)$ and $(12, 2)$, were drawn with the erasing variant of the RHS sampler, which gives every function rule exactly one reference, so that a two-argument rule drops an argument (Section~\ref{secD}).
\item \emph{At $|x| = 2$ with $|y| \geq 9$, and at $(3, 11)$}, the output requires a single function with more than the released maximum of eight references. We pinned $|\mathrm{RHS}|$ to $|y|$ and every function to a left-hand-side form whose arguments are bare primitives, one at $|x| = 2$ and two at $(3, 11)$, so that one application produces the output.
\item \emph{At $|x| = 1$}, the input is a single primitive, and every released primitive has a one-token output. This column comes from the held-out primitive values of the evaluation with injected rules, so that the value of a wider primitive must be inferred from the support items that contain it (Sections~\ref{secG} and~\ref{sec:F_queries}).
\item \emph{At $|x| = 12$}, the cells with $|y| = 7$ to $9$ come from a run of $50$ episodes per cell with ten queries each, which pinned both lengths with the support set at the released bounds and the nesting degree unbounded, as in the input-length sweep. Every checkpoint scores at most $0.2\%$ there, as at $(11, 7)$ to $(11, 9)$, so these settings do not change the reading. The cells $(12, 10)$ to $(12, 12)$ were not obtained.
\end{itemize}
Of the $87$ cells the plane attempted, $50$ returned all five queries in every episode, $36$ returned every episode but fewer than five queries in some, and only $(3, 11)$ returned none. At $(2, 7)$ and $(2, 8)$, where the one-argument function needed is admissible but rarely sampled, we allowed $2{,}000$ instead of $500$ grammar draws per episode. With $500$ draws, generation stopped after $22$ and $10$ episodes.

Accuracy declines with both lengths well inside the meta-training range (Fig.~\ref{fig:supp_length_grid}). All five checkpoints answer $99\%$ of the queries at $(2, 2)$. At $(7, 7)$ our Transformer answers $79\%$, the released checkpoints $78$ to $87\%$ and the LSTM $13\%$, and at $(8, 8)$ these fall to $58\%$, $53$ to $71\%$ and $4\%$. Outside the range the decline is abrupt. At $|x| = 9$ the Transformer answers $31\%$ at $|y| = 5$ and $9\%$ at $|y| = 8$. Beyond $|y| = 8$ only the LSTM answers any query, in at most $1.7\%$ of cases. From $|x| = 10$ on, accuracy stays below $5\%$ except along $|y| = 5$ (Section~\ref{sec:C_window}). Pinning the two lengths does not fix which rules the queries apply, so at fixed $|x|$ accuracy can rise with $|y|$. At $|x| = 4$ the Transformer answers $71\%$ at $|y| = 7$ but $95\%$ at $|y| = 8$, where $90\%$ of the queries apply a one-argument function to a multi-token argument, against none at $|y| = 7$.

Sampling does not ensure that the support set demonstrates the function a query applies, and a function with $r$ references is demonstrated only by support items with at least $r$ output tokens. At $|x| = 2$, where one application produces the whole output, $6\%$ of the queries apply an undemonstrated function at $|y| = 3$ and $4$ but $22$--$33\%$ at $|y| = 5$ to $8$. Every checkpoint answers $95$--$100\%$ of the demonstrated ones. For the others, the checkpoints mostly emit five identical tokens ($74$--$96\%$ of cases; LSTM $54\%$), which is correct only at $|y| = 5$, so accuracy stays at $85$--$97\%$ at $(2, 5)$.

\subsubsection{Argument widths of one function}\label{sec:C_widths} 

\begin{figure}[!htbp]
  \centering
  \includegraphics[width=1\textwidth]{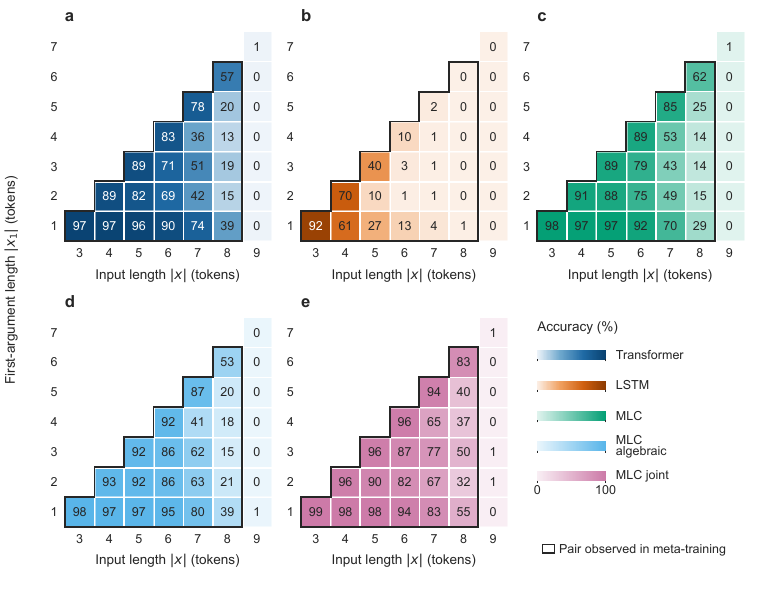}
  \caption{\textbf{Argument widths of one injected function for the five checkpoints.} \textbf{a--e,} Mean accuracy on constructed queries that apply a reversing function $x_1\,F\,x_2 \rightarrow [x_2]\,[x_1]$ to $w_1$ tokens on its left and $w_2$ on its right, arranged by input length $|x| = w_1 + w_2 + 1$ and the width of the first argument. Each cell pools the $4{,}940$ queries of the $494$ episodes whose support set demonstrates the operator. The outline encloses the width pairs that meta-training poses ($|x| \leq 8$).}
  \label{fig:supp_planes_all_checkpoints}
\end{figure}

A reversing function $x_1\,i\,x_2 \rightarrow [x_2]\,[x_1]$ was injected in place of one sampled two-argument function (see Methods, ``OOT Evaluation with injected rules''), and the support set was sampled at the released bounds. For every $(w_1, w_2)$ with $w_1 + w_2 \leq 8$ we constructed ten inputs $u_1 \ldots u_{w_1}\,F\,u_{w_1+1} \ldots u_{w_1+w_2}$ over the primitives the support set demonstrates. Accuracy is computed over the queries in which the injected function fires and where the support demonstrates the operator, resulting in $4{,}940$ per cell with $494$ of the $500$ episodes. All checkpoints show the same pattern (Fig.~\ref{fig:supp_planes_all_checkpoints}). Accuracy deteriorates as the input grows and is near zero at $|x| = 9$, beyond the meta-training range. At the same input length it also depends on how the tokens are divided between the two arguments, although the applied rule is the same. It is highest when one argument is a single token. At $|x| = 7$, for instance, our Transformer answers $74$ to $78\%$ of the queries with a width-one argument but $36$ to $51\%$ when both arguments are multi-token. The LSTM stays below $5\%$ from $|x| = 7$ on. The released checkpoints show the same gap. At $|x| = 8$, \emph{MLC(algebraic only)} answers $39$ to $53\%$ of the queries with a width-one argument but $15$ to $21\%$ when both arguments are multi-token. \emph{MLC(joint)}, the most accurate checkpoint at this length, answers $55$ to $83\%$ with a width-one argument and $32$ to $50\%$ when both arguments are multi-token.

\subsubsection{Nesting degree}\label{sec:C_nesting} 
\begin{figure}[!htbp]
  \centering
  \includegraphics[width=1\textwidth]{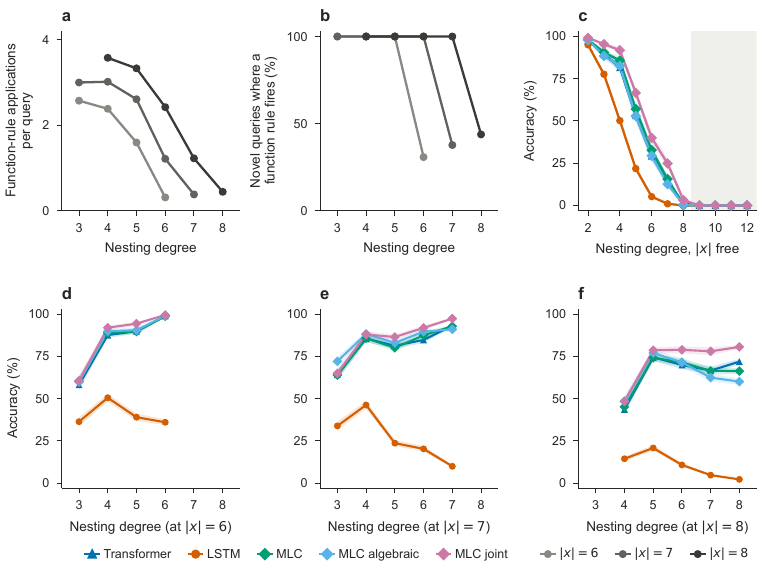}
  \caption{\textbf{At fixed input length, higher nesting means more concatenation.} \textbf{a,b,} Mean function-rule applications per novel query and percentage of queries applying a function, respectively, at $|x|=6$, $7$ and $8$. \textbf{c,} Novel-query accuracy with input length free, including targets longer than eight tokens. \textbf{d--f,} Accuracy at fixed $|x|=6$, $7$ and $8$ with $|y|\leq8$ ($500$ episodes per cell). Curves span degrees that returned episodes. Bands are $95\%$ episode-clustered bootstrap intervals; grey shading marks degrees beyond the meta-training range.}
  \label{fig:supp_nesting_fixed_x}
\end{figure}

Both designs vary the query nesting degree with $|\mathrm{RHS}|=2$ (Methods and Table~\ref{tab:supp_generation_summary}). Nesting degree counts concatenation as well as function applications, so it does not measure the depth of function composition. The first design pins $d\in\{2,\ldots,12\}$, with $|x|\leq d+1$ and $|y|\leq d+6$. Outputs exceed eight tokens in $20$ to $46\%$ of queries at $d=5$--$8$ and in all queries from $d=9$. Our Transformer falls from $98\%$ accuracy at $d=2$ to $0\%$ from $d=9$ (Fig.~\ref{fig:supp_nesting_fixed_x}c).

The second design jointly pins $|x|\in\{6,7,8\}$ and $d\in\{2,\ldots,|x|\}$, with $|y|\leq8$; retrieval is $99.6$ to $100\%$. Function applications fall from a mean of $2.6$--$3.6$ per query at the lowest degree to $0.3$--$0.4$ at $d=|x|$ (Fig.~\ref{fig:supp_nesting_fixed_x}a,b). Accuracy therefore rises, e.g., at $|x|=6$ from $58\%$ at $d=3$ to $99\%$ at $d=6$ (Fig.~\ref{fig:supp_nesting_fixed_x}d--f). LSTM accuracy still declines with nesting degree, consistent with its failure on long concatenations (Section~\ref{sec:C_primitives}).

\subsection{Emission window}\label{sec:C_window}
Across the sweeps, our Transformer, \emph{MLC(algebraic+biases)} and \emph{MLC(joint)} never exceed the meta-training maximum of eight output tokens. \emph{MLC(algebraic only)} exceeds it once and the LSTM in $0.09\%$ of predictions. Only ten predictions reach the decoding caps (Table~\ref{tab:supp_generation_summary}), so these do not explain the limit.

\begin{figure}[!htbp]
  \centering
  \includegraphics[width=\textwidth]{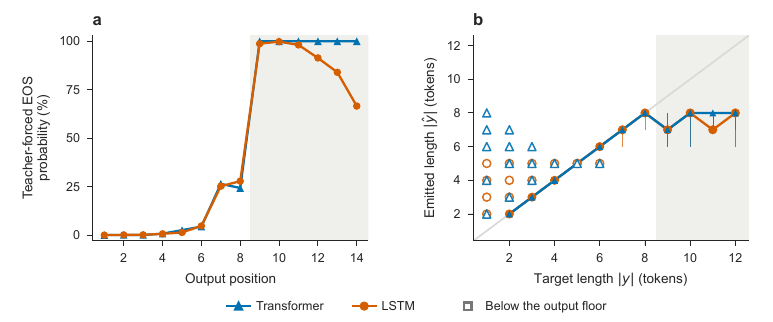}
  \caption{\textbf{Stopping behaviour and predicted output length.} \textbf{a,} Teacher-forced end-of-sequence probability by output position. \textbf{b,} Median emitted length and interquartile range in the output-length sweep ($4{,}762$--$5{,}000$ queries per target length); the diagonal marks equality. Open markers give medians for the $30$ cells below the output floor ($|y|\leq\lfloor|x|/2\rfloor$, $|x|\in\{2,\ldots,12\}$) with at least $100$ queries and $|\mathrm{RHS}|\in[1,2]$. Grey shading marks lengths beyond the meta-training range.}
  \label{fig:supp_emission_window}
\end{figure}

We generated $500$ episodes with $|\mathrm{RHS}|\in[2,6]$, $d\leq6$, support lengths $|x|,|y|\leq8$ and query outputs up to $14$ tokens, yielding $662$ queries with $|y|>8$. We recorded end-of-sequence probabilities at each position given the correct output prefix. Beyond position eight, where another target token was required, the mean was $1.00$ for the Transformer and $0.67$--$1.00$ for the LSTM (Fig.~\ref{fig:supp_emission_window}a).

For targets below the released output floor, both models often produced longer outputs than required (Fig.~\ref{fig:supp_emission_window}b). At input lengths ten to twelve, five-token outputs were most common overall, consistent with the residual accuracy for five-token targets (Section~\ref{sec:C_plane}).

\subsection{Primitive-only length control}\label{sec:C_primitives}
\begin{figure}[!htbp]
  \centering
  \includegraphics[width=1\textwidth]{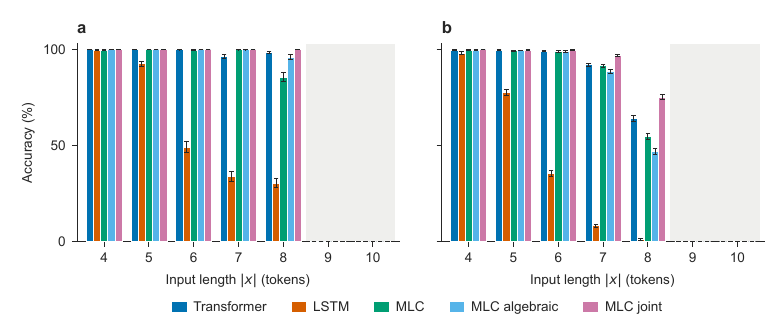}
  \caption{\textbf{Length generalization without function applications.} Accuracy for five checkpoints on constructed strings of \textbf{a,} one repeated primitive or \textbf{b,} several primitives with at least one repetition ($500$ episodes per condition). Both conditions share grammars and support at each length. Error bars show $95\%$ episode-clustered bootstrap intervals; grey shading marks lengths beyond meta-training.}
  \label{fig:ood_length_primitives}
\end{figure}
To separate sequence length from function application, we evaluated $500$ episodes per tested input length, using grammars with four primitives and three function rules and support at the released bounds. For each episode, we constructed ten query strings from primitives present in the support inputs. Strings either repeated one primitive ($|x|\in\{2,\ldots,10\}$) or combined several primitives with at least one repetition ($|x|\in\{3,\ldots,10\}$). The two conditions shared grammars and support sets. Queries were sampled with replacement, retaining duplicates and matches to support inputs. Targets concatenated the corresponding colours, so $|y|=|x|$ and no function was applied.

Retrieval remains near ceiling. At $|x|=6$, LSTM query accuracy is $49.2\%$ for repeated-primitive strings and $35.6\%$ for several-primitive strings; at $|x|=8$, the corresponding Transformer accuracies are $98.8\%$ and $64.0\%$ (Fig.~\ref{fig:ood_length_primitives}). All checkpoints score at most $0.1\%$ at lengths nine and ten, even without function applications.

\subsection{Within-grammar verification}\label{sec:C_within}

\begin{figure}[!htbp]
  \centering
\includegraphics[width=\textwidth]{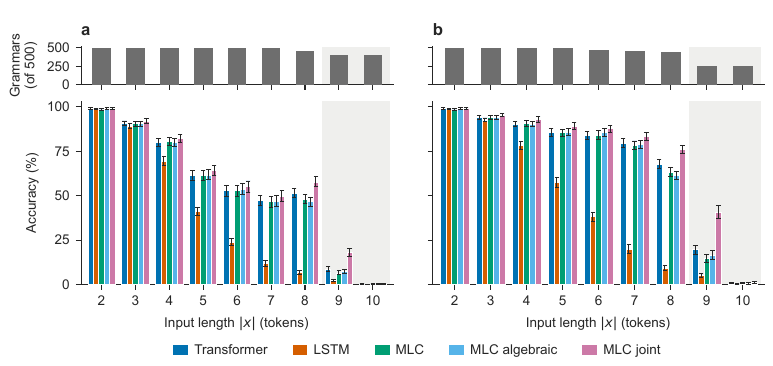}
\caption{\textbf{Length decline within fixed grammars.} Pooled novel-query accuracy for five checkpoints on $500$ grammars, each with a fixed $14$-item support set: \textbf{a,} all queries ($|y|\leq12$); \textbf{b,} queries with $|y|\leq8$. Strips show contributing grammars; error bars are $95\%$ grammar-clustered bootstrap intervals. Grey shading marks inputs beyond the meta-training range.}
\label{fig:supp_within_grammar}
\end{figure}

We fixed a $14$-item support set at the released bounds for each of $500$ novel grammars matching the meta-training configuration. At each $|x|\in\{2,\ldots,12\}$, we requested ten queries with inputs absent from support, with $|y|\leq12$ and $d$ unbounded, omitting grammar--length combinations returning fewer than ten. Figure~\ref{fig:supp_within_grammar} shows accuracy over all queries and those with $|y|\leq8$. Both include queries without function applications, and contributing grammars may vary with length.

For a matched comparison, we retained the $293$ grammars with at least one query that applies a function and has $|y|\leq8$ at every input length from three to eight. On these queries, accuracy falls from $92.7\%$ to $67.1\%$ for the Transformer and from $88.7\%$ to $14.5\%$ for the LSTM between $|x|=3$ and $8$. The decline therefore persists with grammar and support held fixed across lengths.

\clearpage
\section{Evaluation with injected rules}
\label{secG}

This section reports the conditions with injected rules and the checkpoints' responses to them (Algorithms~\ref{alg:inject-rule} and~\ref{alg:pose-queries}; see Methods, ``OOT Evaluation with injected rules''). The argument-width evaluation and the held-out primitive values used for the $|x|=1$ column of the length plane are reported in Section~\ref{secC}. All conditions are decoded greedily with a cap of $12$ tokens.

\subsection{Single-manipulation conditions}
\label{sec:G_overview}

\begin{figure}[p]
\centering
\includegraphics[width=\textwidth]{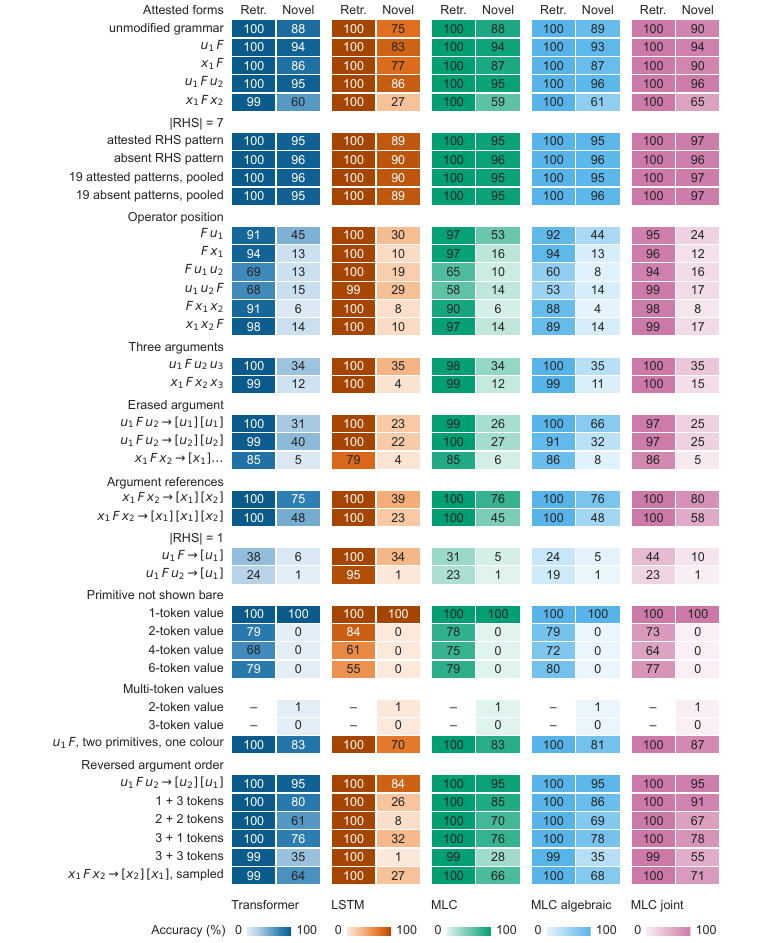}
\caption{\textbf{Retrieval and query accuracy under rule and primitive manipulations.} Paired cells show retrieval (left) and novel-query accuracy (right) for five checkpoints; $u_j$ denotes an atom slot, $x_j$ a string slot and $F$ the operator. Unspecified RHSs contain two to four references; token-count labels give argument widths for a reversing rule. Each condition uses $500$ episodes, except held-out values ($1{,}000$, one bare query each); pooled seven-reference rows combine $19$ conditions. Retrieval covers support items that apply the injected function (any function for the control) or, for held-out values, contain the primitive within the released output-length bound. Dashes mark conditions without a retrieval score.}
\label{fig:supp_battery_overview}
\end{figure}

Each condition changes one property of a function rule or a primitive mapping, including changes that lead outside the meta-grammar (see Methods, ``Controlled manipulations''). Rule conditions move the operator, add or erase an argument, or change the number or order of references on the RHS. Primitive conditions widen a primitive's value, give two primitives the same colour, or withhold a primitive's bare demonstration. Most rule conditions are tested with both atom slots, which bind one token, and with string slots, which bind a non-empty string. Fig.~\ref{fig:supp_battery_overview} groups the conditions by the property changed; its first row is a control without injection.

Novel-query accuracy is high for attested rule forms and falls for forms outside the meta-grammar (Fig.~\ref{fig:supp_battery_overview}). With atom slots, the five checkpoints reach $83$--$96\%$ on the attested unary and binary forms. Accuracy falls to $8$--$53\%$ when the operator is moved to the first or last position, $34$--$35\%$ with a third argument, $23$--$66\%$ when the second argument is erased and $1$--$34\%$ when the right-hand side has a single reference. In these comparisons, string-slot accuracy is never higher than the atom-slot accuracy of the same rule, and for our Transformer it is up to $36$ percentage points lower. The four Transformer checkpoints differ most on $u_1\,F\,u_2 \rightarrow [u_1]\,[u_1]$: \emph{MLC(algebraic only)} reaches $66\%$ accuracy and the other three $25$--$31\%$.

In three kinds of conditions, the checkpoints already fail on support examples (left cells in Fig.~\ref{fig:supp_battery_overview}). The four Transformer checkpoints reproduce only $19$--$44\%$ of the support examples of a rule that returns a single token, and three of them $53$--$69\%$ of those of a binary rule with a moved operator. All five checkpoints reproduce $40$--$84\%$ of the support examples that contain a held-out multi-token value. A support example with the same arguments as the query helps only for the operator-initial rules. Transformer accuracy on $F\,u_1$ is $49\%$ with such an example and $26\%$ without, and on $F\,u_1\,u_2$ it is $20\%$ and $11\%$, with similar gaps for the other checkpoints.

Bare queries apply the injected rule once to primitives (for example $a\,F\,b$), and for each rule they mostly receive the same incorrect response (Table~\ref{tab:supp_predicted_answers}). For the operator-initial rules, this response has one colour per input word: a colour at the operator's position, followed by the arguments' colours in input order ($76$--$95\%$ of bare queries). A response of this form could stem from the heuristic targets used in meta-training, some of which translate the input word by word. Yet it is just as frequent in \emph{MLC(algebraic only)}, which was meta-trained without them ($90$--$95\%$ against $76$--$95\%$). For the rules that erase an argument, the checkpoints return both arguments' colours, except \emph{MLC(algebraic only)}, which answers $86\%$ of the bare queries for $u_1\,F\,u_2\rightarrow[u_1]\,[u_1]$ correctly. The ternary rule has no dominant response and is not listed.


\begin{table}[!htbp]
\centering
\caption{\textbf{Most frequent incorrect response to bare queries.} For each injected rule, the response most often given instead of the target, the number of bare queries and the share of them that receive this response. $\bullet$ denotes one token of any colour.}
\label{tab:supp_predicted_answers}
\footnotesize
\setlength{\tabcolsep}{2pt}
\begin{tabular*}{\textwidth}{@{\extracolsep{\fill}} l l r c c c c c @{}}
\toprule
Injected rule & Response & \shortstack{Bare\\queries} & \shortstack{Transformer\\(ours)} & \shortstack{LSTM\\(ours)} & \shortstack{\emph{MLC}\\\emph{(algebraic+biases)}} & \shortstack{\emph{MLC}\\\emph{(algebraic only)}} & \shortstack{\emph{MLC}\\\emph{(joint)}} \\
\midrule
$F\,u_1$ & $\bullet\,[u_1]$ & $643$ & $87\%$ & $92\%$ & $80\%$ & $90\%$ & $95\%$ \\
$F\,u_1\,u_2$ & $\bullet\,[u_1]\,[u_2]$ & $2{,}014$ & $94\%$ & $76\%$ & $93\%$ & $95\%$ & $92\%$ \\
$u_1\,u_2\,F$ & $[u_1]\,[u_2]\,[u_2]$ & $2{,}014$ & $80\%$ & $32\%$ & $81\%$ & $81\%$ & $61\%$ \\
$u_1\,F\,u_2\rightarrow[u_1]\,[u_1]$ & $[u_1]\,[u_2]$ & $1{,}896$ & $65\%$ & $84\%$ & $98\%$ & $8\%$ & $97\%$ \\
$u_1\,F\rightarrow[u_1]$ & $[u_1]\,\bullet$ & $653$ & $99\%$ & $63\%$ & $99\%$ & $99\%$ & $99\%$ \\
$u_1\,F\,u_2\rightarrow[u_1]$ & $[u_1]\,[u_2]$ & $1{,}825$ & $95\%$ & $93\%$ & $97\%$ & $81\%$ & $98\%$ \\
\botrule
\end{tabular*}
\end{table}

The checkpoints also keep familiar output lengths. The rule $u_1\,F\rightarrow[u_1]$ is demonstrated with one-token outputs, yet the four Transformer checkpoints answer none of its $653$ bare queries correctly and return the argument's colour twice on $95$--$99\%$ of them; the LSTM answers $36\%$ correctly. Conversely, every checkpoint answers every bare query of a primitive with a two- to six-token value with a single token, even when the two-token value is itself a support example.

Poor support reproduction does not explain this. For the held-out two-token value, our Transformer reproduces all $14$ support items in $328$ of the $1{,}000$ episodes, the LSTM in $450$ and the released checkpoints in $186$--$314$, and in every one of these episodes the bare query is still answered with a single token.

\subsection{Seven-reference patterns}
\label{sec:G_rhs7}

All $126$ seven-reference patterns that use both arguments occur in the $100{,}000$ meta-training grammars. Of these, $19$ appear only with string-slot forms and never with $u_1\,F\,u_2$:
\[
\begin{aligned}
&[u_1][u_1][u_1][u_1][u_2][u_1][u_1],\quad [u_1][u_1][u_1][u_2][u_1][u_1][u_1],\quad [u_1][u_1][u_1][u_2][u_1][u_2][u_1], \\
&[u_1][u_1][u_2][u_1][u_1][u_1][u_1],\quad [u_1][u_1][u_2][u_1][u_1][u_1][u_2],\quad [u_1][u_2][u_1][u_2][u_2][u_1][u_2], \\
&[u_1][u_2][u_2][u_2][u_1][u_1][u_2],\quad [u_1][u_2][u_2][u_2][u_2][u_2][u_1],\quad [u_2][u_1][u_1][u_1][u_1][u_1][u_2], \\
&[u_2][u_1][u_1][u_2][u_1][u_2][u_2],\quad [u_2][u_1][u_1][u_2][u_2][u_1][u_1],\quad [u_2][u_1][u_2][u_1][u_1][u_1][u_1], \\
&[u_2][u_1][u_2][u_2][u_2][u_2][u_1],\quad [u_2][u_1][u_2][u_2][u_2][u_2][u_2],\quad [u_2][u_2][u_1][u_1][u_2][u_1][u_2], \\
&[u_2][u_2][u_1][u_1][u_2][u_2][u_2],\quad [u_2][u_2][u_1][u_2][u_1][u_1][u_1],\quad [u_2][u_2][u_1][u_2][u_1][u_2][u_2], \\
&[u_2][u_2][u_2][u_1][u_2][u_2][u_2].
\end{aligned}
\]
The Transformer is as accurate on these new pairings as on $19$ attested controls ($95.3\%$ and $95.6\%$; $95{,}000$ queries per set). No individual pattern shows a clear gap: across the $19$ new pairings, Transformer accuracy ranges from $93.5\%$ to $97.0\%$, compared with $94.6$--$96.5\%$ across the controls, and for the other checkpoints the range across patterns is at most $5.3$ percentage points (Fig.~\ref{fig:supp_battery_overview}).

\subsection{Arrangements of two rules}
\label{sec:G_arrangements}

\begin{figure}[!htbp]
\centering
\includegraphics[width=\textwidth]{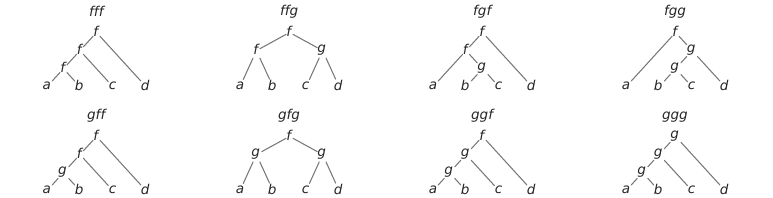}
\caption{\textbf{Interpreter bracketings for three function applications.} The eight written forms with $f$ earlier than $g$ in the rule list. Leaves $a,b,c,d$ retain input order. Rule $f$ reverses its two argument blocks; $g$ preserves them. Exchanging the rules' list positions can change the bracketing. These trees describe the task interpreter, not an inferred model computation.}
\label{fig:supp_composition_bracketings}
\end{figure}

We tested whether the checkpoints consistently compose two functions that are demonstrated in the same support. Function $f$ reverses the order of its two arguments and function $g$ preserves it. When an input contains both functions, the order of their rules in the rule list of $G$ determines how the interpreter brackets the input. For $a\,f\,b\,g\,c\,f\,d$ with primitive inputs $a,b,c,d$, the interpreter produces the chain $f(f(a,g(b,c)),d)$ with $f$ earlier, or the branch $g(f(a,b),f(c,d))$ with $g$ earlier (Fig.~\ref{fig:supp_composition_bracketings}). The rule list was hidden, so rule order had to be inferred from the support.

We used the same $500$ sampled grammars for both rule orders, and switched only the rule-list order of $f$ and $g$. Query inputs are the same in both versions and only the targets differ. We tested all fourteen forms with one to three applications of $f$ and $g$, such as $fg$ or $gfg$, with ten novel queries per form and grammar. The support is constructed so that the target of each query is unambiguously determined, as all grammars of the meta-grammar family that match the support input--output mapping lead to the same tested query targets (Methods, ``Arrangements of two rules''). Every support has $14$ examples and $72$ tokens and always shows the primitives, each function alone, and $f$ and $g$ with an argument of more than one token on either side. In addition, it shows the mixed form $fg$, $gf$, or both. All five checkpoints reproduce every support example.

\begin{figure}[!tbp]
\centering
\includegraphics[width=\textwidth]{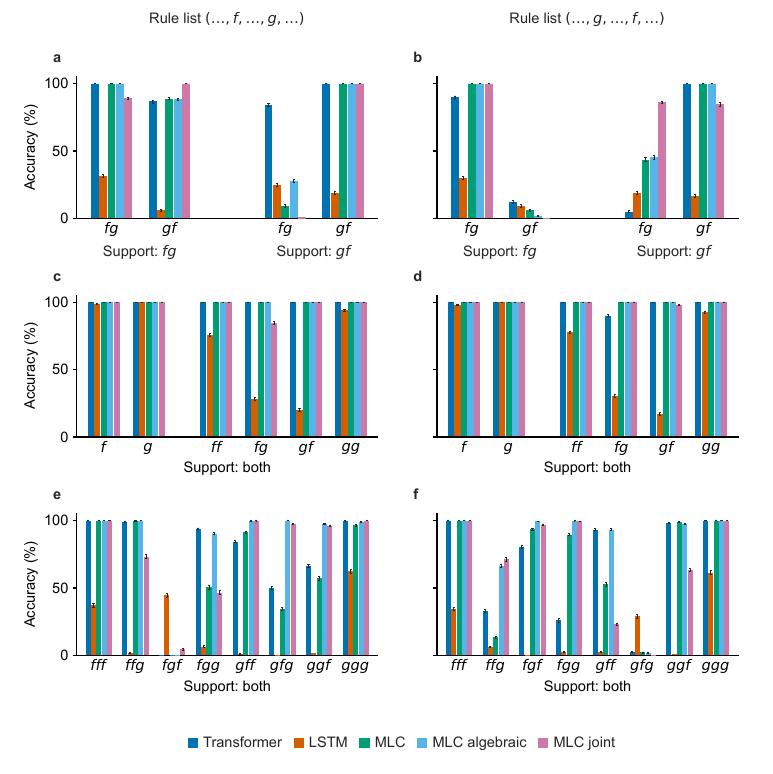}
\caption{\textbf{Paired composition across five checkpoints.} Columns show the two rule-list orders for the same $500$ base grammars. \textbf{a,b,} Mixed two-application queries with $fg$ or $gf$ demonstrated, as labelled. \textbf{c,d,} One- and two-application queries with both mixed forms demonstrated. \textbf{e,f,} Three-application queries with the same support. Bars show accuracy over ten novel queries per form and grammar, with inputs matched across rule lists and both chain and branching derivations included; error bars are $95\%$ bootstrap intervals over base grammars.}
\label{fig:supp_battery_all_checkpoints}
\end{figure}

The four Transformer checkpoints answer single applications and repeated applications of one function almost perfectly (Fig.~\ref{fig:supp_battery_all_checkpoints}c--f). They also answer novel $fg$ and $gf$ queries with $84.7$--$100\%$ accuracy when the queried form is demonstrated, while LSTM accuracy remains at $17.0$--$31.8\%$ (Fig.~\ref{fig:supp_battery_all_checkpoints}a--d). Transfer to the mixed form absent from support depends on rule-list order. With only $fg$ demonstrated, the Transformer checkpoints reach $86.7$--$100\%$ on $gf$ with $f$ earlier but at most $12.4\%$ with $g$ earlier. With only $gf$ demonstrated, our Transformer reaches $84.1\%$ on $fg$ with $f$ earlier and $5.1\%$ with $g$ earlier.

With both mixed forms demonstrated, transfer to three applications remains selective (Fig.~\ref{fig:supp_battery_all_checkpoints}e,f). All four Transformer checkpoints score at most $4.6\%$ on $fgf$ with $f$ earlier and at most $2.6\%$ on $gfg$ with $g$ earlier, yet answer $fff$ perfectly under both orders. Input length, application count and chain depth alone therefore do not explain these failures. On the other mixed forms, our Transformer's accuracy ranges from $26.1\%$ to $98.8\%$, depending on the rule-list order. The LSTM remains below $45\%$ on every mixed form.

\begin{figure}[!tbp]
\centering
\includegraphics[width=\textwidth]{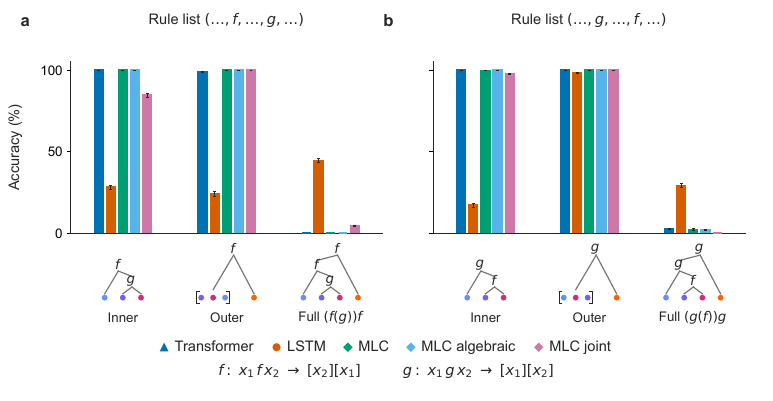}
\caption{\textbf{Inner, outer and full-query accuracy across five checkpoints.} \textbf{a,b,} $fgf$ with $f$ earlier and $gfg$ with $g$ earlier in the rule list, respectively. The outer query replaces the inner expression with its correct output as primitive words (bracketed tokens in the trees; colours identify primitives). The three queries are decoded independently from the same support, which demonstrates both $fg$ and $gf$. Bars show accuracy on all $5{,}000$ matched triples from $500$ paired base grammars; error bars are $95\%$ bootstrap intervals over base grammars.}
\label{fig:supp_composition_matched_queries}
\end{figure}

For $fgf$ with $f$ earlier and $gfg$ with $g$ earlier, we posed the inner expression and the outer application as separate queries under the same support (Fig.~\ref{fig:supp_composition_matched_queries}). For $a\,f\,b\,g\,c\,f\,d$ with $f$ earlier, these are $a\,f\,b\,g\,c$ and $b\,c\,a\,f\,d$; the latter supplies the correct inner output as primitive words and has the same target as the full query. Our Transformer answers the inner query in $100\%$ and the outer query in $98.8\%$ of the $5{,}000$ triples, but none of the full queries. For $gfg$ with $g$ earlier, the corresponding values are $100\%$, $100\%$ and $2.6\%$, and the released Transformer checkpoints show the same pattern. The LSTM answers both shorter queries correctly in only $6.9\%$ and $17.1\%$ of triples, so the comparison is not informative for it.

This pattern also occurs without the human-like heuristic target replacements used in \emph{MLC(algebraic+biases)}. With both mixed forms demonstrated, \emph{MLC(algebraic only)} answers $fg$ and $gf$ perfectly but scores $0\%$ on $fgf$ with $f$ earlier and $2.0\%$ on $gfg$ with $g$ earlier.

A system that had inferred the demonstrated rules would respond consistently under the same support~\citep{FodorPylyshyn1988connectionism}. Our Transformer answers both parts correctly in $98.8\%$ and $100\%$ of triples, but the full query in $0\%$ and $2.6\%$. Nothing in the MLC design enforces agreement between responses, as each query is decoded separately without access to previous responses.

\begin{figure}[p]
\centering
\includegraphics[width=\textwidth]{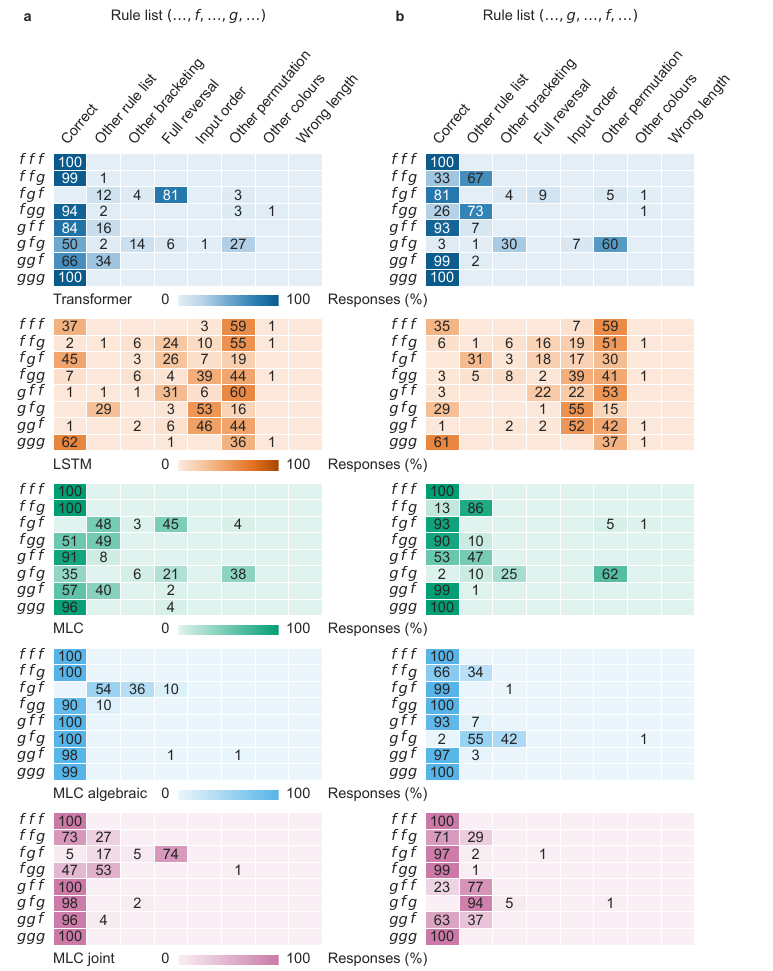}
\caption{\textbf{Composition responses by output category.} \textbf{a,b,} The two rule-list orders, with both mixed forms demonstrated. Rows contain $5{,}000$ responses each and sum to $100\%$; values below $0.5\%$ are unlabelled. Each response is assigned to the first matching category from left to right. ``Other rule list'' denotes the target under the exchanged rule order, ``other bracketing'' the output of another bracketing of the same written form, and ``other permutation'' any other ordering of the target colours.}
\label{fig:supp_composition_responses}
\end{figure}

The composition errors predominantly concern output order: across the three-application forms, $99.2$--$100\%$ of each checkpoint's errors preserve the target colours and their counts (Fig.~\ref{fig:supp_composition_responses}). For $fgf$ with $f$ earlier, full reversal accounts for $80.5\%$ of our Transformer's responses, whereas \emph{MLC(algebraic only)} mostly returns the target under the opposite rule list ($54.0\%$) or the output of another bracketing ($36.0\%$). These categories classify query responses independently and thus do not refer to a consistent alternative rule order or bracketing across queries of the same episode.

\FloatBarrier
\clearpage

\section{Construction of evaluation episodes}
\label{secD}

The algorithms describe episode sampling, rule injection, and query
filtering and construction.

\algrenewcommand\algorithmicrequire{\textbf{Input:}}
\algrenewcommand\algorithmicensure{\textbf{Output:}}

\subsection{Sampling grammars and examples}
\label{sec:F_constrained}

\begin{algorithm}[H]
\caption{Constructing an evaluation episode}
\label{alg:generate-episode}
\small
\begin{algorithmic}[1]
\Require meta-grammar parameters $\theta$, support/query constraints
$C_s,C_q$, example counts $n_s,n_q$, minimum query count $n_q^{\min}$,
excluded grammars $\mathcal R_{\mathrm{train}}$, optional rule
restrictions $\mathcal T,n_u,n_b$
\Ensure episode $(G,S,Q)$ or failure $\bot$
\State $G\gets\Call{SampleGrammar}{\theta}$
  \Comment{restrict LHS forms to $\mathcal T$ if specified}
\State \textbf{if} $G=\bot$ \textbf{then return} $\bot$
\State \textbf{if} $G$ violates a specified unary or binary rule count
  \textbf{then return} $\bot$
\State \textbf{if} function rules match the human few-shot patterns
  \textbf{then return} $\bot$
\State \textbf{if} $R(G)\in\mathcal R_{\mathrm{train}}$
  \textbf{then return} $\bot$
\State $S\gets\Call{CollectExamples}{G,C_s,n_s,n_s,\emptyset}$
\State \textbf{if} $S=\bot$ \textbf{then return} $\bot$
\State $D\gets\Call{CollectExamples}{
  G,C_q,n_q,n_q^{\min},\operatorname{inputs}(S)}$
\State \textbf{if} $D=\bot$ \textbf{then return} $\bot$
\State $Q\gets\operatorname{inputs}(D)$
\State \Return $(G,\Call{Shuffle}{S},\Call{Shuffle}{Q})$
\Statex
\Procedure{CollectExamples}{$G,C,n,n^{\min},E$}
  \Comment{target $n$, minimum $n^{\min}$}
  \State $D\gets\emptyset$
    \Comment{accepted input--output pairs}
  \State $\mathrm{attempts}\gets 0$
  \While{$|D|<n$ \textbf{and} $\mathrm{attempts}<1000\times n$}
    \State $\mathrm{attempts}\gets\mathrm{attempts}+1$
    \State $x\gets\Call{DrawInput}{G,C}$
    \State \textbf{if} $x=\bot$ \textbf{or}
      $x\in E\cup\operatorname{inputs}(D)$
      \textbf{then continue}
      \Comment{$E$: excluded inputs}
    \State $y\gets G(x)$ \Comment{interpret $x$ under $G$}
    \State \textbf{if} $y=\bot$ \textbf{or} $y$ is empty
      \textbf{then continue}
    \State \textbf{if} $y$ contains symbols outside $\mathcal O$
      \textbf{then continue}
    \State \textbf{if} $(x,y)$ violates $C$
      \textbf{then continue}
    \State $D\gets D\cup\{(x,y)\}$
  \EndWhile
  \State \textbf{if} $|D|<n^{\min}$ \textbf{then return} $\bot$
  \State \Return $D$
\EndProcedure
\end{algorithmic}
\end{algorithm}

An episode $(G,S,Q)$ consists of an interpretation grammar $G$, a support
set $S$ of input--output pairs and a set of query inputs $Q$.
The grammar $G$ is sampled from the meta-grammar with
\textsc{SampleGrammar}$(\theta)$ using the supplied meta-parameters
$\theta=(\mathcal I,\mathcal O,n_p,n_f,\min_f,\max_f)$.
Admissible left-hand-side (LHS) forms $\mathcal T$ and unary and binary
rule counts $n_u$ and $n_b$ may also be specified. The grammar is rejected
if it violates the specified parameters or these restrictions, if its
function rules match the human few-shot rules after operator renaming,
or if its sorted rule list $R(G)$ occurs in the meta-training collection
$\mathcal R_{\mathrm{train}}$.

Using $G$, $n_s$ support pairs and up to $n_q$ query inputs are sampled
to form $S$ and $Q$, with $\operatorname{inputs}(S)\cap Q=\emptyset$.
The query set is accepted only if it contains at least $n_q^{\min}$
inputs, where $n_q^{\min}=n_q$ by default.
The sequence-level constraints in Extended Data Table~1 are specified
separately as $C_s$ for support examples and $C_q$ for query examples.
Additional restrictions may exclude particular input--output length
combinations or require query inputs to contain a specified template.

For each set, \textsc{CollectExamples} samples candidate inputs $x$ with
\textsc{DrawInput}, allowing up to $1000n$ attempts for $n$ requested
examples. \textsc{DrawInput} selects an admissible LHS pattern from $G$
and fills its argument variables. Each atom variable $u_j$ is replaced
by a primitive input token, while each string variable $x_j$ is replaced
by either a primitive token or an input generated by a recursive call to
\textsc{DrawInput}. The bounds on input length and nesting degree guide
these choices. The resulting input $x$ is interpreted under $G$ to obtain
its output $y=G(x)$. The pair $(x,y)$ is retained only if it satisfies
the corresponding constraints and $y$ is a non-empty sequence consisting
only of output symbols in $\mathcal O$.
Duplicate inputs are discarded. If either set cannot be obtained,
sampling continues with a new grammar, using a limit of $500$ grammar
attempts ($2{,}000$ for $(|x|,|y|)=(2,7)$ and $(2,8)$).

\subsection{Rule injection}
\label{sec:F_interventions}

Rule injection replaces primitive outputs or function rules using the
ordered lists $w$ and $\mathcal F$, respectively
(Algorithm~\ref{alg:inject-rule}). Support and query examples are sampled
from the modified grammar as in Section~\ref{sec:F_constrained}.

\paragraph{Primitive mappings.}
Primitive replacements are specified by an ordered list $w$ of output
patterns written as sequences of variables $c_i$. Each distinct variable
$c_i$ is replaced throughout $w$ by one output symbol sampled uniformly
without replacement from $\mathcal O$. The resulting output sequences
are assigned in order to distinct primitives selected at random. If a
modified primitive produces the same output as a primitive left
unchanged, the output symbols are sampled again. The condition stops
if no valid symbol assignment is found within $100$ attempts for a grammar.

\paragraph{Function rules.}
Function replacements are specified by an ordered list $\mathcal F$ whose
entries pair a new LHS pattern $\tau$ with an RHS specification $\rho$.
By default, the original rule has the same number of arguments as the new
LHS pattern $\tau$. If $h$ is specified, every original rule selected for
replacement must have $h$ arguments.\footnote{The meta-grammar samples
only unary and binary function rules. Replacements with more than two
arguments therefore require an explicit $h$ ($h=2$ in the reported
conditions).}
The RHS specification $\rho$ contains either a fixed RHS expression or
sampling bounds, optionally with an initial list of references to argument
outputs. By default, this list $R_0$ contains one reference per argument
of the new LHS $\tau$. \textsc{SampleRHS} extends and shuffles $R_0$ within
the length bounds (see Methods, ``Meta-grammar'').

The set $U$ records the indices of already replaced function rules.
Entries in $\mathcal F$ are applied in order, selecting each applicable
rule $G_i$ uniformly with $i\notin U$. Each replacement retains the
original operator and position $i$ in $G$. If an entry has no matching
rule, a new grammar is sampled. The condition stops after $100$
unsuccessful grammar attempts.

\begin{algorithm}[H]
\caption{Injecting primitive mappings and function rules}
\label{alg:inject-rule}
\small
\begin{algorithmic}[1]
\Require meta-grammar parameters $\theta$, primitive output patterns $w$,
function replacements $\mathcal F$, optional arity $h$ of rules to replace
\Ensure grammar $G$ or failure $\bot$
\For{at most $100$ grammar attempts}
  \State $G\gets\Call{SampleGrammar}{\theta,w}$
    \Comment{apply primitive output patterns $w$}
  \State \textbf{if} $G=\bot$ \textbf{then return} $\bot$
  \State $U\gets\emptyset$
    \Comment{indices of replaced function rules}
  \ForAll{$(\tau,\rho)\in\mathcal F$}
    \Comment{$\tau$: LHS pattern, $\rho$: RHS specification}
    \State $h_\tau\gets h$ if specified, otherwise the number of
      arguments in $\tau$
    \State $H\gets\{i\notin U:G_i\text{ is a function rule with }
      h_\tau\text{ arguments}\}$
    \State \textbf{if} $H=\emptyset$ \textbf{then break}
    \State $i\sim\operatorname{Uniform}(H)$
    \State $\tau'\gets\tau$ with $G_i$'s operator
    \If{$\rho$ is a fixed expression}
      \State $\rho'\gets\rho$
    \Else
      \State $(\ell_{\min},\ell_{\max})\gets$ bounds in $\rho$,
        defaulting to $(\min_f,\max_f)$
      \State $R_0\gets$ initial reference list in $\rho$, defaulting
        to one output reference per argument of $\tau$
      \State $\rho'\gets\Call{SampleRHS}{
        R_0,\ell_{\min},\ell_{\max}}$
    \EndIf
    \State $G_i\gets(\tau'\to\rho')$
    \State $U\gets U\cup\{i\}$
  \EndFor
  \State \textbf{if} $|U|=|\mathcal F|$
    \textbf{then return} $G$
\EndFor
\State \Return $\bot$
\end{algorithmic}
\end{algorithm}

\subsection{Filtered and constructed queries}
\label{sec:F_queries}

\begin{algorithm}[H]
\caption{Constructing queries from a template}
\label{alg:pose-queries}
\small
\begin{algorithmic}[1]
\Require grammar $G$, support set $S$, query template $t$, optional
tested rules $J$, query count $n_q$, allowed primitives $\mathcal A$
(default: all primitives of $G$), string lengths $L$
\Ensure query set $Q'$ or failure $\bot$
\State \textbf{if} an operator in $t$ is absent from support inputs
  \textbf{then return} $\bot$
\State $Q'\gets\emptyset$
\State $\mathrm{attempts}\gets 0$
\While{$|Q'|<n_q$ \textbf{and} $\mathrm{attempts}<200\times n_q$}
  \State $\mathrm{attempts}\gets\mathrm{attempts}+1$
  \State $x\gets\Call{FillTemplate}{t,G,L}$
  \State \textbf{if} $x\in\operatorname{inputs}(S)\cup Q'$
    \textbf{then continue}
  \State \textbf{if} $x$ contains a primitive outside $\mathcal A$
    \textbf{then continue}
  \State $y\gets G(x)$ \Comment{interpret $x$ under $G$}
  \State \textbf{if} $y=\bot$ \textbf{or} $y$ is empty
    \textbf{then continue}
  \State \textbf{if} $y$ contains symbols outside $\mathcal O$
    \textbf{then continue}
  \State \textbf{if} $J$ is specified \textbf{and} interpreting $x$
    applies no rule in $J$ \textbf{then continue}
  \State $Q'\gets Q'\cup\{x\}$
\EndWhile
\State \textbf{if} $|Q'|<n_q$ \textbf{then return} $\bot$
\State \Return $Q'$
\end{algorithmic}
\end{algorithm}

A query template $t$ defines an input pattern containing fixed symbols
and argument variables. Atom variables $u_j$ bind single input tokens,
while string variables $x_j$ bind non-empty input sequences. Query
filtering uses \textsc{CollectExamples}
(Section~\ref{sec:F_constrained}) to select pairs satisfying $C_q$
whose inputs contain the specified pattern.

Query construction uses \textsc{FillTemplate}$(t,G,L)$ to fill each
$u_j$ with one token and each $x_j$ with a sequence of the length
specified in $L$ (Algorithm~\ref{alg:pose-queries}).
For each distinct variable, tokens are sampled independently and
uniformly from the primitive inputs of $G$. The same filling is used
at every occurrence of that variable.
Each constructed input $x$ is paired with its output $y=G(x)$.
Accepted inputs contain only primitives from $\mathcal A$, which defaults
to all primitive inputs of $G$. If a set $J$ of tested rules is specified,
interpreting $x$ must apply at least one of them, at any depth.
Once $n_q$ distinct inputs are accepted, they replace $Q$.
Analyses of constructed queries include only episodes with successful
construction.
\FloatBarrier

\clearpage
\section{Systematic variation of shared architecture parameters for meta-learning neural networks} \label{secE}

\begin{figure}[!htbp]
    \centering
    \includegraphics[width=\linewidth]{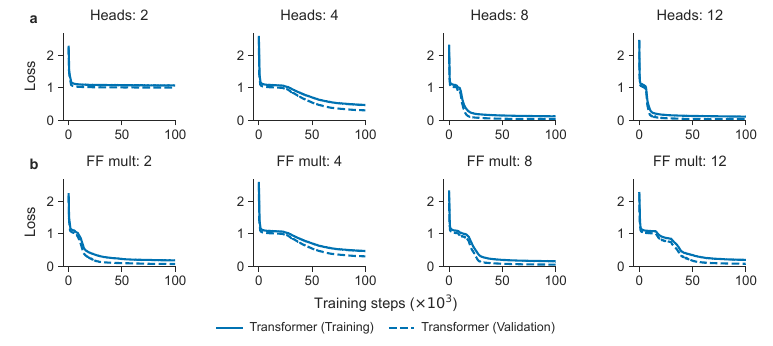}
    \caption{\textbf{Training curves for the Transformer-specific ablations.}
    Token cross-entropy during training (solid) and validation
    (dashed) over the first $100{,}000$ steps, with one run per setting.
    \textbf{a,} Attention-head count at $16$ hidden units per head
    and feed-forward multiplier $4$. The corresponding hidden sizes
    are $32$, $64$, $128$ and $192$, respectively; head count and
    model width therefore increase together.
    \textbf{b,} Feed-forward multiplier at four heads and hidden
    size $64$.}
    \label{fig:supp_transformer_architecture_ablations}
\end{figure}

We varied width, depth and dropout in the Transformer with context
and the LSTM without context, which retains its
key--value memory. Increasing width reduced validation loss in both models.
A second Transformer layer substantially reduced loss, while
further depth produced smaller and non-monotonic changes.
LSTM loss varied little with depth or dropout. For the Transformer,
dropout $0.3$ slowed convergence and $0.5$ produced unstable
validation loss. Validation loss can lie below training loss
because training targets include heuristic and perturbed responses,
whereas validation targets are algebraic
(Supplementary Section~\ref{secA}). In the Transformer-specific sweeps, validation loss generally
decreased as attention-head count and hidden size increased
together, while the feed-forward multiplier had a non-monotonic
effect at fixed width and head count
(Fig.~\ref{fig:supp_transformer_architecture_ablations}).

\begin{figure}[!htbp]
    \centering
    \includegraphics[width=1\textwidth]{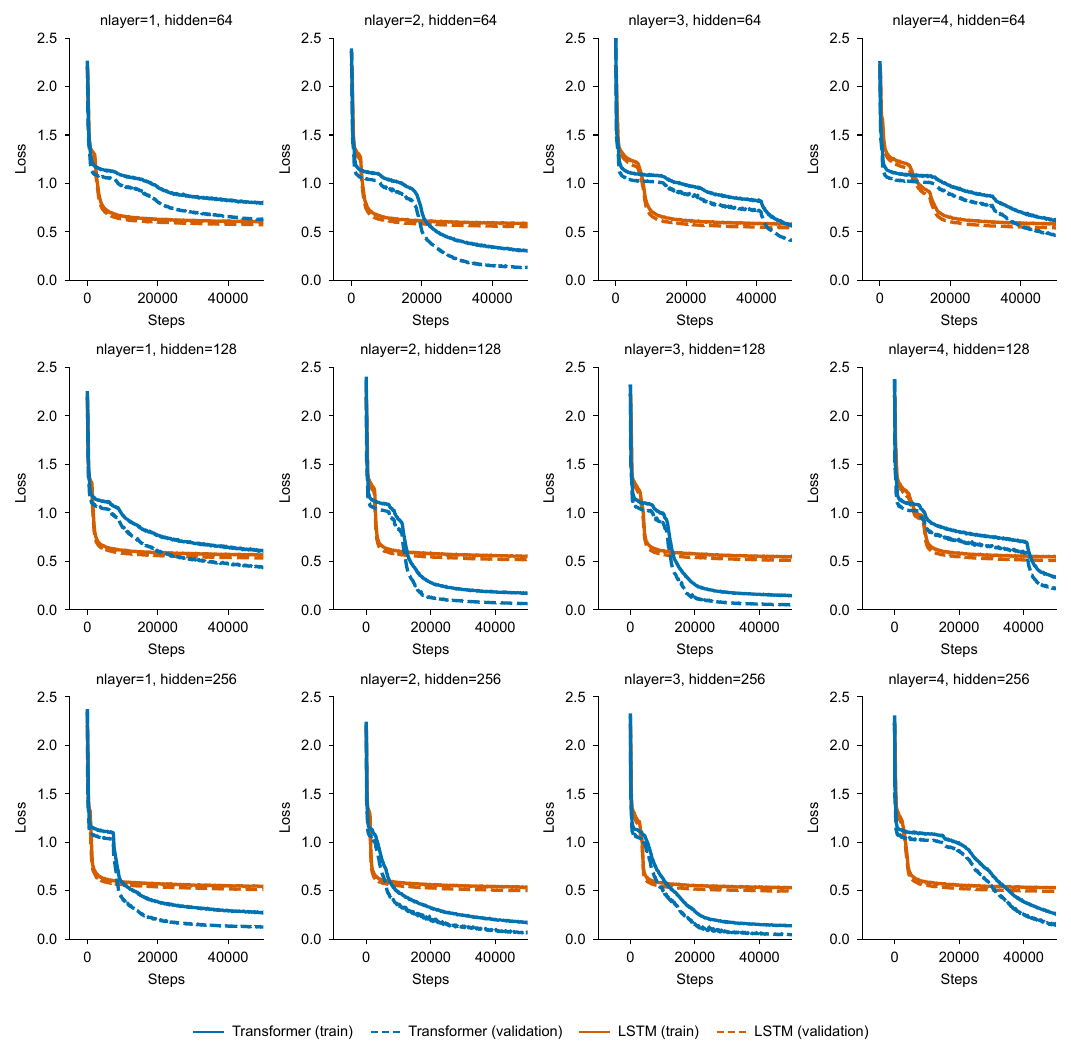}
    \caption{\textbf{Depth$\times$width ablation at fixed dropout ($p=0.1$).}
Training (solid) and validation (dashed) loss over optimization steps for the Transformer (blue) and LSTM (orange) across a grid of encoder--decoder depths ($n_{\mathrm{layer}}=1$--$4$, columns) and hidden sizes ($64$,$128$, and $256$, rows), with all other settings held fixed (see Methods).
Across most settings, the Transformer continues to reduce validation loss as depth and/or hidden size increase, whereas the LSTM curves are comparatively similar across the grid and tend to plateau earlier, indicating weaker sensitivity of optimization dynamics to these architectural changes under this dropout setting and fixed meta-training design choices.}
    \label{fig:complete_architecture_ablation_dropout01}
\end{figure}

\begin{figure}[!htbp]
    \centering
    \includegraphics[width=1\textwidth]{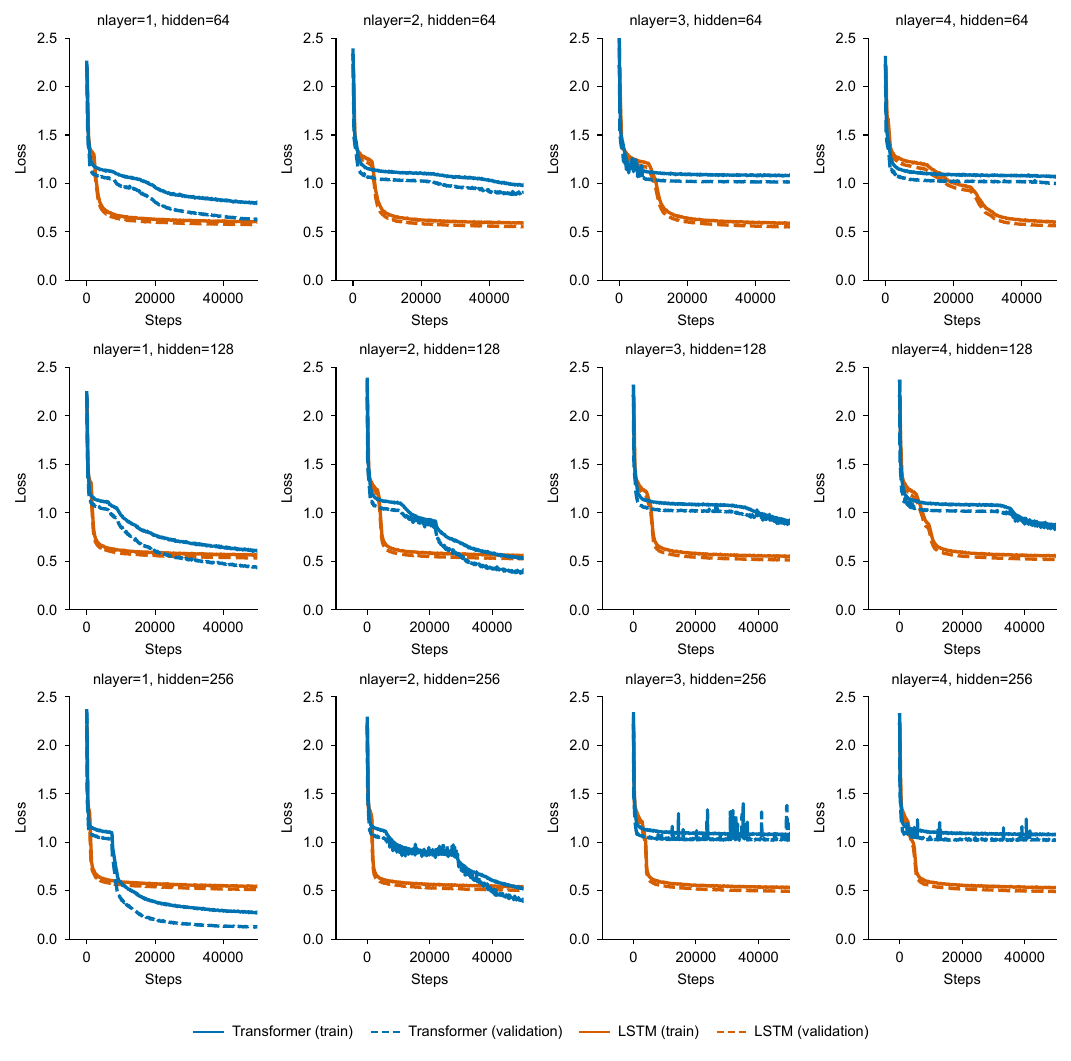}
    \caption{\textbf{Depth$\times$width ablation at fixed dropout ($p=0.3$).}
Training (solid) and validation (dashed) loss over optimization steps for the Transformer (blue) and LSTM (orange) across a grid of encoder--decoder depths ($n_{\mathrm{layer}}=1$--$4$, columns) and hidden sizes ($64$, $128$ and $256$, rows), with all other settings held fixed (see Methods).
Across most settings, increasing depth and/or hidden size yields larger reductions in validation loss for the Transformer than for the LSTM. The LSTM loss curves remain comparatively similar across the grid and tend to plateau early, whereas several higher-capacity Transformer settings exhibit reduced but occasionally less smooth validation-loss trajectories under this dropout setting.}
    \label{fig:complete_architecture_ablation_dropout03}
\end{figure}

\begin{figure}[!htbp]
    \centering
    \includegraphics[width=1\textwidth]{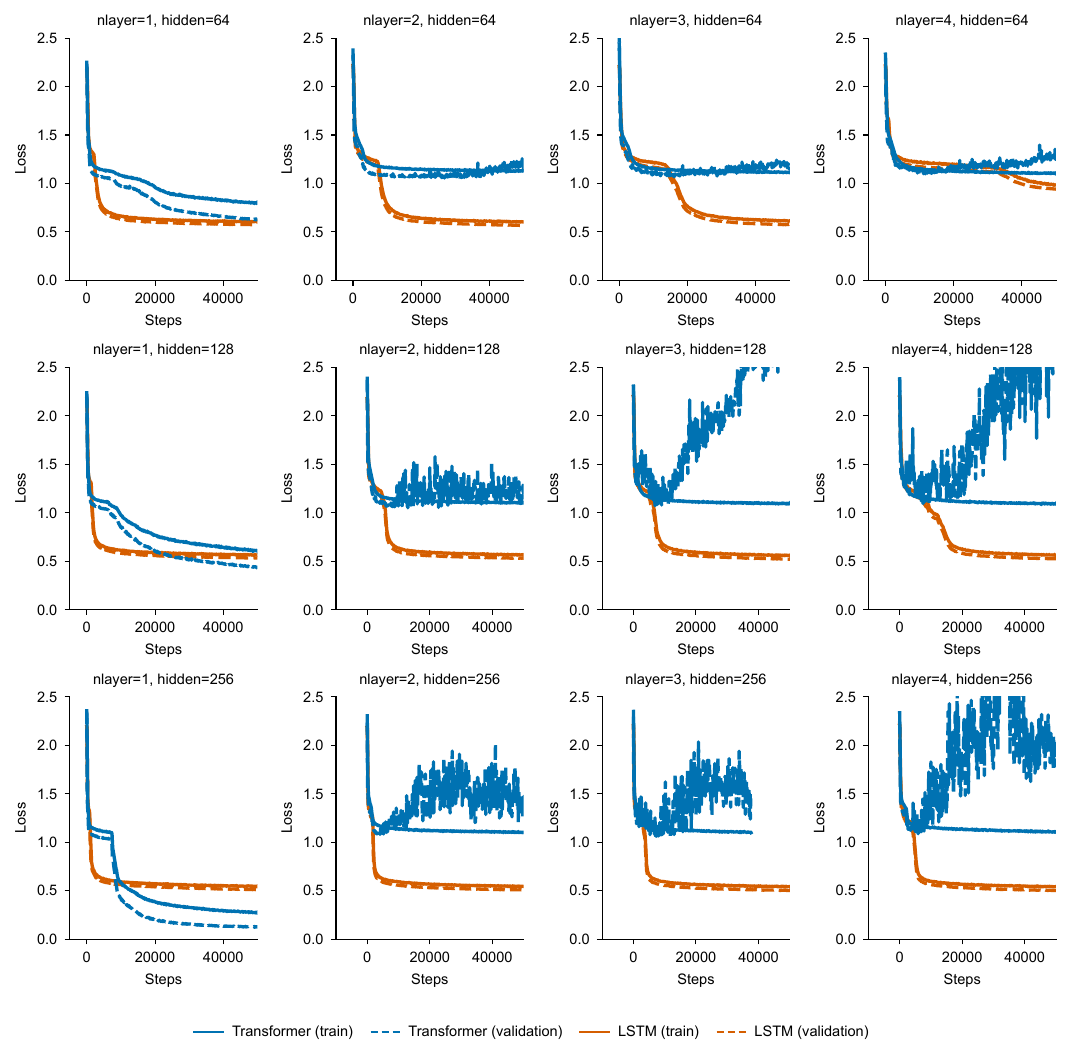}
    \caption{\textbf{Depth$\times$width ablation at fixed dropout ($p=0.5$).}
Training (solid) and validation (dashed) loss over optimization steps for the Transformer (blue) and LSTM (orange) across a grid of encoder--decoder depths ($n_{\mathrm{layer}}=1$--$4$, columns) and hidden sizes ($64$, $128$ and $256$, rows), with all other settings held fixed (see Methods).
At this higher dropout rate, many Transformer configurations exhibit unstable optimization, with validation loss increasing and/or becoming highly variable after the initial decrease, particularly at larger hidden sizes and greater depth. In contrast, the LSTM loss curves remain comparatively stable across the grid and converge to similar values, indicating lower sensitivity of its training dynamics to these architectural changes under strong dropout.}
    \label{fig:complete_architecture_ablation_dropout05}
\end{figure}

Figs.~\ref{fig:complete_architecture_ablation_dropout01}--\ref{fig:complete_architecture_ablation_dropout05} give training and validation loss over the complete grid of depths and hidden sizes at the three dropout rates. To relate width to task performance, we scored the Transformer at the reported depth and dropout (three encoder and three decoder layers, $p = 0.1$) across hidden sizes $8$ to $256$ on the $200$ released validation episodes (Fig.~\ref{fig:supp_hidden_size_accuracy}).

%% file: sn-bibliography.bib
@inproceedings{lake2019compositional,
  title={Compositional generalization through meta sequence-to-sequence learning},
  author={Lake, Brenden M.},
  booktitle={Advances in Neural Information Processing Systems},
  volume={32},
  year={2019},
  publisher={Curran Associates Inc.}
}

@article{lake2023human,
  title={Human-like systematic generalization through a meta-learning neural network},
  author={Lake, Brenden M. and Baroni, Marco},
  journal={Nature},
  volume={623},
  number={7985},
  pages={115--121},
  year={2023},
  doi={10.1038/s41586-023-06668-3}
}

@inproceedings{lake2018generalization,
  title={Generalization without systematicity: on the compositional skills of sequence-to-sequence recurrent networks},
  author={Lake, Brenden M. and Baroni, Marco},
  booktitle={Proceedings of the 35th International Conference on Machine Learning},
  series={Proceedings of Machine Learning Research},
  volume={80},
  pages={2873--2882},
  year={2018},
  publisher={PMLR}
}

@article{FodorPylyshyn1988connectionism,
  title={Connectionism and cognitive architecture: a critical analysis},
  author={Fodor, Jerry A. and Pylyshyn, Zenon W.},
  journal={Cognition},
  volume={28},
  number={1--2},
  pages={3--71},
  year={1988},
  doi={10.1016/0010-0277(88)90031-5}
}

@article{Fodor2001,
  title={Language, thought and compositionality},
  author={Fodor, Jerry A.},
  journal={Mind \& Language},
  volume={16},
  number={1},
  pages={1--15},
  year={2001},
  doi={10.1111/1468-0017.00153}
}

@misc{nezhurina2024alicewonderlandsimpletasks,
  title={Alice in {W}onderland: simple tasks showing complete reasoning breakdown in state-of-the-art large language models},
  author={Nezhurina, Marianna and Cipolina-Kun, Lucia and Cherti, Mehdi and Jitsev, Jenia},
  year={2024},
  note={Preprint at \url{https://arxiv.org/abs/2406.02061}}
}

@inproceedings{dziri2024faith,
  title={Faith and fate: limits of transformers on compositionality},
  author={Dziri, Nouha and Lu, Ximing and Sclar, Melanie and Li, Xiang Lorraine and Jiang, Liwei and Lin, Bill Yuchen and Welleck, Sean and West, Peter and Bhagavatula, Chandra and {Le Bras}, Ronan and Hwang, Jena D. and Sanyal, Soumya and Ren, Xiang and Ettinger, Allyson and Harchaoui, Zaid and Choi, Yejin},
  booktitle={Advances in Neural Information Processing Systems},
  volume={36},
  pages={70293--70332},
  year={2023},
  publisher={Curran Associates Inc.},
  doi={10.52202/075280-3081}
}

@misc{dinu2024symbolicaiframeworklogicbasedapproaches,
  title={{SymbolicAI}: a framework for logic-based approaches combining generative models and solvers},
  author={Dinu, Marius-Constantin and Leoveanu-Condrei, Claudiu and Holzleitner, Markus and Zellinger, Werner and Hochreiter, Sepp},
  year={2024},
  note={Preprint at \url{https://arxiv.org/abs/2402.00854}}
}

@article{ellis2020dreamcodergrowinggeneralizableinterpretable,
  title={{DreamCoder}: growing generalizable, interpretable knowledge with wake--sleep {Bayesian} program learning},
  author={Ellis, Kevin and Wong, Lionel and Nye, Maxwell and Sabl{\'e}-Meyer, Mathias and Cary, Luc and Anaya Pozo, Lore and Hewitt, Luke and Solar-Lezama, Armando and Tenenbaum, Joshua B.},
  journal={Philosophical Transactions of the Royal Society A},
  volume={381},
  number={2251},
  pages={20220050},
  year={2023},
  doi={10.1098/rsta.2022.0050}
}

@inproceedings{shindo2025iclr,
  title={{BlendRL}: a framework for merging symbolic and neural policy learning},
  author={Shindo, Hikaru and Delfosse, Quentin and Dhami, Devendra Singh and Kersting, Kristian},
  booktitle={International Conference on Learning Representations},
  year={2025},
  url={https://openreview.net/forum?id=60i0ksMAhd}
}

@inproceedings{bayat2025the,
  title={The pitfalls of memorization: when memorization hurts generalization},
  author={Bayat, Reza and Pezeshki, Mohammad and Dohmatob, Elvis and Lopez-Paz, David and Vincent, Pascal},
  booktitle={International Conference on Learning Representations},
  year={2025},
  url={https://openreview.net/forum?id=vVhZh9ZpIM}
}

@inproceedings{meszaros2024rule,
  title={Rule extrapolation in language modeling: a study of compositional generalization on {OOD} prompts},
  author={M{\'e}sz{\'a}ros, Anna and Ujv{\'a}ry, Szilvia and Brendel, Wieland and Reizinger, Patrik and Husz{\'a}r, Ferenc},
  booktitle={Advances in Neural Information Processing Systems},
  volume={37},
  pages={34870--34899},
  year={2024},
  publisher={Curran Associates Inc.},
  doi={10.52202/079017-1101}
}

@inproceedings{deletang2023neural,
  title={Neural networks and the {Chomsky} hierarchy},
  author={Del{\'e}tang, Gr{\'e}goire and Ruoss, Anian and Grau-Moya, Jordi and Genewein, Tim and Wenliang, Li Kevin and Catt, Elliot and Cundy, Chris and Hutter, Marcus and Legg, Shane and Veness, Joel and Ortega, Pedro A.},
  booktitle={International Conference on Learning Representations},
  year={2023},
  url={https://openreview.net/forum?id=WbxHAzkeQcn}
}

@inproceedings{mirzadeh2024gsmsymbolic,
  title={{GSM-Symbolic}: understanding the limitations of mathematical reasoning in large language models},
  author={Mirzadeh, Iman and Alizadeh, Keivan and Shahrokhi, Hooman and Tuzel, Oncel and Bengio, Samy and Farajtabar, Mehrdad},
  booktitle={International Conference on Learning Representations},
  year={2025},
  url={https://openreview.net/forum?id=AjXkRZIvjB}
}

@inproceedings{nguyen2024ngram,
  title={Understanding transformers via {N-gram} statistics},
  author={Nguyen, Timothy},
  booktitle={Advances in Neural Information Processing Systems},
  volume={37},
  pages={98049--98082},
  year={2024},
  publisher={Curran Associates Inc.},
  doi={10.52202/079017-3112}
}

@inproceedings{nikankin2024arithmetic,
  title={Arithmetic without algorithms: language models solve math with a bag of heuristics},
  author={Nikankin, Yaniv and Reusch, Anja and Mueller, Aaron and Belinkov, Yonatan},
  booktitle={International Conference on Learning Representations},
  year={2025},
  url={https://openreview.net/forum?id=O9YTt26r2P}
}

@inproceedings{keysers2020measuring,
  title={Measuring compositional generalization: a comprehensive method on realistic data},
  author={Keysers, Daniel and Sch{\"a}rli, Nathanael and Scales, Nathan and Buisman, Hylke and Furrer, Daniel and Kashubin, Sergii and Momchev, Nikola and Sinopalnikov, Danila and Stafiniak, Lukasz and Tihon, Tibor and Tsarkov, Dmitry and Wang, Xiao and {van Zee}, Marc and Bousquet, Olivier},
  booktitle={International Conference on Learning Representations},
  year={2020},
  url={https://openreview.net/forum?id=SygcCnNKwr}
}

@inproceedings{kim2020cogs,
  title={{COGS}: a compositional generalization challenge based on semantic interpretation},
  author={Kim, Najoung and Linzen, Tal},
  booktitle={Proceedings of the 2020 Conference on Empirical Methods in Natural Language Processing},
  pages={9087--9105},
  year={2020},
  publisher={Association for Computational Linguistics},
  doi={10.18653/v1/2020.emnlp-main.731}
}

@inproceedings{gontier2020measuring,
  title={Measuring systematic generalization in neural proof generation with transformers},
  author={Gontier, Nicolas and Sinha, Koustuv and Reddy, Siva and Pal, Christopher},
  booktitle={Advances in Neural Information Processing Systems},
  volume={33},
  pages={22231--22242},
  year={2020},
  publisher={Curran Associates Inc.}
}

@article{hupkes2020compositionality,
  title={Compositionality decomposed: how do neural networks generalise?},
  author={Hupkes, Dieuwke and Dankers, Verna and Mul, Mathijs and Bruni, Elia},
  journal={Journal of Artificial Intelligence Research},
  volume={67},
  pages={757--795},
  year={2020},
  doi={10.1613/jair.1.11674}
}

@article{geirhos2020shortcut,
  title={Shortcut learning in deep neural networks},
  author={Geirhos, Robert and Jacobsen, J{\"o}rn-Henrik and Michaelis, Claudio and Zemel, Richard and Brendel, Wieland and Bethge, Matthias and Wichmann, Felix A.},
  journal={Nature Machine Intelligence},
  volume={2},
  number={11},
  pages={665--673},
  year={2020},
  doi={10.1038/s42256-020-00257-z}
}

@article{firestone2020performance,
  title={Performance vs. competence in human--machine comparisons},
  author={Firestone, Chaz},
  journal={Proceedings of the National Academy of Sciences},
  volume={117},
  number={43},
  pages={26562--26571},
  year={2020},
  doi={10.1073/pnas.1905334117}
}

@article{riveland_compositionality_2026,
  title={The compositionality continuum as a principle for studying the neural basis of intelligence},
  author={Riveland, Reidar and Pouget, Alexandre and Driscoll, Laura},
  journal={Nature Neuroscience},
  volume={29},
  number={9},
  pages={2067--2080},
  year={2026},
  doi={10.1038/s41593-026-02382-1}
}

@article{fedorenko2024language,
  title={Language is primarily a tool for communication rather than thought},
  author={Fedorenko, Evelina and Piantadosi, Steven T. and Gibson, Edward A. F.},
  journal={Nature},
  volume={630},
  number={8017},
  pages={575--586},
  year={2024},
  doi={10.1038/s41586-024-07522-w}
}

@article{mahowald2024dissociating,
  title={Dissociating language and thought in large language models},
  author={Mahowald, Kyle and Ivanova, Anna A. and Blank, Idan A. and Kanwisher, Nancy and Tenenbaum, Joshua B. and Fedorenko, Evelina},
  journal={Trends in Cognitive Sciences},
  volume={28},
  number={6},
  pages={517--540},
  year={2024},
  doi={10.1016/j.tics.2024.01.011}
}

@article{kemp2009structured,
  title={Structured statistical models of inductive reasoning},
  author={Kemp, Charles and Tenenbaum, Joshua B.},
  journal={Psychological Review},
  volume={116},
  number={1},
  pages={20--58},
  year={2009},
  doi={10.1037/a0014282}
}

@article{tenenbaum2011grow,
  title={How to grow a mind: statistics, structure, and abstraction},
  author={Tenenbaum, Joshua B. and Kemp, Charles and Griffiths, Thomas L. and Goodman, Noah D.},
  journal={Science},
  volume={331},
  number={6022},
  pages={1279--1285},
  year={2011},
  doi={10.1126/science.1192788}
}

@article{lake2015human,
  title={Human-level concept learning through probabilistic program induction},
  author={Lake, Brenden M. and Salakhutdinov, Ruslan and Tenenbaum, Joshua B.},
  journal={Science},
  volume={350},
  number={6266},
  pages={1332--1338},
  year={2015},
  doi={10.1126/science.aab3050}
}

@inproceedings{camposampiero2026scalable,
  title={Scalable evaluation and neural models for compositional generalization},
  author={Camposampiero, Giacomo and Barbiero, Pietro and Hersche, Michael and Wattenhofer, Roger and Rahimi, Abbas},
  booktitle={Advances in Neural Information Processing Systems},
  volume={38},
  pages={8219--8279},
  year={2025},
  publisher={Curran Associates Inc.},
  doi={10.52202/085713-0252}
}

@misc{beger2025ai,
  title={Do {AI} models perform human-like abstract reasoning across modalities?},
  author={Beger, Claas and Yi, Ryan and Fu, Shuhao and Denton, Kaleda and Moskvichev, Arseny and Tsai, Sarah W. and Rajamanickam, Sivasankaran and Mitchell, Melanie},
  year={2025},
  note={Preprint at \url{https://arxiv.org/abs/2510.02125}}
}

@inproceedings{wust2026synthesizing,
  title={Synthesizing visual concepts as vision-language programs},
  author={W{\"u}st, Antonia and Stammer, Wolfgang and Shindo, Hikaru and Helff, Lukas and Dhami, Devendra Singh and Kersting, Kristian},
  booktitle={Proceedings of the IEEE/CVF Conference on Computer Vision and Pattern Recognition},
  year={2026},
  publisher={IEEE}
}

@article{mitchell2026six,
  title={Six principles for evaluating cognitive capabilities in {AI} models},
  author={Mitchell, Melanie},
  journal={AI Magazine},
  volume={47},
  number={2},
  pages={e70061},
  year={2026},
  doi={10.1002/aaai.70061}
}

@article{goodale2026fodor,
  title={Fodor and {Pylyshyn's} systematicity challenge still stands},
  author={Goodale, Michael and Mascarenhas, Salvador},
  journal={Transactions of the Association for Computational Linguistics},
  volume={14},
  pages={1598--1611},
  year={2026},
  doi={10.1162/tacl.a.743}
}

@misc{woydt2025fodor,
  title={Fodor and {Pylyshyn's} legacy: still no human-like systematic compositionality in neural networks},
  author={Woydt, Tim and Willig, Moritz and W{\"u}st, Antonia and Helff, Lukas and Stammer, Wolfgang and Rothkopf, Constantin A. and Kersting, Kristian},
  year={2025},
  note={Preprint at \url{https://arxiv.org/abs/2506.01820}}
}

@article{irie2025overcoming,
  title={Overcoming classic challenges for artificial neural networks by providing incentives and practice},
  author={Irie, Kazuki and Lake, Brenden M.},
  journal={Nature Machine Intelligence},
  volume={7},
  number={10},
  pages={1602--1611},
  year={2025},
  doi={10.1038/s42256-025-01121-8}
}

@inproceedings{yang2025emergent,
  title={Emergent symbolic mechanisms support abstract reasoning in large language models},
  author={Yang, Yukang and Campbell, Declan and Huang, Kaixuan and Wang, Mengdi and Cohen, Jonathan D. and Webb, Taylor W.},
  booktitle={Proceedings of the 42nd International Conference on Machine Learning},
  series={Proceedings of Machine Learning Research},
  volume={267},
  pages={70515--70549},
  year={2025},
  publisher={PMLR}
}

@inproceedings{newman2020eos,
  title={The {EOS} decision and length extrapolation},
  author={Newman, Benjamin and Hewitt, John and Liang, Percy and Manning, Christopher D.},
  booktitle={Proceedings of the Third BlackboxNLP Workshop on Analyzing and Interpreting Neural Networks for NLP},
  pages={276--291},
  year={2020},
  publisher={Association for Computational Linguistics},
  doi={10.18653/v1/2020.blackboxnlp-1.26}
}

@inproceedings{razeghi2022impact,
  title={Impact of pretraining term frequencies on few-shot numerical reasoning},
  author={Razeghi, Yasaman and Logan IV, Robert L. and Gardner, Matt and Singh, Sameer},
  booktitle={Findings of the Association for Computational Linguistics: EMNLP 2022},
  pages={840--854},
  year={2022},
  publisher={Association for Computational Linguistics},
  doi={10.18653/v1/2022.findings-emnlp.59}
}

@article{mccoy2024embers,
  title={Embers of autoregression show how large language models are shaped by the problem they are trained to solve},
  author={McCoy, R. Thomas and Yao, Shunyu and Friedman, Dan and Hardy, Mathew D. and Griffiths, Thomas L.},
  journal={Proceedings of the National Academy of Sciences},
  volume={121},
  number={41},
  pages={e2322420121},
  year={2024},
  doi={10.1073/pnas.2322420121}
}

@article{smolensky1988proper,
  title={On the proper treatment of connectionism},
  author={Smolensky, Paul},
  journal={Behavioral and Brain Sciences},
  volume={11},
  number={1},
  pages={1--23},
  year={1988},
  doi={10.1017/S0140525X00052432}
}

@book{marcus2003algebraic,
  title={The Algebraic Mind: Integrating Connectionism and Cognitive Science},
  author={Marcus, Gary F.},
  publisher={MIT Press},
  address={Cambridge, MA},
  year={2001}
}

@article{kean2026evidence,
  title={Evidence from formal logical reasoning reveals that the language of thought is not natural language},
  author={Kean, Hope and Fung, Alexander and Jaggers, Paris and Chen, Jason and Rule, Joshua S. and Benn, Yael and Tenenbaum, Joshua B. and Piantadosi, Steven T. and Varley, Rosemary A. and Fedorenko, Evelina},
  journal={Proceedings of the National Academy of Sciences},
  volume={123},
  number={28},
  pages={e2520095123},
  year={2026},
  doi={10.1073/pnas.2520095123}
}

@inproceedings{thomm2024limits,
  title={Limits of transformer language models on learning to compose algorithms},
  author={Thomm, Jonathan and Camposampiero, Giacomo and Terzic, Aleksandar and Hersche, Michael and Sch{\"o}lkopf, Bernhard and Rahimi, Abbas},
  booktitle={Advances in Neural Information Processing Systems},
  volume={37},
  pages={7631--7674},
  year={2024},
  publisher={Curran Associates Inc.},
  doi={10.52202/079017-0245}
}

@inproceedings{wang2022interpretability,
  title={Interpretability in the wild: a circuit for indirect object identification in {GPT-2} small},
  author={Wang, Kevin Ro and Variengien, Alexandre and Conmy, Arthur and Shlegeris, Buck and Steinhardt, Jacob},
  booktitle={International Conference on Learning Representations},
  year={2023},
  url={https://openreview.net/forum?id=NpsVSN6o4ul}
}

@article{tang2026circuit,
  title={Circuit explained: how does a transformer perform compositional generalization},
  author={Tang, Cheng and Lake, Brenden and Jazayeri, Mehrdad},
  journal={PLOS One},
  volume={21},
  number={2},
  pages={e0340088},
  year={2026},
  doi={10.1371/journal.pone.0340088}
}

@inproceedings{makelov2024subspace,
  title={Is this the subspace you are looking for? an interpretability illusion for subspace activation patching},
  author={Makelov, Aleksandar and Lange, Georg and Geiger, Atticus and Nanda, Neel},
  booktitle={International Conference on Learning Representations},
  volume={2024},
  pages={26486--26515},
  year={2024}
}

@misc{chollet2019measure,
      title={On the Measure of Intelligence}, 
      author={François Chollet},
      year={2019},
      eprint={1911.01547},
      archivePrefix={arXiv},
      primaryClass={cs.AI},
      url={https://arxiv.org/abs/1911.01547}, 
}

@inproceedings{conklin2021meta,
  title={Meta-learning to compositionally generalize},
  author={Conklin, Henry and Wang, Bailin and Smith, Kenny and Titov, Ivan},
  booktitle={Proceedings of the 59th Annual Meeting of the Association for Computational Linguistics and the 11th International Joint Conference on Natural Language Processing (Volume 1: Long Papers)},
  pages={3322--3335},
  year={2021}
}

@article{binz2024meta,
  title={Meta-learned models of cognition},
  author={Binz, Marcel and Dasgupta, Ishita and Jagadish, Akshay K and Botvinick, Matthew and Wang, Jane X and Schulz, Eric},
  journal={Behavioral and Brain Sciences},
  volume={47},
  pages={e147},
  year={2024},
  publisher={Cambridge University Press}
}

@inproceedings{russin2024human,
  title={Human curriculum effects emerge with in-context learning in neural networks},
  author={Russin, Jacob and Pavlick, Ellie and Frank, Michael J},
  booktitle={Proceedings of the Annual Meeting of the Cognitive Science Society},
  volume={46},
  year={2024}
}

@inproceedings{mondorf2026compositional,
  title={Compositional-ARC: Assessing systematic generalization in abstract spatial reasoning},
  author={Mondorf, Philipp and Zhou, Shijia and Riedler, Monica and Plank, Barbara},
  booktitle={International Conference on Learning Representations},
  volume={2026},
  pages={108708--108744},
  year={2026}
}

@article{chang2025coverage,
  title={The coverage principle: A framework for understanding compositional generalization},
  author={Chang, Hoyeon and Park, Jinho and Cho, Hanseul and Yang, Sohee and Ko, Miyoung and Hwang, Hyeonbin and Won, Seungpil and Lee, Dohaeng and Ahn, Youbin and Seo, Minjoon},
  journal={arXiv e-prints},
  pages={arXiv--2505},
  year={2025}
}

@inproceedings{press2023measuring,
  title={Measuring and narrowing the compositionality gap in language models},
  author={Press, Ofir and Zhang, Muru and Min, Sewon and Schmidt, Ludwig and Smith, Noah A and Lewis, Mike},
  booktitle={Findings of the Association for Computational Linguistics: EMNLP 2023},
  pages={5687--5711},
  year={2023}
}

@inproceedings{mitchell2021challenges,
  title={Challenges of acquiring compositional inductive biases via meta-learning},
  author={Mitchell, Eric and Finn, Chelsea and Manning, Chris},
  booktitle={AAAI Workshop on Meta-Learning and MetaDL Challenge},
  pages={138--148},
  year={2021},
  organization={PMLR}
}

@article{chan2022data,
  title={Data distributional properties drive emergent in-context learning in transformers},
  author={Chan, Stephanie and Santoro, Adam and Lampinen, Andrew and Wang, Jane and Singh, Aaditya and Richemond, Pierre and McClelland, James and Hill, Felix},
  journal={Advances in neural information processing systems},
  volume={35},
  pages={18878--18891},
  year={2022}
}

@article{chan2022transformers,
  title={Transformers generalize differently from information stored in context vs in weights},
  author={Chan, Stephanie CY and Dasgupta, Ishita and Kim, Junkyung and Kumaran, Dharshan and Lampinen, Andrew K and Hill, Felix},
  journal={arXiv preprint arXiv:2210.05675},
  year={2022}
}
